# Introducing multiplex semantic networks as multifaceted representations of creative associative knowledge across multilingual samples

Edith Haim[1], Kurt Haim[2], Roger E. Beaty[3], Cynthia S. Q. Siew[4], and Massimo Stella[1]

[1] *CogNosco Lab, Department of Psychology and Cognitive Science, University of Trento, Rovereto, Trento*

[2] *Department of Science Education, University of Education Upper Austria, Linz, Austria*

[3] *Department of Psychology, The Pennsylvania State University, Pennsylvania, USA*

[4] *Department of Psychology, National University of Singapore, Singapore*

**Corresponding author:**

Edith Haim - email address: edith.haim@unitn.it

Trento University, Corso Bettini 31, 38068 Rovereto TN, Italy

## Abstract

Creativity is a complex cognitive ability that relies on knowledge organisation and retrieval from semantic memory. Yet most research uses a single task to measure it, capturing only a fraction of this complexity. This study investigates multiplex networks - layered semantic networks obtained from six cognitive tasks - as a more comprehensive approach to modelling the associative knowledge underlying creativity. We collected data from *N=518* individuals from four countries (Austria, USA, Singapore, and Italy). From their responses to verbal fluency, sentence-chain, free association, and narrative writing tasks, we constructed semantic networks and assembled them in a multiplex structure. AI persona-based responses provided a comparison baseline. Structural reducibility analyses showed that different task layers captured distinct, non-redundant information about semantic organisation, supporting the use of multiple tasks over any single one. The networks from high- and low-creative groups remained structurally distinct, while AI-generated networks showed near-identical structures regardless of creativity group. Finally, we used 12 features (network measures, emotional scores, and spreading activation simulations) in a machine learning model using ridge regression to predict individual creativity scores. The combination of structurally similar layers, as identified in the previous stage, improved a proof-of-concept prediction accuracy by 50%. Structural measures showed the highest feature importance with spreading activation dynamics providing additional predictive power. Together, these findings indicate that multiplex semantic networks capture a richer, cross-cultural picture of associative knowledge underlying creativity. We also release our diverse dataset and code to foster diverse computational approaches within the creativity community.

# Introducing multiplex semantic networks as multifaceted representations of creative associative knowledge across multilingual samples

## 1. Introduction

Creativity is a core component of human cognition. It is the ability to produce new ideas that are both novel and appropriate in a given context (Runco & Jaeger, 2012). It is related to important skills such as problem solving and flexible thinking. Therefore, creativity provides a critical cognitive resource enabling progress in science, technology and society as a whole (Ahmed & Feist, 2021). For instance, creative reasoning is important for generating hypotheses and identifying new connections between concepts in science. It also promotes innovation in technology through creating new systems or redesigning existing ones. Additionally, creativity is essential for decision-making, where it helps to address challenges such as climate change, public health crises and rapid development of digital technologies (Jacobson et al., 2016). Understanding the cognitive foundations of creativity is necessary for improving flexible thinking and designing novel AI-supported methods of measuring and modelling creativity.

Yet creativity is not a single skill. Even during relatively simple verbal tasks, various cognitive mechanisms need to work together. One mechanism is *associative flexibility*, the ability to form associations between ideas that are typically distant in terms of meaning (Beaty & Kenett, 2023). According to classical theories, creative individuals can produce remote associations that can lead to novel insights (Mednick, 1962). Another dimension is *executive control*, that supports goal-directed thinking by suppressing irrelevant ideas and choosing useful ones (Beaty & Silvia, 2012). Creative behaviour also builds on *linguistic knowledge* and *domain-specific memory* (Kenett et al., 2014). Among other cognitive components, *verbal fluency* reflects how easily individuals can access words in semantic memory. This process is comparable to efficient retrieval in information systems. Rather than only accessing common concepts, creative individuals have been found to produce remote associations (Benedek et al., 2017; Ganor et al., 2025). This may be guided by strong *semantic navigation skills*, which describes the ability to move through memory structures in flexible and non-linear ways (Ganor et al., 2025). Furthermore, the creative process is shaped by motivational and affective aspects. Creative performance is further shaped by motivation and mood. Positive mood has been found to broaden semantic exploration, while negative mood may narrow it (Forgeard & Elstein, 2014).

As discussed previously, these cognitive processes, or facets, interact with one another continuously (Beaty & Kenett, 2023). Cognitive operations do not function in separate isolated modules. On the contrary, they can be viewed as part of a dynamic system in which memory retrieval, semantic search, linguistic processing and affective signals interact with each other producing various creativity patterns. From a computational perspective, creativity emerges from the interaction between several sub-systems working under a number of constraints. Capturing this multi-dimensional interaction requires methods that enable us to integrate heterogeneous data - single words, short associations, full sentences or narratives - into a unified representation (Stella et al., 2024). Understanding the mechanisms that underlie these interactions is crucial for developing theories regarding creative thinking as a result of certain structural and organisational properties of semantic memory.

Network science offers such a framework. Specifically, cognitive network science models the organisation of knowledge by treating words, concepts or ideas as nodes linked through associative or syntactic relations (E. Haim et al., 2026; E. Haim & Stella, 2026; Kenett et al., 2014; Stella et al., 2024). This allows us to study the topology of semantic memory, providing insights into underlying cognitive processes. Research grounded in both associative theories of creativity (Mednick, 1962) and executive theories focusing on controlled recall processes (Beaty & Silvia, 2012) has revealed that creative individuals display semantic structures marked by flexible associations, lower modularity and easy access to remote concepts. Such easy accessibility is found either in terms of memory recalls (Kenett et al., 2014) or across semantic and phonological associations (Ganor et al., 2025; Samuel et al., 2023; Stella & Kenett, 2019). Network analysis, therefore, offers a powerful means of understanding how creativity is encoded within the architecture of memory.

However, approaches that model only single-layer networks, where concepts are linked by only one type of association, are not enough for capturing the complex processes of creativity. Recent research has highlighted that multiple cognitive layers must be considered when analysing creative thinking. Such layers can involve semantic associations, phonological similarities, syntactic structures and affective connotations. *Multiplex cognitive networks*, also termed *multiplex lexical networks*, are key instruments for the integration of multiple cognitive dimensions in one network framework that treats each layer as interconnected but distinct (Stella et al., 2024). In a multiplex cognitive network, each layer encodes a different type of relation between the same set of concepts, allowing researchers to assess how cognitive processes coordinate across linguistic, associative and emotional domains (Aleta et al., 2026). This approach is promising for studying creativity since novel ideas often emerge from the interplay of multiple representational layers. The power of multiplex networks in studying creativity was demonstrated in earlier research (Ganor et al., 2025; Samuel et al., 2023; Stella & Kenett, 2019), indicating how various cognitive layers jointly support flexible and innovative thinking.

In order to study creativity, various tasks have been used to elicit verbal and conceptual output. These vary in terms of instructions, cognitive load and linguistic output. For instance, word fluency tasks, free associations and continuous associations are designed to tap directly into associative knowledge (Kenett, 2025). In other tasks, such as the Alternative Uses Task (Gilhooly et al., 2007), sentence-chain production (E. Haim et al., 2026) and short narratives

(E. Haim, Fischer, et al., 2026), contextual and syntactic constraints are involved. It is not clear whether each task accesses different components of semantic memory and executive functions, or whether they provide redundant information about participants' semantic knowledge organisation.

From a methodological point of view, these considerations are relevant for deciding whether multiple tasks measure the same cognitive network or different network layers, each one corresponding to a single task. In the first case, researchers could streamline creativity assessment by selecting a smaller set of tasks that maximises information while minimising participant burden. Conversely, if tasks reveal complementary structures, a multiplex approach would be necessary in order to account for the complexity of creative thought (Beaty & Kenett, 2023). Multiplex network analysis allows one to quantify the structural similarity or distinctiveness of layers via reducibility techniques to assess whether layers can be collapsed without losing relevant information (Stella et al., 2024).

While investigating network structure, it is important to explore further whether a multi-layer network contains enough information to make predictions about creativity at the individual level. Prior work has demonstrated that individual creativity scores can be predicted from neural connectivity patterns (Beaty et al., 2018) and from structural and emotional features of text-derived semantic networks (E. Haim, Fischer, et al., 2026). Furthermore, spreading activation dynamics, which simulate how concepts activate one another during semantic retrieval (Collins & Loftus, 1975), have recently been applied to predict creativity ratings from text-based networks (Passaro et al., 2026). In addition, spreading activation on a semantic-phonological multiplex network was found useful to make predictions about the difficulty of items in the Remote Associates Test (Citraro et al., 2026). It remains an open question whether structural, emotional and dynamical features extracted from a multiplex, multi-task network can predict individual creativity in a machine learning model and which task layers contribute most.

The present work uses a multiplex network approach with six different tasks designed for eliciting associative, linguistic and conceptual behaviour: two verbal fluency tasks, two sentence-chain tasks (Woseco), a continuous association task (behavioural forma mentis network), and one text writing task. Each task was aligned by topic to ensure any detected differences in network structure are due to differences between the tasks and not due to content domains. Data was collected in four cultural settings (Austria, USA, Singapore, and Italy) to assess the generality of results across linguistic and educational environments. Participants were classified into high- and low-creative groups based on their originality scores in the Alternative Uses Task, thus allowing the investigation of whether differing levels of creativity manifest in distinct network topologies. AI-generated responses were based on human persona profiles and provided a synthetic comparison condition, similar to digital shadowing (Ardebili & Stella, 2026).

In the second analytical step, structural network measures, emotion features and spreading activation dynamics were extracted from individual-level networks. These were used as inputs for a proof-of-concept machine learning aimed at predicting individual creativity scores on the Alternative Uses Task, assessing which network layers and feature types contain the most relevant information for such predictions.

Across both steps, our research aims are threefold:

(i) to determine whether the above-listed tasks capture convergent or divergent aspects of semantic memory and creative cognition;

(ii) to provide an evidence-based recommendation on whether creativity research should rely on a streamlined set of tasks or adopt a multiplex framework;

(iii) to investigate whether structural and dynamical features of individual-level networks can predict individual creativity scores, and to identify which network layers and feature types contribute most to these predictions.

By systematically analysing structural reducibility across tasks, creativity groups and cultural contexts, and by linking individual network features to creativity outcomes through interpretable machine learning, this study contributes to a mechanistic understanding of how creative thought is encoded in cognitive networks and how best to represent the cognitive architecture underlying it.

## 2. Methodology

In this section, we outline the methodological pipeline used to construct and compare the cognitive network representations derived from the different tasks across all datasets. Figure 1 (top) outlines the different instructions. From each task, we constructed semantic networks specifically for the output format (e.g., co-occurrence networks for fluency tasks; syntactic dependency networks for the text-based task). As shown in Figure 1 (middle) for the verbal fluency, Woseco and Woseco forma mentis we constructed two networks for each task - one for the prompt “school” and one for “science”. This resulted in a total of eight single-layer networks per dataset.

To examine how these layers collectively represent participants’ associative knowledge, we integrated the eight networks into a multiplex lexical network, where each network constitutes a distinct layer. First, we considered the unaggregated multiplex structure, i.e. the full configuration in which all eight layers are retained separately (Figure 1, bottom left). Then we compared this against all possible aggregated multiplex variants, where two or more layers are merged if they exhibit similar structural properties (Figure 1, bottom right). Such an aggregation of layers serves to reduce model complexity, while retaining the maximal amount of structural information (De Domenico et al., 2015). Thus, this allows us to evaluate whether some tasks capture redundant aspects of semantic memory.

These steps in the pipeline are followed for every dataset in our study separately (for each nationality and humans vs simulations).

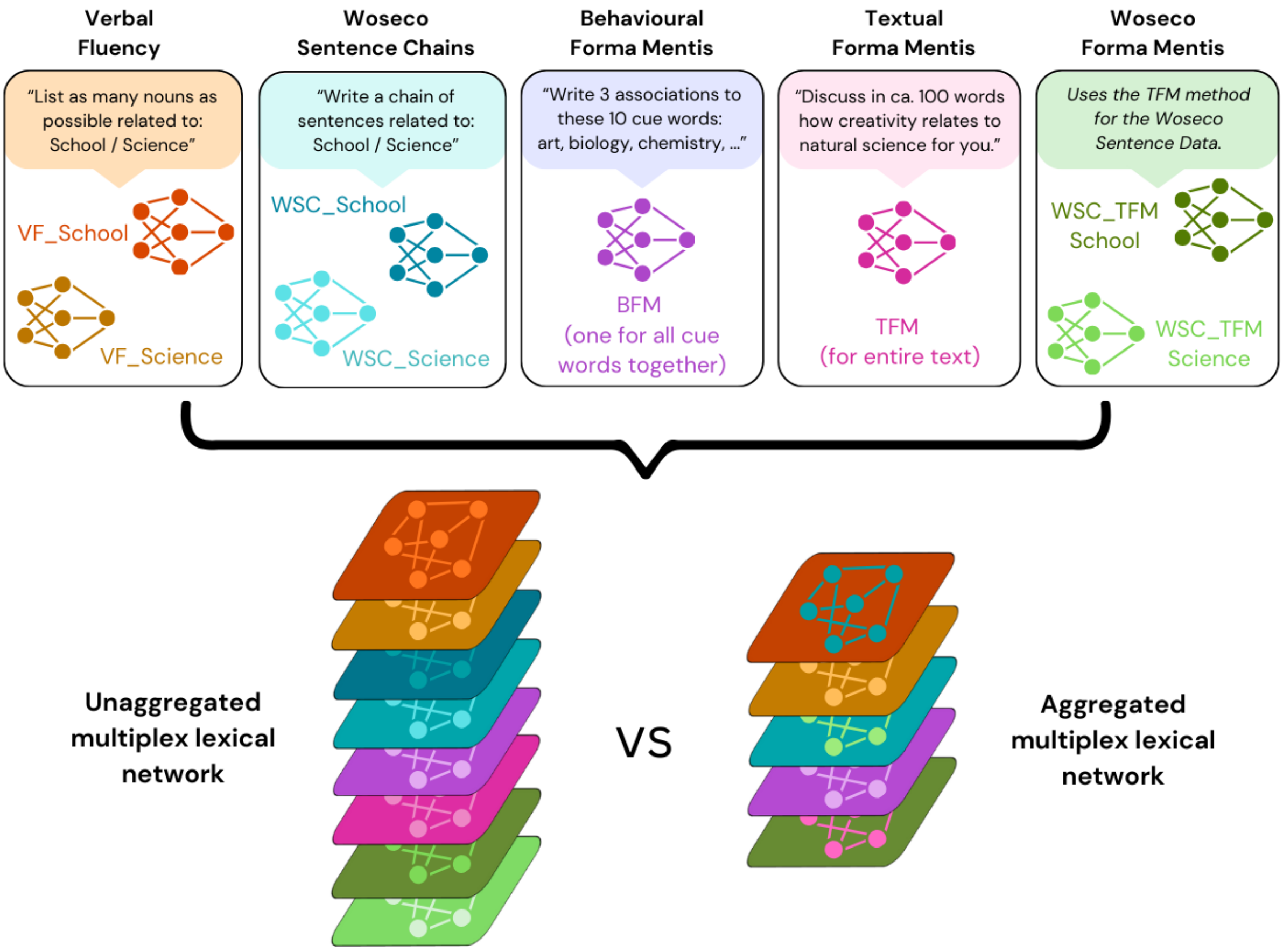

**Figure 1**. Overview of the methodological pipeline. The upper panel illustrates the six tasks and instructions used in the study (two verbal fluency, two Woseco, behavioural forma mentis, textual forma mentis). Only for the Woseco forma mentis networks no new data was collected. In this case, the sentence chains collected in the Woseco tasks were merged into a full text per participant and analysed like a short narrative with the textual forma mentis method. The lower panel contrasts the unaggregated multiplex network (left), where all layers are kept separate, with an example aggregated multiplex network (right). In the aggregated multiplex network, structurally similar layers are merged to reduce complexity. The networks differing in colour from the layer colour represent an example for layers that have been merged. For instance, the turquoise network on the upper aggregated orange layer indicates that the networks VF_School and WSC_School have been merged. This illustrative example does not reflect the empirical results but visualises the general principle of the aggregation procedure tested in our analyses.

## 2.1 Participants and Groups

Data for this research was collected in four studies conducted in Austria, the United States of America, Singapore and Italy. The complete data used in this study can be found publicly available on OSF (https://osf.io/ns952/overview). All studies followed the same experimental design and research questions, allowing methodological comparability across those four cultural contexts. The participant samples were largely composed of university students enrolled in STEM-related or psychology courses, i.e. convenience sampling.

The study was approved by the Ethics Committee of the University of Trento (Protocol 2024-021, approved 21 March 2024) and the Ethics Committee of the National University of Singapore (Ref. 000498). Data collection at the remaining institutions was conducted in accordance with their respective institutional ethical guidelines.

In the following, the participant characteristics of each study are described in detail.

Study 1 - Austria

The Austrian dataset consisted of 132 teaching students enrolled in teacher education programs. Participants were recruited from the University of Education Upper Austria via university mailing lists and convenience sampling, and participated on laptops or computers in a university classroom. The average age was 24 years (ranging from 18 to 51) with 57% female participants. Almost all students had Austrian nationality, with only 5 students coming from other European countries. We recruited 96 students in the Bachelor level, 35 were studying at the Master level, and one was pursuing a Doctorate degree. They were recruited from various teaching disciplines, with focus on natural science subjects. In Austria, teacher education programs require teachers in training to obtain qualification in at least two teaching subjects (Proyer et al., 2022). Thus, Austrian teachers typically teach and get training in two subjects, resulting in overlapping subject distributions.

Our participants represent a broad spectrum of STEM and related disciplines typical of Austrian secondary education. The most common subjects in our sample were Mathematics and Informatics (n=84), followed by Biology and Nutritional Science (n=34), Chemistry (n=25), and Physics (n=25). Smaller subsets were studying Languages (n=19), Psychology (n=6), and 40 different less frequent subjects in the “Other” category.

Study 2 - USA

The U.S. dataset comprised 160 undergraduate students who were recruited from Pennsylvania State University. The average age was 19 years (ranging from 18 to 24), with 84% female participants. In terms of self-reported nationality and ethnic background, 82% identified as White or North American, 5% as Asian, 4% as South American, 4% as European, 2% as African American and 3% as biracial.

All students were enrolled in undergraduate degree programs. The largest group of students were enrolled in programs related to Biology, Medicine and Nursing (n=63), followed by Psychology (n=40). Smaller numbers of participants studied in fields related to Advertisement and Management (n=16), Criminology, Forensics and Security (n=10), Art and Media (n=8), as well as further subjects that were less frequent in our dataset (n=23). Participants were recruited through the university's Sona System, an online research platform. They completed the study online on their own devices, at a time of their choosing.

Study 3 - Singapore

For the Singaporean dataset, 148 undergraduate students from the National University of Singapore (NUS) participated in our study. All students were enrolled at the Bachelor level. The average age was 20 years (ranging from 18 to 51). Our sample included 82% female, 17% male and 1% participants identifying as diverse in gender identity. Regarding nationality, 86% were Singaporean citizens, 12% originated from other Asian countries (e.g., China, India, South Korea, Malaysia), and 2% came from Europe.

The majority were enrolled in Psychology (n=80), and courses related to Health Sciences such as Medicine, Nursing and Life Science (n=40). Further study fields were Business and Economics (n=14), Social Sciences such as Languages, Politics and other humanities-related fields (n=13), and Engineering and Technology (n=6). Only two participants indicated unspecified study programs. Participants were recruited via the university's Sona research participant pool. As with the US sample, participants completed the study remotely online.

Study 4 - Italy

For the Italian dataset, 78 undergraduate students from the University of Trento participated in the study. All participants were enrolled in first-year Bachelor programs in the fields of cognitive psychology. The average age was 19 years (ranging from 18 to 23) and the sample included 76% female participants. Regarding nationality, 97% were Italian citizens and 2 participants reported nationalities from outside Europe. Participants were recruited through convenience sampling during a regular class at the University of Trento on their own devices.

Study 5 - AI simulations

In addition to the human datasets, we generated four synthetic comparison datasets using a large language model. The purpose of including AI-generated data is to create a null-model baseline against which the structure of the human networks could be interpreted. For each of the four national datasets (Austria, USA, Singapore, Italy), we created a set of AI personas based on the anonymised demographic profiles of the human participants. These personas included nationality, age, gender, and study field, as well as the highest education level of the parents. The same tasks, with slightly adapted instructions for the AI (e.g., replacing time limits with maximum number of responses) were completed by an AI model acting in the role of each persona.

The synthetic dataset was produced using Qwen3-4B-Instruct, a 4-billion-parameter instruction-tuned language model released on 25th July 2025 (Yang et al., 2025). The Instruct version of Qwen3 is designed to follow prompts reliably but does not incorporate chain-of-thought reasoning or multi-step inference. This configuration ensured that the model responded directly to the task instructions in a manner comparable to human participants. Although larger models were tested during preliminary piloting (i.e., Gemma 3 27B, Mistral Small 3.1), they generated substantially noisier and less consistent outputs for our task structures. Qwen3-4B-Instruct therefore provided the most stable and interpretable synthetic dataset for the purposes of this study.

Since the AI data is only meant to serve as a null-model baseline, we do not aim to draw conclusions about comparisons between human cognitive networks and AI systems more broadly. For this reason, we used only a single AI model (namely Qwen3-4B-Instruct) to get a controlled reference condition, rather than comparing the performance of multiple AI models.

High creative vs low creative groups

To investigate differences between participants with higher and lower levels of creativity, participants were divided into a high-creative and a low-creative group. Group assignment was based on each participant's total score in the Alternative Uses Task. The total score was calculated as the mean of their originality scores across both task items (see Section *2.3 Calculation of Creativity Score* for details on the scoring procedure). To obtain comparable group sizes within each dataset, we applied a median split on the total AUT score for each dataset separately. Participants with scores equal to or above the median were classified as high-creative, and those below the median as low-creative. This approach ensures a balanced distribution of participants per group and avoids skewed group sizes that may bias statistical comparisons (Iacobucci et al., 2015).

We chose a median split per dataset over fixed absolute creativity thresholds across datasets because the primary goal of this analysis was not to compare creativity levels between countries, but to investigate relative differences within each sample. By contrasting the top 50% creative participants against the bottom 50% of the same group, we examine within-group contrasts in the same cultural context for each dataset. This procedure aligns with previous research using relative thresholds like median split rather than fixed cut-offs (Beaty & Silvia, 2012).

## 2.2 Task Design

The task design was identical across all four datasets, differing only in the language in which it was conducted. The study was programmed in PsychoPy (version 2024.1.5), an open-source platform for designing and running behavioural and cognitive experiments (Peirce, 2007; Peirce et al., 2019). Participation took place online via the Pavlovia platform on the participants' own devices. The instructions were presented in German (study 1), English (study 2 & 3) and Italian (study 4). Each version was piloted with native speakers to ensure it was easily understandable and sounded natural. The translations of the instructions can be found in the Supplementary Materials on OSF.

Participants completed a set of tasks to measure cognitive, linguistic and emotional aspects of creative thinking. These included network-based tasks (verbal fluency, Woseco, behavioural and textual forma mentis networks), as well as Alternative Uses tasks for two items. Each of these tasks captures a distinct aspect of semantic memory and cognitive processing relevant to creative thought. Table 1 summarizes the primary cognitive processes captured by each measure, as reported and discussed in past publications introducing or testing such cognitive tasks.

| Task | Primary cognitive process targeted | Reference |
|---|---|---|
| Verbal Fluency | Semantic search, associative retrieval | Siew & Guru (2023) |
| Woseco Sentence Chains | Conceptual integration, controlled retrieval, semantic and syntactic relations | E. Haim et al. (2024) |
| Behavioural Forma Mentis Task | Spontaneous associative structure, semantic navigation, affective dimensions | Stella et al. (2019) |
| Textual Forma Mentis Task | Discourse-level processing, contextualised semantic organisation, affective dimensions | Stella (2020) |
| Woseco Forma Mentis | Conceptual integration plus affective dimensions | / [New Approach] |
| Alternative Uses Task | Divergent thinking, originality | Gilhooly et al. (2007) |

**Table 1**. Cognitive dimensions targeted by each task.

Verbal fluency tasks provide direct access to associative retrieval under time constraints and are widely used to assess semantic search efficiency (Benedek et al., 2017; Siew & Guru, 2023). The Woseco sentence-chain task captures flexible conceptual integration with minimal syntactic structure (E. Haim et al., 2026), while its forma mentis variant adds explicit affective and evaluative information to the produced concepts. Behavioural forma mentis networks reflect spontaneous associative structure (E. Haim, Bergh, et al., 2026; Stella et al., 2019), and textual forma mentis networks derived from short texts quantify contextualised semantic organisation that integrates linguistic and affective dimensions (E. Haim, Fischer, et al., 2026). The Alternative Uses Task serves as an established behavioural measure of divergent thinking and originality (Gilhooly et al., 2007). In the following, each task is described in detail and how it was administered.

Verbal Fluency

The verbal fluency task assesses participants' capacity to generate as many words as possible within a specified topic and time limit (Siew & Guru, 2023). Fluency tasks are widely used in cognitive and creativity research to provide insight into the breadth and organisation of associative knowledge, by reconstructing semantic networks based on the associative relations between concepts (E. Haim et al., 2026; Lai et al., 2024; Siew & Guru, 2023). In the present study, participants completed two category fluency tasks on the topics "school" and "natural science". The chosen topics align with the study's focus on scientific creativity in educational contexts and are recurring stimuli throughout the other tasks. Participants had a time limit of one minute for each topic and were instructed to list as many nouns as possible relating to these topics. To stimulate divergent thinking, participants were explicitly encouraged to "be creative" and to consider uncommon associations rather than searching only for typical responses. This explicit instruction for creative output was based on prior evidence showing that prompting participants to adopt a creative mindset during a task can increase the level of creativity of their output (Chen et al., 2005; Runco et al., 2005).

Woseco Sentence Chains

The Word-Sentence-Construction (Woseco) task is a sentence-chain paradigm, originally introduced by Haim and Aschauer (2022) as part of the Scientific Creativity in Practice (SCIP) program. SCIP is an educational framework designed to foster associative and divergent thinking in STEM learning. In a Woseco task, participants write a sequence of consecutive sentences in which each sentence must incorporate a key concept from the previous one. For instance, a participant might produce the following sentence-chain: "The study of biology includes **anatomy**. In anatomy we learn about **bones**. Bones are crucial in the **body**. The body…" In this example, the word highlighted in bold is a new concept introduced in the sentence, which is then picked up in the following sentence (highlighted through underlining). This creates a sentence chain, where sentences are linked together through repeated concepts (K. Haim & Aschauer, 2022). From this data, one can investigate how participants navigate semantic knowledge while maintaining semantic coherence across sentences. The ability to flexibly connect concepts within a given domain is closely linked to creativity-related cognitive processes (E. Haim et al., 2026).

In this study, participants completed two Woseco tasks using the same prompts as in the verbal fluency task ("school" and "science"). This allowed comparison of network structures across tasks within the same thematic frame. For each topic, participants had three minutes to produce as many consecutive sentences as possible. Again, participants were explicitly instructed to be creative and use remote associations.

### Behavioural Forma Mentis Networks

Forma mentis networks can be constructed in two distinct ways: from behavioural data (Stella et al., 2019) or textual data (Stella, 2020). Textual forma mentis networks (TFMNs, see below) are derived from samples of written narratives or discourse (Semeraro et al., 2025). In contrast, behavioural forma mentis networks (BFMNs) are built from a continued free association task, in which participants respond to a cue word with the first three words that come to mind (De Deyne et al., 2013; E. Haim, Bergh, et al., 2026). The free associations to the cue words collected in this way form the semantic structure of the BFMN network. Nodes represent cue words and associations, and edges link each cue to its three responses. In addition to the network structure, BFMNs also incorporate an emotional dimension. After generating associations, participants rate the emotional valence of each cue and response on a 1-5 point scale, with 1 indicating "very negative" and 5 "very positive" (Stella et al., 2019). This method has been shown to reflect broader patterns of knowledge organisation and affective perception, including how negative concepts cluster and how emotional framing shapes math anxiety (Stella, 2022; Stella et al., 2019).

In the present study, participants completed the BFM task using the following ten cue words related to education and STEM: art, biology, chemistry, creativity, life, mathematics, physics, school, sustainability, university. The stimuli were presented in randomised order to minimise sequence effects.

Textual Forma Mentis Networks

Textual forma mentis networks represent another form of forma mentis networks that are based on text rather than free association cues. TFMNs model the semantic and emotional structure of short texts by linking words according to their syntactic dependencies (Stella, 2020). Unlike the behavioural form, which collects valence information directly from the participants, TFMNs employ the EmoLex database (Mohammad & Turney, 2013) of emotion words to assign valence and compute emotion z-scores based on Plutchik's eight basic emotions (E. Haim, Fischer, et al., 2026; Plutchik, 1980). By considering syntactic dependencies instead of a fixed adjacency window, TFMNs provide a linguistically richer representation of sentence structure than linear co-occurrence networks (Passaro et al., 2026) and even come within a simple-to-use Python package for language analysis, called EmoAtlas (Semeraro et al., 2025).

TFMNs have been applied to examine mindset structure in domains such as public discourse on STEM gender gaps (Stella, 2020) and creativity in short narratives (E. Haim, Fischer, et al., 2026; Semeraro et al., 2025). In the present study, participants were instructed to write a short text (minimum ~ 100 words) in six minutes discussing the question: "To what extent does creativity play a role for you in science?"

Alternative Uses Task

Finally, we used the Alternative Uses Task (AUT), a widely established measure of divergent thinking, to assess individuals' ability to generate novel and varied ideas (Guilford, 1967). In this task, participants are asked to produce as many unusual and alternative uses as possible for a common object within a limited time. The AUT is considered an indicator of creative potential because it taps into core components of divergent thinking (Guilford, 1967). These components include the capacity to generate many ideas (fluency), to shift between conceptual categories (flexibility), and to produce uncommon responses (originality) (Beaty et al., 2022).

In the present study, participants completed the AUT for two everyday objects: a fork and a bottle. The order of the two objects was randomised across participants to avoid systematic order effects. Participants were instructed to come up with as many unusual and creative uses as they could within a 3-minute time limit. The responses were processed to derive an originality-based score, which was the basis for classifying participants into high-creative and low-creative groups. Section 2.3 discusses in detail how we calculated the AUT scores for each participant.

## 2.3 Calculation of Creativity Scores

Creativity scores for each participant were derived from responses to the Alternative Uses Task. Traditionally, the AUT is scored using human raters or category-based metrics. However, these scoring approaches have been found to be influenceable by factors such as response length and elaboration (Beaty et al., 2022; Forthmann et al., 2019). Thus, approaches using AI support for scoring have become increasingly popular. However, these AI-supported tools also face some limitations such as model-specific biases (Beaty et al., 2022). Therefore, to enhance the reliability and robustness of our creativity estimates, we employed two independent automated scoring systems: CLAUS (Patterson et al., 2024) and OCSAI (Organisciak et al., 2023).

CLAUS (Cross-Lingual Alternate Uses Scoring) is a multilingual creativity-scoring model based on a fine-tuned XLM-RoBERTa architecture trained on AUT responses in 12 languages (Patterson et al., 2024). It is available through the Creativity Assessment Platform (CAP, https://cap.ist.psu.edu/claus, Last Access 03. December 2025) and returns originality scores in the 0-1 range (0 = uncreative; 1 = highly creative) (Patterson et al., 2025).

OCSAI (Open Creativity Scoring with Artificial Intelligence) is built on a GPT-4o-mini-based supervised model trained on approximately 27.000 human-rated AUT responses (Organisciak et al., 2023). It is available on the Open Creativity Scoring website (https://openscoring.du.edu/uploadllm, Last Access 03. December 2025) and provides originality scores on a 1-5 scale (1 = uncreative; 5 = highly creative). To ensure comparability with values provided by CLAUS, OCSAI values were linearly rescaled to a 0–1 range.

Previous studies have compared the performance of both AI scoring approaches and found positive correlations between LLM ratings and human scorings, with OCSAI performing better than CLAUS (Primi et al., 2026; Saretzki et al., 2025). Using these two distinct automated raters - one multilingual transformer-based model and one supervised model trained on a large expert-rated dataset - reduces the likelihood of model-specific scoring artefacts and provides a more stable, convergent estimate of participants' originality (Primi et al., 2026; Saretzki et al., 2025). Both systems provide support for the three languages present in our dataset (English, German, Italian).

For each AUT item (fork and bottle), we obtained an originality score for every response and averaged them per participant, resulting in an average AUT score per item. This procedure was followed by both rating systems (CLAUS and OCSAI) separately, providing averaged scores within each system. Finally, we calculated an overall creativity score as the mean of all averaged scores per participant for each item and rating system. Based on this overall score, participants were classified into high-creative and low-creative groups via a median split. This relative classification ensures balanced group sizes and avoids fixed creativity thresholds across our culturally diverse datasets.

### 2.4 Network Construction

All analyses were carried out and all networks were constructed using the original language of each dataset. We processed all data in the original language as translations can distort lexical relations, alter syntactic structure and introduce incorrect semantic similarities or differences. Working directly in the source languages preserves the associative and linguistic patterns produced by the participants without introducing artefacts of cross-language alignment. For each dataset, we lowercased all words during data cleaning to avoid duplicates in the networks from spellings in upper- or lowercase letters. Even though in German nouns are spelled with an initial capital letter, we lowercased these as well for consistency.

Below, we outline the procedures used to construct the networks for each task type. Each task produced one or more single-layer networks, which were later integrated into the multiplex framework described in Figure 1.

#### Verbal Fluency

Semantic networks from the verbal fluency data were constructed following the general principles of correlation-based networks as used in the SemNA package by Christensen and Kenett (2023). We implemented the procedure in Python and made the full code available on OSF (https://osf.io/ns952/overview). Each group in the datasets (high-creative students, low-creative students) was processed separately to capture the associative structure characteristic of each group.

Responses were first lemmatised, and a binary response matrix was created, with rows representing participants and columns representing unique words. A word was retained if it was produced by at least two participants within that group, which reduces idiosyncratic noise and ensures stable association estimates (Christensen & Kenett, 2023). For each group-specific binary matrix, an association matrix was computed using cosine similarity, a commonly used association metric for binary fluency data because it yields values between 0 and 1 and avoids negative associations (Christensen et al., 2018). The resulting association matrix was then filtered using the Triangulated Maximally Filtered Graph (TMFG) algorithm (Massara et al., 2016), an information-filtering network that retains the strongest associations under planarity constraints and produces a sparse but structurally informative semantic network (Christensen & Kenett, 2023).

In contrast to some implementations of correlation-based networks (Christensen & Kenett, 2023), node-equating across groups was intentionally not applied. In our code, networks are constructed independently for each group, and equating is only available as an option but not used in constructing the final networks. This choice avoids artificially shrinking the networks and is appropriate given that the analyses focus on within-group structure and comparisons across different tasks (fluency, textual forma mentis, behavioural forma mentis), rather than strict cross-group comparisons for the fluency task. The resulting TMFG-filtered networks derived from each group were used for all subsequent analyses.

<u>Woseco</u>

For the Woseco task, networks were constructed using the same correlation-based pipeline as for the verbal fluency data, but an additional preprocessing step was required to extract the relevant words from the sentence chains (E. Haim et al., 2026). To extract the recurring words from the sentence chains, each participant's Woseco responses were parsed with spaCy, and all nouns, proper nouns, and verbs were lemmatised and extracted from each sentence. To identify the sequential linking intended by the task, an algorithm matched each sentence to the previous one by selecting the concept with the highest semantic similarity to the prior sentence's concepts. This produced a single ordered list of linking concepts for each participant, functionally analogous to a fluency list.

After this extraction, the Woseco networks were generated identically to the fluency networks: we built binary response matrices, computed cosine-based association matrices, and applied the Triangulated Maximally Filtered Graph to obtain semantic networks.

Behavioural Forma Mentis Networks

The behavioural forma mentis networks (BFMNs) were constructed directly from participants' free association responses. Only complete rows containing a cue word and all three associative responses were retained; missing valence ratings for existing words were replaced with a neutral value of 3. Associative links were extracted by combining all cue–association pairs into a single edge list. For the group networks, links occurring only once were removed, and duplicates were collapsed. Later, when constructing individual networks for the machine learning analysis, we also retained words that occurred only once within the individual's network (E. Haim, Bergh, et al., 2026).

The resulting set of connections was used to build an unweighted directed graph representing the semantic layer of the network. To add the emotional layer, each concept's valence distribution was compared against the distribution of all remaining words using a Wilcoxon rank-sum test ($\alpha = 0.1$). Concepts with significantly higher or lower median valence were labelled *positive* or *negative*, respectively; all others were classified as *neutral*. These valence labels were used to colour nodes and edges according to the emotional relations they represent (positive = blue, negative = red, neutral = grey). Finally, the empirical clustering coefficient was evaluated against a null distribution generated from 500 configuration-model networks preserving the original degree sequence. This allowed assessment of whether the observed clustering coefficients exceeded chance expectations (E. Haim, Bergh, et al., 2026).

### Textual Forma Mentis Networks

Textual forma mentis networks (TFMNs) were constructed for each participant group (high-creative and low-creative) using the EmoAtlas library (Semeraro et al., 2025). Following standard preprocessing (tokenisation, lemmatisation), each sentence was parsed into a dependency tree, and all non-stopwords within three syntactic steps were connected. This dependency window captures structurally meaningful relations that are not detectable via linear co-occurrence (Passaro et al., 2026). Sentence-level networks were then merged into a text-level network, with lemmatisation ensuring that different inflected forms of a word mapped to a single conceptual node (Semeraro et al., 2025).

TFMNs also include an emotional layer. Valence (positive, negative, neutral) was assigned using the EmoLex lexicon (Mohammad & Turney, 2013). Broader affective information was quantified using emotion z-scores based on Plutchik's eight basic emotions (Plutchik, 1980), computed relative to null-model expectations (Semeraro et al., 2025).

### Woseco-Textual Forma Mentis Networks

We also used a novel exploratory approach that applies the textual forma mentis (TFM) method to the Woseco task. Unlike the standard Woseco network construction where only the repeated linking concepts are retained, this approach preserves the full linguistic content of participants' sentence chains. The Woseco networks described previously reduce the sentences to a sequence of target concepts, omitting the rich syntactic and semantic information contained in pronouns and additional concepts. This loss is especially pronounced in longer or more syntactically complex sentences.

To capture this broader structure, all sentences produced in the Woseco task were first concatenated into a single short text per participant. We then applied the standard TFMN pipeline described in the previous subsection: preprocessing, dependency parsing, construction of syntactic networks within a three-step dependency window, and integration of valence and emotion labels. This exploratory variant allows us to examine whether TFMNs can serve as a richer alternative representation of the associative and linguistic processes elicited by the Woseco task.

## 2.5 Calculating Network Measures

For the TFM and BFM networks for each participant, a set of structural network measures was computed on the largest connected component (LCC) of the individual semantic-level network. TFM measures were extracted using *NetworkX* (version 3.6.1) in Python, and *igraph* (version 2.1.4) in R was used for the BFM layer, with the community detection library using the Louvain implementation for modularity. The extracted measures are defined below.

**Diameter:** The diameter $D$ is the maximum shortest-path-distance between any two nodes (M. Newman, 2018).

$$D = max_{i,j}\, d(i,j)$$

where $d(i,j)$ is the shortest-path-length between nodes $i$ and $j$. A larger diameter indicates the presence of semantically distant concept pairs, reflecting broader associative reach.

**Average shortest path length (ASPL):** ASPL is the mean shortest-path-distance across all node pairs (M. Newman, 2018).

$$ASPL = \frac{2}{n(n-1)} \sum_{i \neq j} d(i,j)$$

Lower ASPL reflects more efficient information flow. In fluency-derived networks, lower ASPL has been linked to higher creative performance (Kenett et al., 2014; Siew & Guru, 2023). In contrast, higher ASPL values in narrative networks have been associated with greater creativity, possibly reflecting broader thematic organisation (E. Haim, Fischer, et al., 2026).

**Average clustering coefficient**: The local clustering coefficient $C_i$ of node $i$ measures what proportion of its neighbours are also connected to each other (E. Haim & Stella, 2026).

$$C_i = \frac{|\{(k,l) \in E: k,l \in N_i\}|}{|N_i| \, (|N_i| - 1)}$$

where $N_i$ is the set of neighbours of $i$. Values range from 0 (no neighbours are linked) to 1 (all neighbours form a clique). Higher values indicate that concepts in a participant's network tend to appear in tightly interconnected clusters. We report the average local clustering coefficient across all nodes.

**Modularity:** Modularity Q quantifies how strongly a network decomposes into internally dense, externally sparse communities (Blondel et al., 2008; M. Newman, 2018). We use the Louvain method for community detection.

$$Q = \frac{1}{2m} \sum_{i,j} \left[A_{ij} - \frac{k_i k_j}{2m}\right] \delta(c_i, c_j)$$

where $m = |E|, A_{ij}$ is the adjacency matrix. $k_i$ is the degree of node $i$; $c_i$ is the community assignment of node $i$; and $\delta(c_i, c_j) = 1$ if $i$ and $j$ are in the same community. Higher $Q$ reflects more clearly separated conceptual clusters, which may indicate compartmentalised semantic organisation.

**Degree centrality:** The degree $k_i$ of node $i$ is a count for its number of edges (E. Haim & Stella, 2026).

$$k_i = \sum_j s_{ij}$$

where $s_{ij}$ is the corresponding adjacency matrix entry. We report the mean normalised degree centrality $\underline{DC} = \frac{1}{n}\sum_i \frac{k_i}{n-1}$ across all nodes, reflecting the average connectivity of concepts in the network. Higher values indicate that nodes are, on average, more densely connected.

**Degree assortativity:** Degree assortativity $r$ is the Pearson correlation coefficient of degrees across all connected node pairs (M. Newman, 2003).

$$r = \frac{\sum_{(i,j)\epsilon E} k_i k_j - M^{-1}\left[\sum_{(i,j)\epsilon E} \frac{k_i + k_j}{2}\right]^2}{\sum_{(i,j)\epsilon E} \frac{k_i^2 + k_j^2}{2} - M^{-1}\left[\sum_{(i,j)\epsilon E} \frac{k_i + k_j}{2}\right]^2}$$

where $M = |E|$. Values of $r > 0$ indicate that high-degree nodes tend to connect to other high-degree nodes, while $r < 0$ indicates a structure where hubs connect to peripheral nodes. Assortativity reflects whether semantically central concepts cluster together or serve as bridges to less connected regions of the network.

**PageRank range:** The PageRank value $p_i$ of node $i$ reflects the node's relative importance in the network, derived from the distribution of a random walk that follows edges with probability $1 - \alpha$ and jumps to an arbitrary node with probability $\alpha$ (Griffiths et al., 2007; Page et al., 1999). We use $\alpha = 0.85$. To quantify how unevenly PageRank is distributed across the network, we compute its range:

$$C_{PR} = max(p_i) - min(p_i)$$

Higher $C_{PR}$ indicates that a small number of concepts dominate the flow of activation relative to the rest of the network. Note that because all networks in this study are undirected and unweighted, PageRank reduces to a degree-proportional stationary distribution. $C_{PR}$ therefore serves primarily as a proxy for degree heterogeneity rather than directed influence, and should be interpreted accordingly.

### 2.6 Extracting individual-level measures for a Machine Learning model

Structural and emotional measures from TFMN and BFMN

To identify network-based predictors of high or low performance in the Alternative Uses Task (AUT), we extracted structural features at the individual network level. For textual forma mentis networks (TFMN, including the Woseco variant) and behavioural forma mentis networks (BFMN), this is methodologically straightforward. A TFMN can be constructed from a single narrative because the network is derived from syntactic dependencies within that text. While group-level networks are obtained by merging story-level networks from multiple participants, an individual's TFMN is simply the network built from their own text. Likewise, individual emotion z-scores are computed on the text of just the individual participant (E. Haim, Fischer, et al., 2026).

Similarly, BFMNs are constructed from cue-response associations provided by a participant. Even at the individual level, these networks may contain clusters and closed triads if participants reuse cue words as responses or repeat responses across different cues (E. Haim, Bergh, et al., 2026). For assigning positive, neutral or negative valence to the words produced by individuals, we use the values taken directly from the participant's ratings rather than performing the Wilcoxon rank-sum test (Stella & Zaytseva, 2020). In cases where a word is repeated within a single participant's answers (e.g. writing the cue "biology" also as association to “school” and “sustainability”), we take the mean rating value of all occurrences to assign valence to that word.

Spreading activation values from VF and WSC networks

Unlike TFM and BFM networks, extracting individual network measures from verbal fluency and Woseco networks is not as straightforward. In current semantic network research, networks obtained from verbal fluency data are usually constructed at the group level by aggregating responses across participants and retaining only words produced by at least two individuals within that group. This procedure reduces noise from idiosyncratic responses (Christensen & Kenett, 2023). If one were to construct a semantic fluency network for a single participant using the same method (e.g., the method proposed in the SemNA R package by Christensen and Kenett (2023), the result would be a simple chain of successive responses. Unless the participant repeated earlier responses, there would be no opportunity for clustering or direct links between non-consecutive responses. As a result, the current standard for constructing group-based fluency networks does not produce meaningful individual-level verbal fluency or Woseco networks.

To nevertheless include verbal fluency and Woseco network features as predictors in our machine learning analyses, we adopted an exploratory spreading activation approach. Spreading activation models conceptualise semantic memory as a network in which activation spreads along associative links between concepts (Collins & Loftus, 1975; Siew, 2019). Such models have been used to simulate retrieval dynamics and associative search processes, including the study of creativity and semantic memory structure (Citraro et al., 2026; Kenett et al., 2014).

We implemented spreading activation simulations using the Python library SpreadPy (Citraro et al., 2026). The library has been previously used in studies to predict the difficulty of items in the Remote Associates Test (Citraro et al., 2026) and to model which network or spreading dynamic features predict creativity ratings of short stories (Passaro et al., 2026).

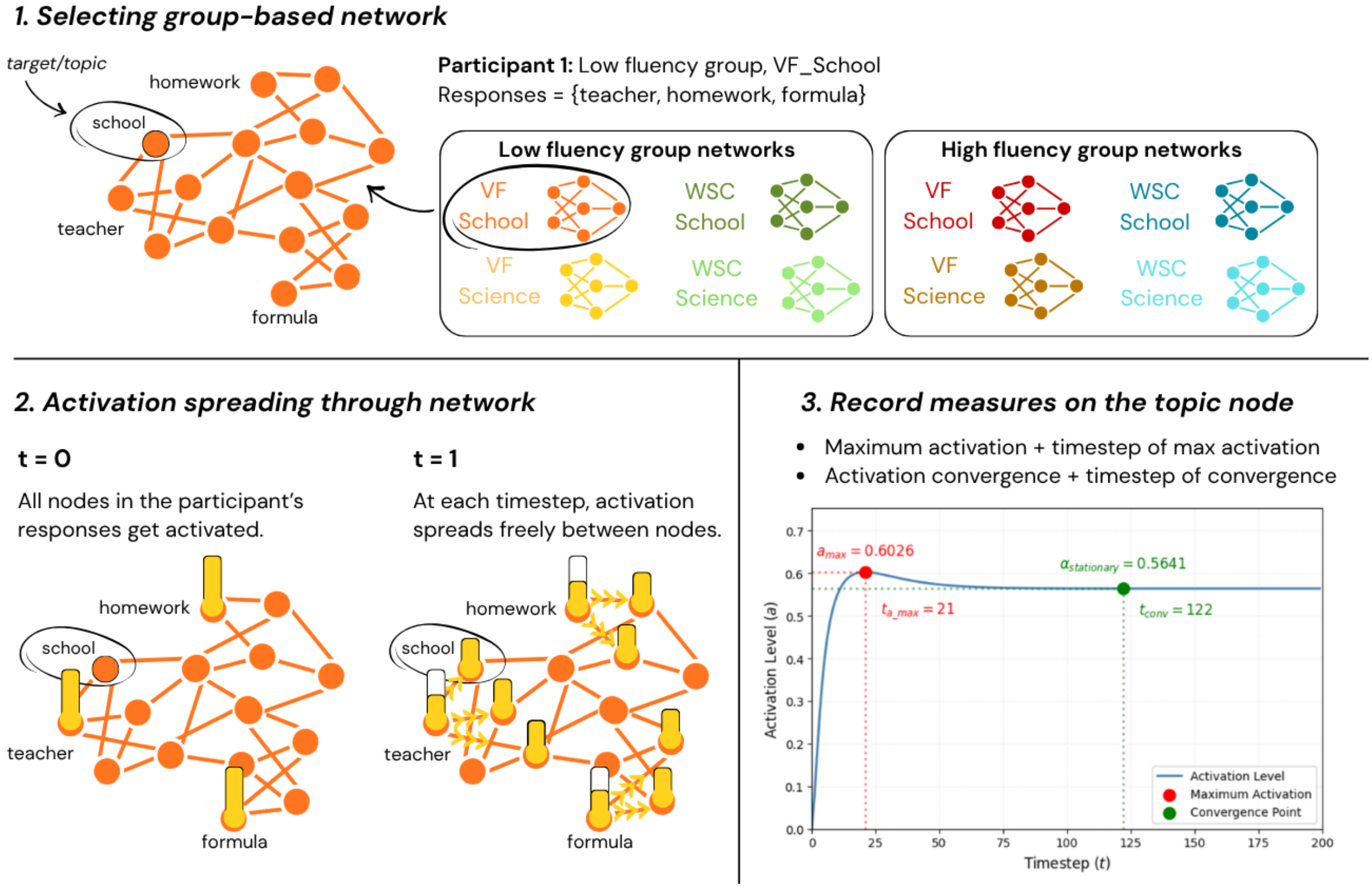

**Figure 2.** Methodological pipeline for extracting spreading activation measures on verbal fluency and Woseco networks for individuals.

Figure 2 outlines the pipeline used to extract spreading activation measures. We performed the activation simulations on the group-level verbal fluency and Woseco networks divided by task and fluency level. To define fluency groups, we calculated a fluency score for each participant as the total number of responses produced across both verbal fluency tasks and both Woseco tasks. By performing a median split, we then assigned participants to a low fluency group or a high fluency group. This was done for each nationality sample separately to compare participants within their own dataset.

It is important to note that these fluency-based networks are distinct from the AUT-based high and low creative networks described in Section 2.1. The AUT-split networks are used exclusively for the descriptive network analyses reported in Section 4.2, where the goal is to characterise structural differences between participants of different creativity levels. For the machine learning analyses, however, constructing VF and WSC networks based on AUT group membership would introduce data leakage: the group-level network structure — and thus the spreading activation features derived from it — would already be informed by the very outcome variable the model is trained to predict (Kaufman et al., 2012). Using fluency-based groups instead ensures that the network features fed into the machine learning model are independent of the AUT scores used as the prediction target. For participants in the low-fluency group we assumed their underlying network to be the low-fluency group networks for either topic, and the same for participants in the high-fluency group. This shall ensure that activation spreads on a semantic structure derived from the associative responses of the participant's own group, thereby approximating their likely underlying semantic organisation. Networks were kept separate by topic ("school", "science") and by task (verbal fluency, Woseco). Only single-layer, undirected, unweighted networks were used.

For a given participant, all nodes corresponding to their responses were activated at the initial time step. Each of these nodes received an initial activation value of $N/n > 1$, where $N$ denotes the total number of nodes in the group-level network and $n$ represents the number of responses produced by the participant. Activation would then spread through the network for 5000 discrete time steps. This number was chosen based on preliminary analyses indicating that in many cases maximum activation and alpha stationary values were reached after approximately 4000 iterations. Running the simulations for 5000 steps ensures that peak activation is captured even in slower-spreading cases. We set the retention parameter to 0.5 meaning that at each time step, nodes retained 50% of their current activation and distributed the remaining 50% equally among their neighbours. A retention rate of 0.5 reflects a balance of retaining and spreading activation, aligning with the idea that activation in semantic memory is neither fully conserved locally nor immediately dissipated globally. Because the networks were unweighted and undirected, activation was distributed evenly across all adjacent nodes. Similar parameter choices have been used in prior spreading activation studies of creativity (Citraro et al., 2026; Passaro et al., 2026). Activation could return to previously activated nodes via neighbouring paths, allowing reinforcement effects that depend on network structure. Consequently, activation trajectories reflect topological properties of the network such as degree, clustering and network size. Over time, activation dynamics approach a stationary regime in which changes between consecutive steps become negligible.

From our spreading activation simulations we extracted four measures for each participant per task and topic. First, we recorded the maximum activation value reached by the target node, i.e. the topic cue word (“school” or “science”). Second, we recorded the time step at which this peak occurred. Third, we calculated the stationary activation level by computing the mean activation over the final stable window once changes in activation fell below a small convergence threshold of $\varepsilon = 10^{-6}$. Stability was assessed across a window of 100 consecutive time steps. Fourth, we recorded the time step at which convergence was reached. Together, these spreading activation measures quantify how strongly and how rapidly a participant’s responses activate the target node within the group-level semantic structure.

## 3. Analysis

In our analyses, we first examined whether the eight task-based network layers encoded distinct associative knowledge structures or whether some layers were structurally redundant and could be merged without meaningful information loss. This was assessed using a multiplex reducibility analysis (Section 3.1). Then we extracted structural, emotional, and spreading activation features from the individual-level networks. These measures were used as features in a machine learning model to predict participants' creativity scores on the Alternative Uses Task (Section 3.2). This allowed us to determine: (i) how the cognitive tasks relate to one another as representations of semantic memory, and (ii) which features of those representations carry the most predictive signal for creative performance.

### 3.1 Multiplex Reducibility Analysis

We first examined whether the different task-based networks within each participant group should be treated as distinct layers of a multiplex system or whether some of them could be aggregated without substantial loss of structural information. Multiplex networks represent systems with multiple layers, in which the same set of nodes is connected by different types of relations (De Domenico et al., 2015; Stella et al., 2024). In our case, each layer corresponded to one task-derived network. Because different tasks and groups produced partly different vocabularies, layers were aligned on the union of all nodes before multiplex reducibility analysis. Words absent from a given layer were represented as isolated nodes in that layer. This alignment allowed all layers to share the same node set while preserving task-specific differences in edge structure.

Each task (verbal fluency, Woseco, behavioural and textual forma mentis) causes a distinct pattern of edges to emerge between these words. Structural reducibility analysis asks whether all these layers are necessary to describe the system or whether some of them encode redundant structural information that can be collapsed into a simpler representation. Reducing layers is desirable when it decreases model complexity without discarding meaningful structural variation (De Domenico et al., 2015).

In our analysis, structural similarity between layers was quantified using the Jensen-Shannon distance (JSD), an information-theoretic measure that captures how different network structures are in their overall connectivity patterns (De Domenico et al., 2015; Lin, 2002). The JSD evaluates global structural differences based on the spectral representation of the networks. Using the *multired framework* (De Domenico et al., 2015), we performed the structural reducibility analysis iteratively over all layers. First, the JSD matrix is computed between all pairs of layers. Layers that are structurally most similar are then merged. After each aggregation step, the entropy of the reduced multiplex representation is recalculated and compared to the entropy of the fully aggregated single-layer network. This process continues until all layers have been collapsed into one. At each stage, a relative entropy value $q$ quantifies how distinguishable the current multilayer configuration is from the fully aggregated network (De Domenico et al., 2015). Higher $q$ values indicate that the retained layer structure preserves more information relative to full aggregation, whereas lower values indicate information loss. The optimal reduced representation is typically identified as the configuration that maximizes $q$, as it balances structural richness against redundancy (De Domenico et al., 2015).

This procedure allows us to determine whether semantic networks derived from different linguistic tasks encode distinct structural information or whether some tasks produce network topologies that may be aggregated. In our analyses, structural reducibility was performed separately for each nationality (AT, US, SG, IT) and participant group (high vs low creative).

### 3.2 Machine Learning Analysis

Following the multiplex reducibility analysis, the resulting optimal layer structure was used as the basis for a machine learning model predicting numeric AUT scores. The aim was to examine which structural properties of the semantic networks are most informative for predicting creative performance. Participants with missing values in any of the features were excluded from the machine learning analysis.

ML feature selection

From the semantic networks derived from the six tasks (VF School, VF Science, WSC School, WSC Science, BFMN, TFMN) we extracted a total of 57 potential features for our machine learning model. These features capture various structural aspects of the networks, including size (number of nodes and edges), connectivity patterns (average shortest path length, diameter, clustering coefficient, modularity), and properties related to spreading activation dynamics and emotional characteristics. Because a number of 57 potential predictors is very large compared to the sample size, we applied a two-stage correlation filter approach to select the most relevant features. This filter approach was guided by the principle that a good feature set should contain predictors that are individually relevant to the outcome while being mutually non-redundant (Hall, 1999; Yu & Liu, 2004). All four national datasets were stacked into a single combined dataset for this step, so that the same feature set would be used consistently across all datasets rather than selecting features separately per country. In the first stage, we computed a correlogram with Spearman rank correlations of all 57 features with the AUT score and identified all features that significantly correlated with AUT scores ($p < 0.05$) with a minimum Spearman correlation coefficient of $\rho > 0.10$. In the second stage, features that were highly intercorrelated with a higher-ranked feature were excluded as redundant. For instance, TFM_Nodes_LCC, TFM_Edges_G, and TFM_Edges_LCC were discarded because they were highly correlated with TFM_Nodes_G.

Because the outcome of the structural reducibility analysis (see Results Section 4.2) indicates that it would be ideal to merge the TFM layer with the WSC-TFM layers for reduced complexity and better interpretability, we tested the correlations for both versions of the TFM layer as either a single layer or a merged network. To ensure a fair comparison between the two configurations and avoid biasing the ML model against either one, we took the union of the relevant features identified across both conditions. Non-redundant features that reached significance in either configuration were retained as candidates, regardless of whether they were significant in both. This two-stage selection approach yielded a final set of 12 features used as input to the machine learning model (see Table 2).

| Category | Feature | Spearman's ρ | p-value |
|---|---|---|---|
| Structural | TFM_Nodes_G | 0.258 | < 0.001 |
| | TFM_CC | -0.131 | < 0.01 |
| | TFM_ASPL | 0.109 | < 0.05 |
| Spreading activation | VF_School_alpha_stationary | 0.250 | < 0.001 |
| | WSC_Science_max_activation | 0.184 | < 0.001 |
| | VF_Science_t_convergence | 0.165 | < 0.001 |
| | WSC_Science_pct_words_valid | -0.137 | < 0.01 |
| | VF_School_max_activation | 0.112 | < 0.05 |
| | VF_School_time_max_activation | 0.102 | < 0.05 |
| Emotional | TFM_anger | -0.122 | < 0.05 |
| | TFM_disgust | -0.115 | < 0.05 |
| | TFM_surprise | 0.116 | < 0.05 |

**Table 2**. Non-redundant features that are highly correlated with the AUT scores in either the TFM single layer or TFM merged condition. These 12 features were selected for the ML model. The features are ranked by their Spearman's rank correlation coefficient and are grouped into structural, emotional and spreading activation features.

Identification of best regressor and explainable AI

After selecting the optimal non-redundant features, we compared three regression models to identify the most predictive approach: Ridge, Random Forest, and XGBoost regression. Predictors were standardised for the linear model using z-score normalisation.

Model performance was evaluated using nested cross-validation with three outer folds and three inner folds, which repeatedly trains and tests the model on different subsets of the data (Varoquaux & Colliot, 2023; Zhou & Kang, 2025). The inner cross-validation loop was used for hyperparameter optimization via grid search, while the outer loop estimated out-of-sample predictive performance.

We assessed the model's performance across the outer folds using the coefficient of determination ($R^2$) across folds to capture model stability. We further recorded root mean squared error (RMSE), mean absolute error (MAE), Pearson correlation and Spearman correlation. For all four datasets stacked together, the Ridge regressor showed the highest predictive performance (see Table 3).

| **Model** | **Mean $R^2$** | **RMSE** | **Pearson_r** | **Spearman_r** |
|---|---|---|---|---|
| Ridge | 0.116 | 0.071 | 0.340 | 0.293 |
| Random Forest | 0.080 | 0.073 | 0.291 | 0.288 |
| XGBoost | 0.006 | 0.075 | 0.248 | 0.253 |

**Table 3**. Cross-validated (3-folds) model performance for the regressors tested across all four datasets using the TFM+WSC TFM merged networks.

To interpret the contribution of individual predictors to the model's overall predictions, we used Explainable Artificial Intelligence methods (XAI). We computed SHAP values (SHapley Additive exPlanations) using the Python SHAP package to estimate the importance of each feature in the prediction process (Lundberg et al., 2020; Lundberg & Lee, 2017). Furthermore, we used SHAP beeswarm plots to identify which network properties had the strongest influence on higher or lower predicted creativity scores. This approach allows otherwise "black box" machine learning models to become interpretable.

# 4. Results

Here we present the results of our analyses. First, we compare AUT originality scores for human participants and AI simulations. Next, we present the network measures calculated for each human and AI dataset, followed by the results of the multiplex reducibility analysis. Finally, we report the best combination of predictive features per dataset and the outcome of the machine learning models predicting high or low values in the AUT task.

## 4.1 Differences in AUT scores

We first examined the distribution of AUT originality scores for human participants and AI-generated responses (see Table 4). The AUT scores were calculated as the mean ratings provided by the two AI "raters" CLAUS (Patterson et al., 2025) and OCSAI (Organisciak et al., 2023).

Human responses had an average score of 0.4417 across both high and low creative groups and showed substantial variability with a total range of 0.3646 between the lowest and highest scores across all four datasets. Mean AUT scores were highest in the Singaporean sample (M = 0.478, SD = 0.069), followed by Austria (M = 0.439, SD = 0.052), the United States (M = 0.43, SD = 0.097), and Italy (M = 0.416, SD = 0.080). In contrast, AI-generated responses were more homogeneous in content and in originality. Although AI responses received a higher average originality score across all answers (0.5135), the variability between them was considerably lower, with a total range of only 0.1166.

| Dataset | Low_avg | Low_range | High_avg | High_range | Total_avg | Total_range |
|---|---|---|---|---|---|---|
| Human_avg | 0.3877 | 0.2379 | 0.4955 | 0.1262 | 0.4417 | 0.3646 |
| AI_avg | 0.4953 | 0.0680 | 0.5316 | 0.0484 | 0.5135 | 0.1166 |

**Table 4**. Average values and range for AUT scores in the entire human and AI datasets.

A similar pattern was observed when separating responses into high- and low-creative groups. Human responses displayed broader ranges within both groups (Low_range = 0.2379; High_range = 0.1262), whereas the AI simulations produced only narrow score differences (Low_range = 0.068; High_range = 0.0484). This indicates that AI-generated responses tend to be homogeneous and cluster closely together in their evaluated creativity. Such homogeneity in AI creative outputs is consistent with previous studies (Anderson et al., 2024; Moon et al., 2025; Wenger & Kenett, 2025). At the same time, it is important to note that originality ratings in the present study were produced by two AI-based scoring systems: CLAUS which is based on a fine-tuned XLM-RoBERTa architecture (Patterson et al., 2025), and OCSAI which is built on a GPT-4o-mini-based supervised model (Organisciak et al., 2023) (see further Section 2.3). Prior work has shown that AI evaluators can display systematic biases in favour of AI-generated content (Domanti et al., 2026; E. Haim, Fischer, et al., 2026) which should be considered when interpreting the higher originality scores of AI outputs in comparison to human responses.

The full table displaying AUT score distribution for each dataset separately can be found online in the supplementary materials (https://osf.io/ns952/overview).

### 4.2 Network Measures and Reducibility Analyses

In this section we report the descriptive network measures for the semantic networks constructed from human and AI-generated response. Because of the large number of networks created, we do not include visual figures for these networks in the manuscript. All network visualisations are available in the supplementary materials on OSF (https://osf.io/ns952/overview). We report the measures for all analyses calculated for the Largest Connected Component (LCC). The size of the full graph and the LCC was very similar across datasets, which is why we only report the sizes for the LCC here.

Instead of using configuration models or bootstrap analyses, we used AI-generated networks as a reference to evaluate whether the human networks exhibit non-random structure. This approach was chosen because the study combines several different network construction procedures, for which bootstrapping would introduce additional methodological complexity. Therefore, the AI networks serve as a comparative baseline to illustrate whether the structural properties observed in human networks deviate from patterns that could arise under more homogeneous responses.

The following subsections present the results for each dataset separately. For each dataset, we first report the descriptive network measures for the human high creative and low creative networks (Tables 5, 8, 11, 14). The network measures for their AI counterparts can be found in the Appendix. Afterwards, we show the results of the structural reducibility analysis. This was performed in two steps: First, the reducibility analysis was conducted on the full dataset to examine whether the structural differences between the networks from high- and low-creative participants reflect meaningful variation or whether these layers could be merged without substantial loss of structural information. This was relevant given the homogeneous patterns observed in the AI-generated responses and AUT originality scores (see Section 4.1).

Second, the reducibility analysis was repeated separately for the high- and low-creative groups to examine which task layers contained redundant or distinct structural information within each group. The results of these analyses determine which layers can be aggregated or should be retained as separate in the subsequent machine learning analyses.

When reporting the best configuration for the reducibility analysis, layers that can be merged are listed together in square brackets. Layers appearing within the same brackets are merged into a single layer (e.g. [L1, L2, L3, L4] merge into one aggregated layer). If several sets of brackets are shown, each set represents a separate merge. For example, the notation

"[L1, L2] and [L3, L4]" indicates that L1 merges with L2, and L3 merges with L4 but the two resulting layers are not aggregated with each other.

Study 1 - Austria

Table 5 summarises the network measures for the Austrian student networks across all tasks and creativity groups. Overall, the networks show typical semantic network properties, including relatively high clustering coefficients in verbal fluency tasks and larger, denser structures for the TFMN-based networks. The corresponding AI-generated networks are reported in the Appendix. Overall, the simulated networks were generally smaller and more homogeneous compared to the human networks.

| Task | Group | Nodes LCC | Edges LCC | Diameter | ASPL | CC | Q | Degree centr | Degree assort | PageRank Range |
|---|---|---|---|---|---|---|---|---|---|---|
| VF School | AT High | 148 | 439 | 10 | 4.477 | 0.704 | 0.748 | 0.040 | -0.050 | 0.021 |
| | AT Low | 144 | 426 | 9 | 4.092 | 0.707 | 0.737 | 0.041 | -0.016 | 0.029 |
| VF Science | AT High | 111 | 327 | 7 | 3.230 | 0.738 | 0.672 | 0.054 | -0.062 | 0.029 |
| | AT Low | 124 | 366 | 10 | 4.613 | 0.697 | 0.726 | 0.048 | 0.008 | 0.016 |
| WSC School | AT High | 73 | 213 | 8 | 3.264 | 0.720 | 0.623 | 0.081 | -0.119 | 0.055 |
| | AT Low | 75 | 219 | 7 | 3.636 | 0.717 | 0.613 | 0.079 | -0.082 | 0.037 |
| WSC Science | AT High | 75 | 222 | 8 | 3.239 | 0.710 | 0.632 | 0.080 | -0.095 | 0.049 |
| | AT Low | 59 | 174 | 6 | 2.843 | 0.731 | 0.584 | 0.102 | -0.119 | 0.057 |
| BFMN | AT High | 207 | 257 | 6 | 3.766 | 0.102 | 0.693 | 0.012 | -0.890 | 0.056 |
| | AT Low | 196 | 247 | 6 | 3.652 | 0.106 | 0.666 | 0.013 | -0.876 | 0.058 |
| TFMN | AT High | 924 | 4012 | 7 | 3.037 | 0.582 | 0.405 | 0.009 | -0.135 | 0.026 |
| | AT Low | 854 | 3429 | 8 | 3.116 | 0.545 | 0.426 | 0.009 | -0.143 | 0.033 |
| WSC-TFMN School | AT High | 675 | 2427 | 9 | 3.245 | 0.539 | 0.479 | 0.011 | -0.098 | 0.021 |
| | AT Low | 660 | 2466 | 8 | 3.201 | 0.515 | 0.462 | 0.011 | -0.101 | 0.022 |
| WSC-TFMN Science | AT High | 639 | 2252 | 10 | 3.456 | 0.521 | 0.466 | 0.011 | -0.072 | 0.021 |
| | AT Low | 644 | 2353 | 7 | 3.287 | 0.547 | 0.497 | 0.011 | -0.060 | 0.018 |

| Total Merged | AT High | 1879 | 9571 | 10 | 3.160 | 0.516 | 0.368 | 0.005 | -0.122 | 0.016 |
|---|---|---|---|---|---|---|---|---|---|---|
| | AT Low | 1789 | 9101 | 8 | 3.124 | 0.511 | 0.365 | 0.006 | -0.122 | 0.014 |

**Table 5**. Descriptive network measures for Austrian student semantic networks across tasks and creativity groups. Measures are reported for the largest connected component (LCC).

Figure 3 shows the multiplex reducibility analysis for the full Austrian dataset including both creativity groups. The corresponding optimal configurations and q-values are summarized in Table 6. The optimal configuration indicates merging of VF and WSC for the topic "science", as well as merging TFM layers (TFM and WSC-TFMs) within the low and high creative group each for the human dataset. For the AI dataset, only the high and low creative layers of WSC-TFM for the topic "science" can be merged. Notice how the human network only allows merging within groups (high-high; low-low) while the AI layers can also be merged across high and low creative groups. Figure 4 and Table 7 show the reducibility analyses conducted separately for the high- and low-creative groups.

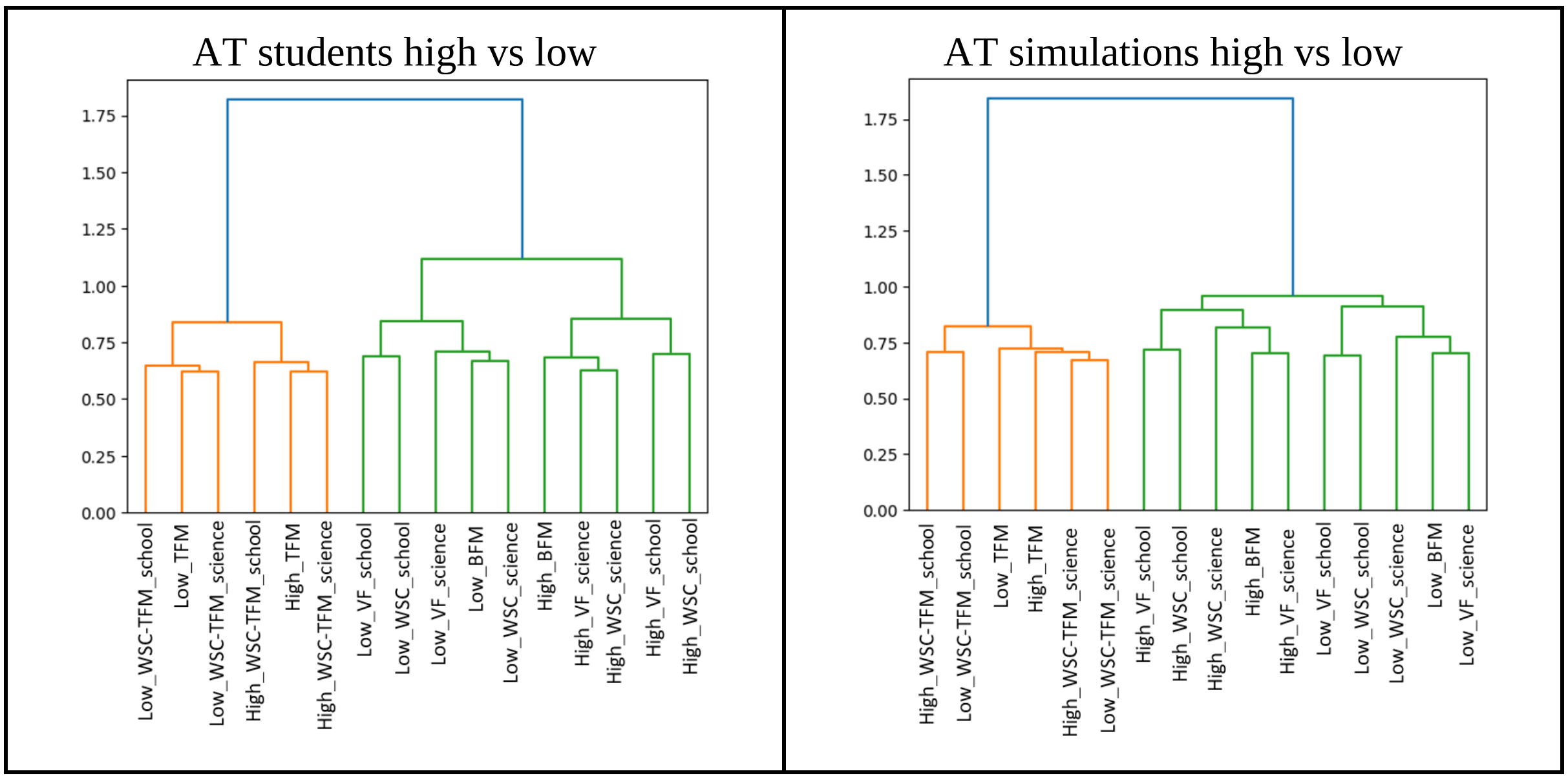

**Figure 3**. Multiplex reducibility analysis for the Austrian dataset including both creativity groups for human (left) and AI-generated (right) networks.

| Group | Optimal Layer Configuration | q-value |
|---|---|---|
| AT students high vs. low | 11 layers total, merging [High_VF_science, High_WSC_science] and [Low_TFM, Low_WSC-TFM_school, Low_WSC-TFM_science] and [High_TFM, High_WSC-TFM_school, High_WSC-TFM_science] | 0.311 |
| AT simulation high vs. low | 15 layers total, merging [High_WSC-TFM_science, Low_WSC-TFM_science] | 0.342 |

**Table 6**. Optimal multiplex layer configuration and q-values for the reducibility analysis of the Austrian dataset including both creativity groups.

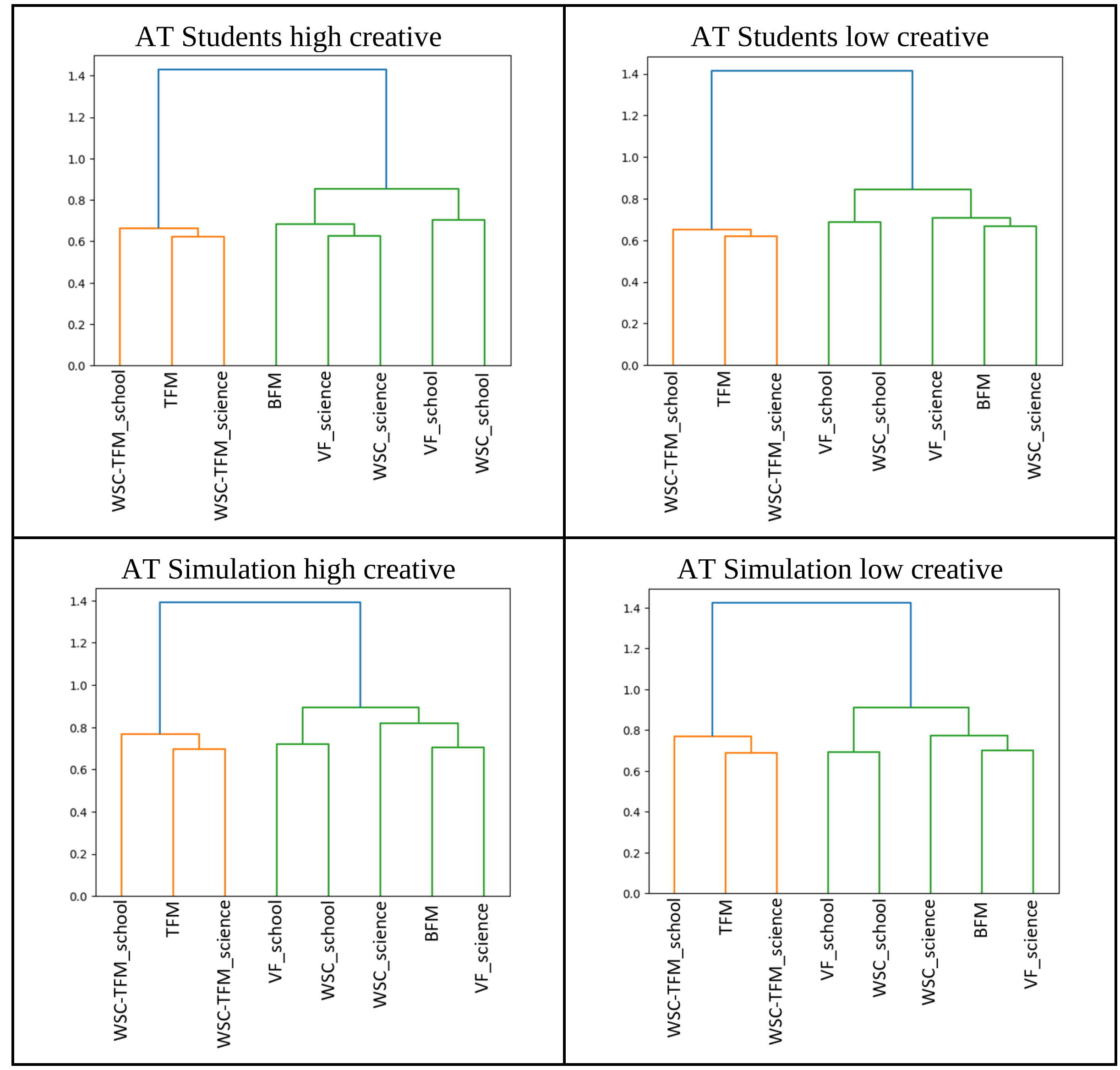

**Figure 4**. Multiplex reducibility analysis conducted separately for high (left) and low (right) creativity groups in the Austrian dataset for human (top) and AI-generated (bottom) networks.

| Group | Optimal Layer Configuration | q-value |
|---|---|---|
| AT Students high | 7 layers total, merging [TFM, WSC-TFM_science] | 0.269 |
| AT Students low | 6 layers total, merging [TFM, WSC-TFM_school, WSC-TFM_science] | 0.289 |
| AT Simulations high | 7 layers total, merging [TFM, WSC-TFM_science] | 0.309 |
| AT Simulations low | 7 layers total, merging [TFM, WSC-TFM_science] | 0.313 |

**Table 7**. Optimal multiplex layer configurations and q-values for reducibility analyses conducted separately for high- and low-creativity groups in the Austrian dataset.

Study 2 - USA

Table 8 summarizes the descriptive network measures for the U.S. student networks across tasks and creativity groups. Similar to the Austrian dataset, networks derived from the TFMN tasks were substantially larger and denser than the verbal fluency or word association networks. AI-generated networks for the same tasks are reported in the Appendix. AI networks again show more homogeneous structures compared to the human networks.

| Task | Group | Nodes LCC | Edges LCC | Diameter | ASPL | CC | Q | Degree centr | Degree assort | PageRank Range |
|---|---|---|---|---|---|---|---|---|---|---|
| VF School | US High | 156 | 462 | 9 | 4.108 | 0.725 | 0.734 | 0.038 | -0.077 | 0.030 |
| | US Low | 133 | 393 | 8 | 3.509 | 0.732 | 0.681 | 0.045 | -0.096 | 0.038 |
| VF Science | US High | 129 | 381 | 8 | 4.023 | 0.726 | 0.701 | 0.046 | -0.077 | 0.022 |
| | US Low | 119 | 351 | 8 | 3.875 | 0.719 | 0.673 | 0.050 | -0.029 | 0.024 |
| WSC School | US High | 97 | 285 | 8 | 3.806 | 0.715 | 0.681 | 0.061 | 0.000 | 0.022 |
| | US Low | 54 | 156 | 6 | 3.084 | 0.708 | 0.596 | 0.109 | -0.093 | 0.038 |
| WSC Science | US High | 68 | 198 | 6 | 3.131 | 0.707 | 0.640 | 0.087 | -0.053 | 0.047 |
| | US Low | 53 | 153 | 8 | 3.461 | 0.703 | 0.589 | 0.111 | -0.131 | 0.035 |
| BFMN | US High | 230 | 308 | 5 | 3.426 | 0.161 | 0.660 | 0.012 | -0.882 | 0.058 |
| | US Low | 228 | 309 | 5 | 3.545 | 0.145 | 0.657 | 0.012 | -0.877 | 0.063 |
| TFMN | US High | 846 | 7509 | 6 | 2.575 | 0.535 | 0.272 | 0.021 | -0.155 | 0.016 |
| | US Low | 606 | 5295 | 6 | 2.542 | 0.561 | 0.269 | 0.029 | -0.148 | 0.017 |
| WSC-TFMN School | US High | 485 | 4019 | 6 | 2.532 | 0.499 | 0.304 | 0.034 | -0.107 | 0.019 |
| | US Low | 340 | 2264 | 6 | 2.632 | 0.516 | 0.354 | 0.039 | -0.081 | 0.022 |
| WSC-TFMN Science | US High | 521 | 3335 | 7 | 2.862 | 0.493 | 0.372 | 0.025 | -0.022 | 0.020 |
| | US Low | 299 | 1862 | 7 | 2.668 | 0.548 | 0.401 | 0.042 | -0.100 | 0.022 |
| Total Merged | US High | 1497 | 14894 | 6 | 2.740 | 0.495 | 0.282 | 0.013 | -0.124 | 0.012 |
| | US Low | 1116 | 9933 | 6 | 2.778 | 0.501 | 0.310 | 0.016 | -0.117 | 0.014 |

**Table 8**. Descriptive network measures for US student semantic networks across tasks and creativity groups. Measures are reported for the largest connected component (LCC).

The reducibility analysis again shows that human networks allow merging of task layers primarily within creativity groups. Specifically, VF and WSC layers merge for the high group (topic school), and TFM-related layers merge within each group separately. As in the Austrian dataset, human layers are not merged across high and low groups. In contrast, the AI dataset again allows merging across creativity groups among the TFM-related layers. This pattern is consistent with the relatively homogeneous response structures observed in the simulated data and the narrow range of AUT originality scores.

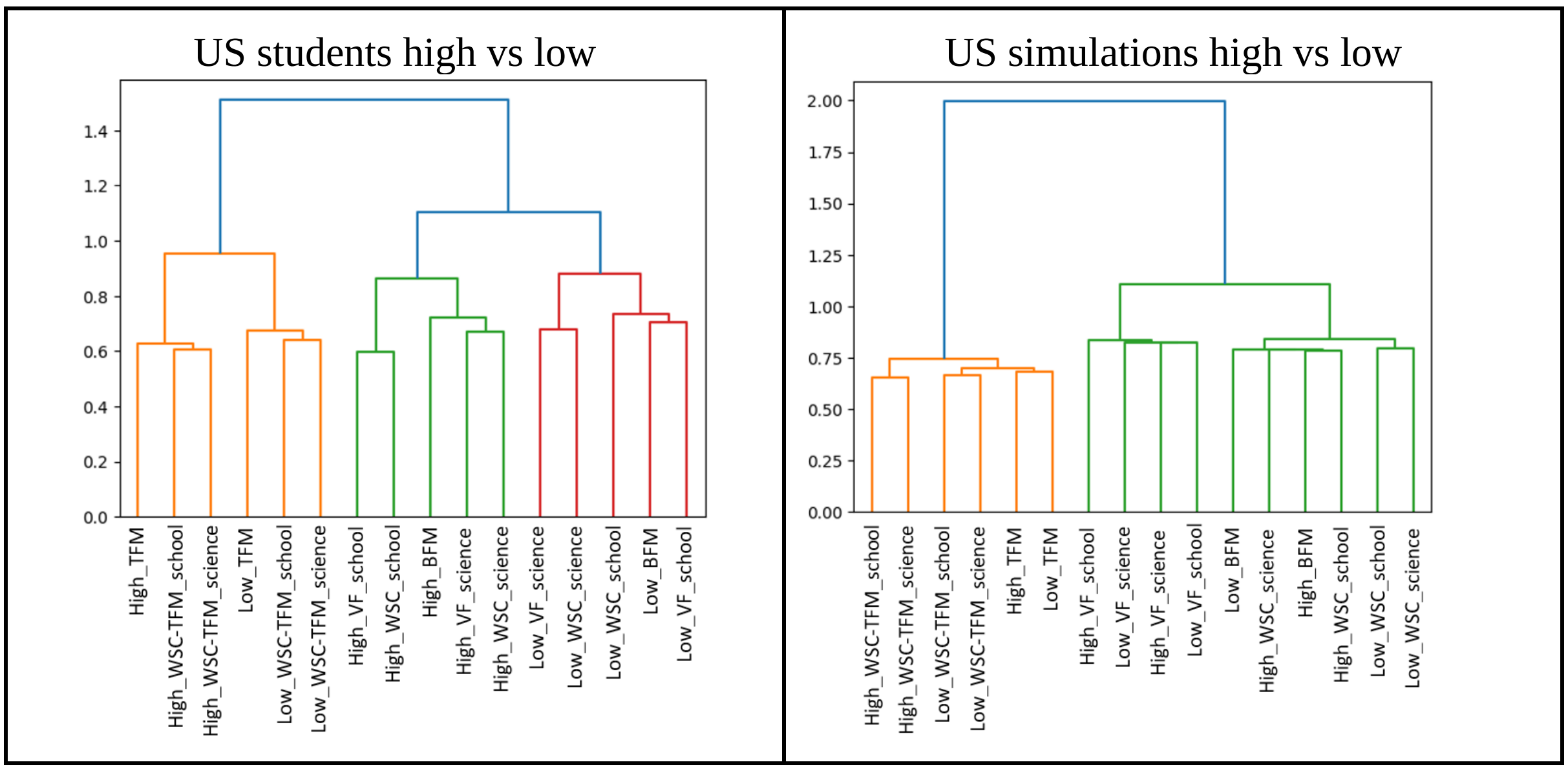

**Figure 5**. Multiplex reducibility analysis for the US dataset including both creativity groups for human (left) and AI-generated (right) networks.

| Group | Optimal Layer Configuration | q-value |
|---|---|---|
| US students high vs. low | 12 layers total, merging [High_VF_school, High_WSC_school] and [High_TFM, High_WSC-TFM_school, High_WSC-TFM_science] and [Low_WSC-TFM_school, Low_WSC-TFM_science] | 0.296 |
| US simulation high vs. low | 11 layers total, merging [High_WSC-TFM_school, High_WSC-TFM_science, Low_WSC-TFM_school, Low_WSC-TFM_science, High_TFM, Low_TFM] | 0.404 |

**Table 9**. Optimal multiplex layer configuration and q-values for the reducibility analysis of the US dataset including both creativity groups.

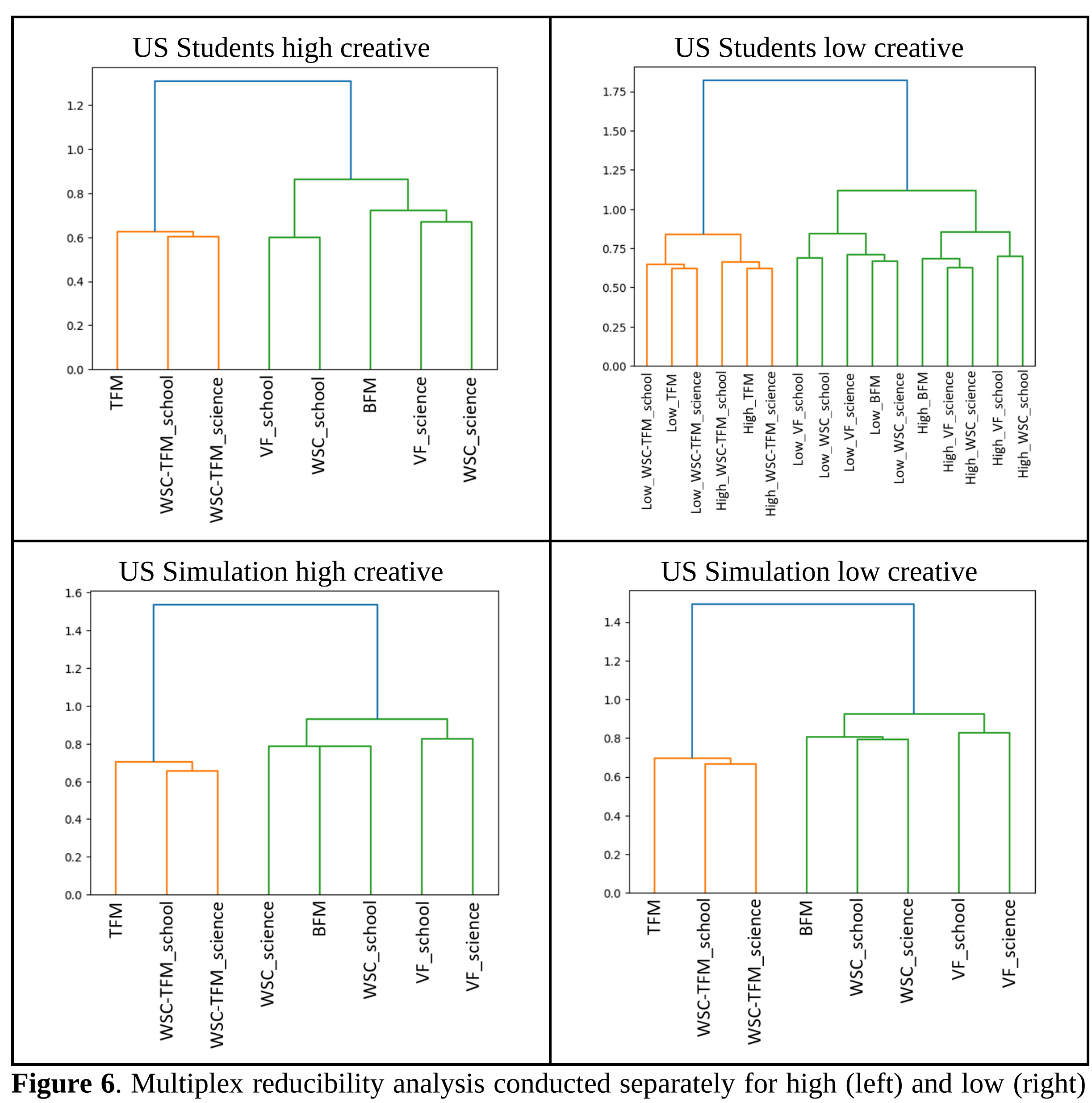

**Figure 6**. Multiplex reducibility analysis conducted separately for high (left) and low (right) creativity groups in the US dataset for human (top) and AI-generated (bottom) networks.

| Group | Optimal Layer Configuration | q-value |
|---|---|---|
| US Students high | 5 layers total, merging [VF_school, WSC_school] and [TFM, WSC-TFM_school, WSC-TFM_science] | 0.250 |
| US Students low | 6 layers total, merging [TFM, WSC-TFM_school, WSC-TFM_science] | 0.275 |
| US Simulations high | 6 layers total, merging [TFM, WSC-TFM_school, WSC-TFM_science] | 0.356 |
| US Simulations low | 6 layers total, merging [TFM, WSC-TFM_school, WSC-TFM_science] | 0.347 |

**Table 10**. Optimal multiplex layer configurations and q-values for reducibility analyses conducted separately for high- and low-creativity groups in the US dataset.

Study 3 - Singapore

Descriptive network measures for the Singapore dataset are shown in Table 11. Overall structural properties were comparable to the previous datasets.

| Task | Group | Nodes LCC | Edges LCC | Diameter | ASPL | CC | Q | Degree centr | Degree assort | PageRank Range |
|---|---|---|---|---|---|---|---|---|---|---|
| VF School | SG High | 145 | 432 | 8 | 3.650 | 0.723 | 0.689 | 0.041 | -0.107 | 0.035 |
| | SG Low | 145 | 435 | 12 | 4.694 | 0.716 | 0.726 | 0.042 | -0.125 | 0.025 |
| VF Science | SG High | 140 | 414 | 12 | 5.654 | 0.697 | 0.730 | 0.043 | -0.052 | 0.018 |
| | SG Low | 124 | 366 | 9 | 4.358 | 0.703 | 0.724 | 0.048 | 0.028 | 0.020 |
| WSC School | SG High | 85 | 249 | 8 | 3.654 | 0.705 | 0.657 | 0.070 | 0.022 | 0.028 |
| | SG Low | 70 | 204 | 7 | 3.254 | 0.724 | 0.620 | 0.084 | -0.079 | 0.036 |
| WSC Science | SG High | 80 | 234 | 8 | 3.318 | 0.727 | 0.602 | 0.074 | -0.093 | 0.040 |
| | SG Low | 78 | 228 | 9 | 4.090 | 0.703 | 0.657 | 0.076 | -0.137 | 0.029 |
| BFMN | SG High | 216 | 287 | 6 | 3.464 | 0.132 | 0.653 | 0.012 | -0.870 | 0.064 |
| | SG Low | 243 | 313 | 5 | 3.530 | 0.151 | 0.687 | 0.011 | -0.903 | 0.055 |
| TFMN | SG High | 927 | 8634 | 6 | 2.628 | 0.512 | 0.274 | 0.020 | -0.116 | 0.017 |
| | SG Low | 864 | 8185 | 6 | 2.603 | 0.515 | 0.271 | 0.022 | -0.124 | 0.016 |
| WSC-TFMN School | SG High | 611 | 4111 | 7 | 2.794 | 0.515 | 0.329 | 0.022 | -0.084 | 0.019 |
| | SG Low | 524 | 3732 | 6 | 2.640 | 0.544 | 0.330 | 0.027 | -0.134 | 0.019 |
| WSC-TFMN Science | SG High | 643 | 4632 | 6 | 2.753 | 0.536 | 0.336 | 0.022 | -0.077 | 0.017 |
| | SG Low | 532 | 4000 | 6 | 2.698 | 0.536 | 0.335 | 0.028 | -0.073 | 0.018 |
| Total Merged | SG High | 1762 | 17279 | 6 | 2.813 | 0.502 | 0.278 | 0.011 | -0.110 | 0.010 |
| | SG Low | 1568 | 15817 | 7 | 2.762 | 0.499 | 0.285 | 0.013 | -0.117 | 0.010 |

**Table 11**. Descriptive network measures for Singaporean student semantic networks across tasks and creativity groups. Measures are reported for the largest connected component (LCC).

The reducibility results follow the same pattern observed in Austria and the United States. In the human dataset, VF and WSC layers merge for the high group (topic school), while TFM-related layers merge within creativity groups but not across them. In contrast, the AI dataset again allows merging across high and low creativity groups, indicating limited structural differentiation between the simulated groups.

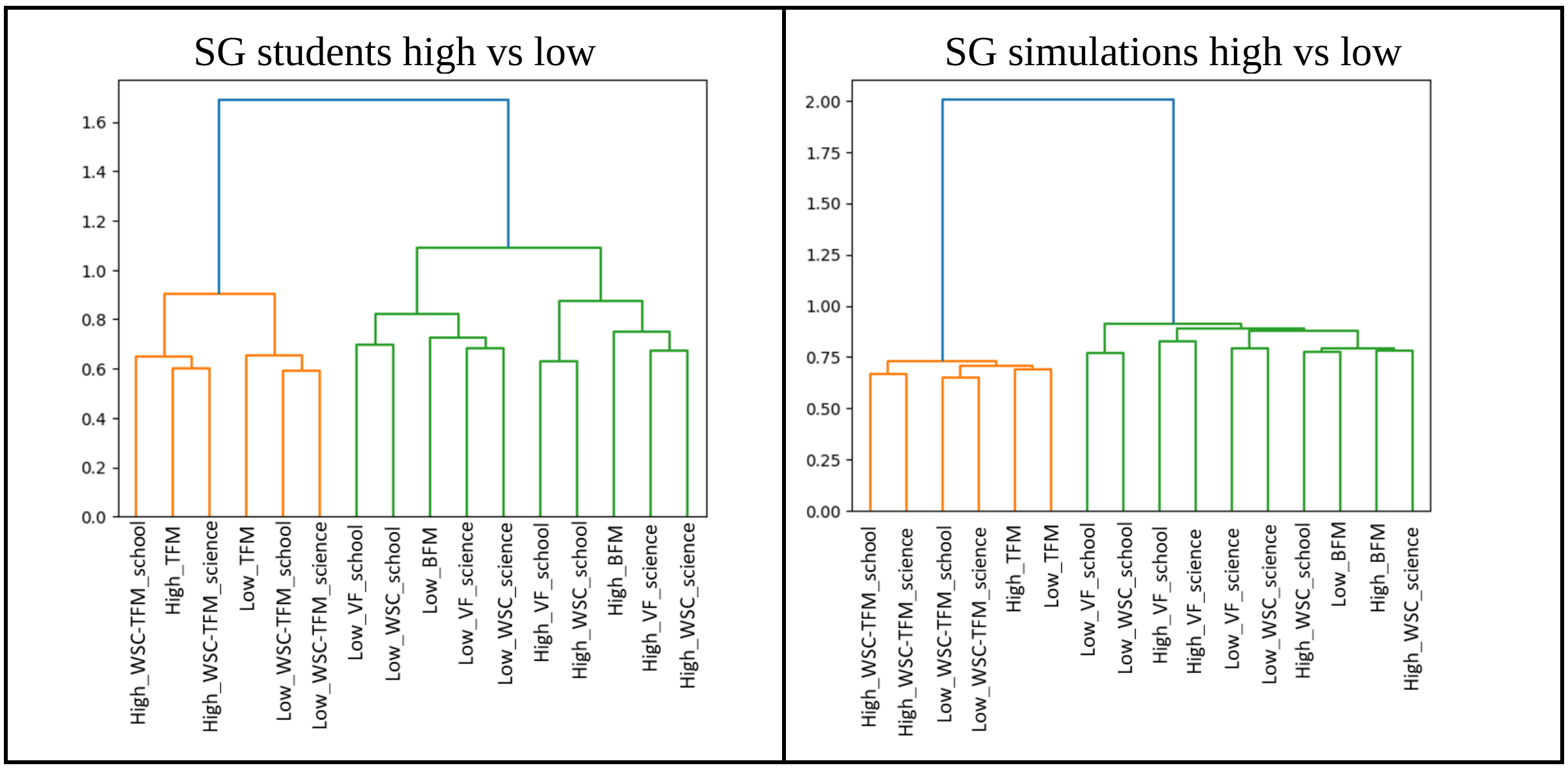

**Figure 7**. Multiplex reducibility analysis for the Singaporean dataset including both creativity groups for human (left) and AI-generated (right) networks.

| Group | Optimal Layer Configuration | q-value |
|---|---|---|
| SG students high vs. low | 11 layers total, merging [High_VF_school, High_WSC_school] and [High_TFM, High_WSC-TFM_school, High_WSC-TFM_science] and [Low_TFM, Low_WSC-TFM_school, Low_WSC-TFM_science] | 0.304 |

| SG simulation high vs. low | 11 layers total, merging [High_WSC-TFM_school, High_WSC-TFM_science, Low_WSC-TFM_school, Low_WSC-TFM_science, High_TFM, Low_TFM] | 0.407 |
|---|---|---|

**Table 12**. Optimal multiplex layer configuration and q-values for the reducibility analysis of the Singaporean dataset including both creativity groups.

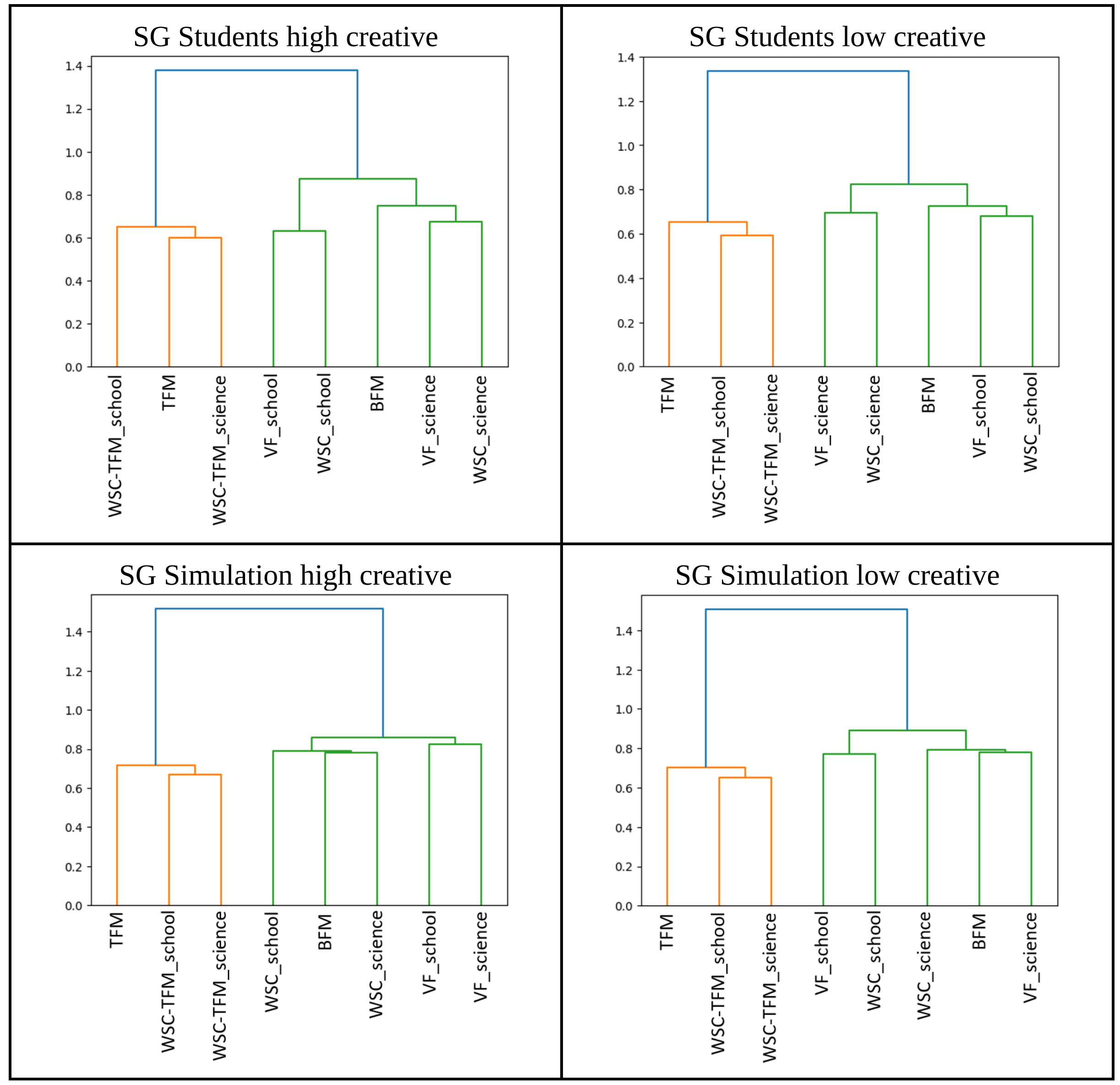

**Figure 8**. Multiplex reducibility analysis conducted separately for high (left) and low (right) creativity groups in the Singaporean dataset for human (top) and AI-generated (bottom) networks.

| Group | Optimal Layer Configuration | q-value |
|---|---|---|
| SG Students high | 7 layers total, merging [TFM, WSC-TFM_science] | 0.259 |
| SG Students low | 6 layers total, merging [TFM, WSC-TFM_school, WSC-TFM_science] | 0.276 |
| SG Simulations high | 6 layers total, merging [TFM, WSC-TFM_school, WSC-TFM_science] | 0.360 |
| SG Simulations low | 6 layers total, merging [TFM, WSC-TFM_school, WSC-TFM_science] | 0.350 |

**Table 13**. Optimal multiplex layer configurations and q-values for reducibility analyses conducted separately for high- and low-creativity groups in the Singaporean dataset.

Study 4 - Italy

Table 14 reports the descriptive network measures for the Italian dataset.

| Task | Group | Nodes LCC | Edges LCC | Diameter | ASPL | CC | Q | Degree centr | Degree assort | PageRank Range |
|---|---|---|---|---|---|---|---|---|---|---|
| VF School | IT High | 89 | 267 | 9 | 3.839 | 0.680 | 0.677 | 0.068 | 0.011 | 0.019 |
| | IT Low | 79 | 234 | 6 | 2.969 | 0.727 | 0.616 | 0.076 | -0.083 | 0.033 |
| VF Science | IT High | 67 | 201 | 7 | 3.269 | 0.702 | 0.597 | 0.091 | -0.048 | 0.037 |
| | IT Low | 61 | 185 | 5 | 2.751 | 0.710 | 0.522 | 0.101 | -0.012 | 0.041 |
| WSC School | IT High | 55 | 159 | 7 | 3.428 | 0.705 | 0.605 | 0.107 | -0.050 | 0.031 |
| | IT Low | 59 | 171 | 6 | 2.929 | 0.715 | 0.571 | 0.100 | 0.023 | 0.042 |
| WSC Science | IT High | 44 | 126 | 6 | 2.820 | 0.711 | 0.551 | 0.133 | -0.103 | 0.055 |
| | IT Low | 54 | 156 | 7 | 3.214 | 0.707 | 0.578 | 0.109 | -0.085 | 0.035 |
| BFMN | IT High | 160 | 185 | 9 | 4.415 | 0.040 | 0.724 | 0.015 | -0.928 | 0.054 |
| | IT Low | 156 | 180 | 8 | 4.359 | 0.048 | 0.729 | 0.015 | -0.923 | 0.056 |
| TFMN | IT High | 532 | 4166 | 7 | 2.589 | 0.577 | 0.331 | 0.029 | -0.096 | 0.024 |
| | IT Low | 493 | 3900 | 5 | 2.530 | 0.588 | 0.319 | 0.032 | -0.102 | 0.024 |
| WSC-TFMN School | IT High | 319 | 6224 | 4 | 2.090 | 0.732 | 0.280 | 0.123 | -0.201 | 0.015 |
| | IT Low | 320 | 12195 | 4 | 1.865 | 0.758 | 0.240 | 0.239 | -0.213 | 0.009 |
| WSC-TFMN Science | IT High | 332 | 4486 | 5 | 2.228 | 0.684 | 0.333 | 0.082 | -0.133 | 0.020 |
| | IT Low | 309 | 15059 | 4 | 1.720 | 0.843 | 0.033 | 0.319 | -0.501 | 0.012 |

| Total Merged | IT High | 1085 | 14946 | 6 | 2.624 | 0.589 | 0.359 | 0.025 | -0.107 | 0.010 |
|---|---|---|---|---|---|---|---|---|---|---|
| | IT Low | 1021 | 30259 | 5 | 2.410 | 0.656 | 0.320 | 0.058 | -0.171 | 0.007 |

**Table 14**. Descriptive network measures for Italian student semantic networks across tasks and creativity groups. Measures are reported for the largest connected component (LCC).

Regarding the network measures we want to point out the unexpectedly high edge count of the WSC-TFM networks for the low creative group. Even though low and high creative groups showed comparable node counts (~309–332 nodes per group), low creative students showed considerably higher edge counts and degree centrality compared to high creative students in both topics "school" (low: 12195 edges, degree centrality = 0.239; high: 6224 edges, degree centrality = 0.123) and "science" (low: 15059 edges, degree centrality = 0.319; high: 4486 edges, degree centrality = 0.082). This effect could be partially explained by the fact that we used the TFM method (intended for coherent texts) to WSC responses (sentence-chains repeating target words) and mixed two separately designed tasks together. The Woseco task forces repetition of words at sentence boundaries to create a chain of sentences. Aggregating these into a single corpus produces a small set of highly repeated hub words that densely interconnect the network. Our exploratory group-level WSC-TFMNs should therefore be interpreted with caution, as network density in this case may reflect task constraints rather than meaningful differences in semantic structure.

The reducibility analysis largely replicates the pattern observed in the other datasets, with TFM-related layers merging within creativity groups. However, the Italian dataset represents the only case in which the reducibility analysis allows for the human dataset to merge across high and low creative groups: the best configuration includes merging the high-creative TFM layer with the low-creative WSC-TFM school layer. The AI dataset again shows broader merging across layers and merges between high and low creative groups.

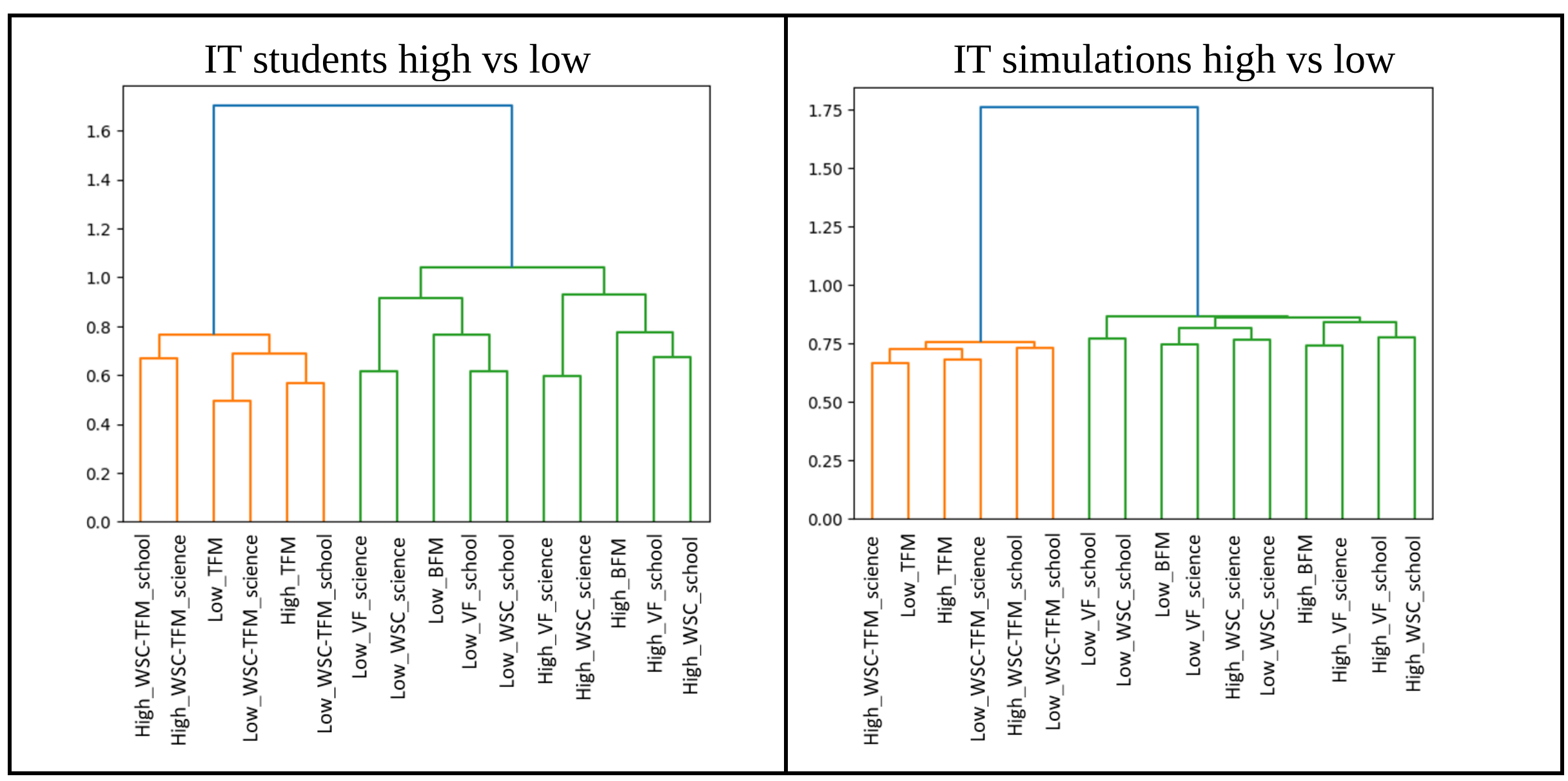

**Figure 9**. Multiplex reducibility analysis for the Italian dataset including both creativity groups for human (left) and AI-generated (right) networks.

| Group | Optimal Layer Configuration | q-value |
|---|---|---|
| IT students high vs. low | 14 layers total, merging [Low_TFM, Low_WSC-TFM_science] and [High_TFM, Low_WSC-TFM_school] | 0.314 |
| IT simulation high vs. low | 12 layers total, merging [High_WSC-TFM_science, Low_TFM, High_TFM, Low_WSC-TFM_science] and [High_WSC-TFM_school, Low_WSC-TFM_school] | 0.377 |

**Table 15**. Optimal multiplex layer configuration and q-values for the reducibility analysis of the Italian dataset including both creativity groups.

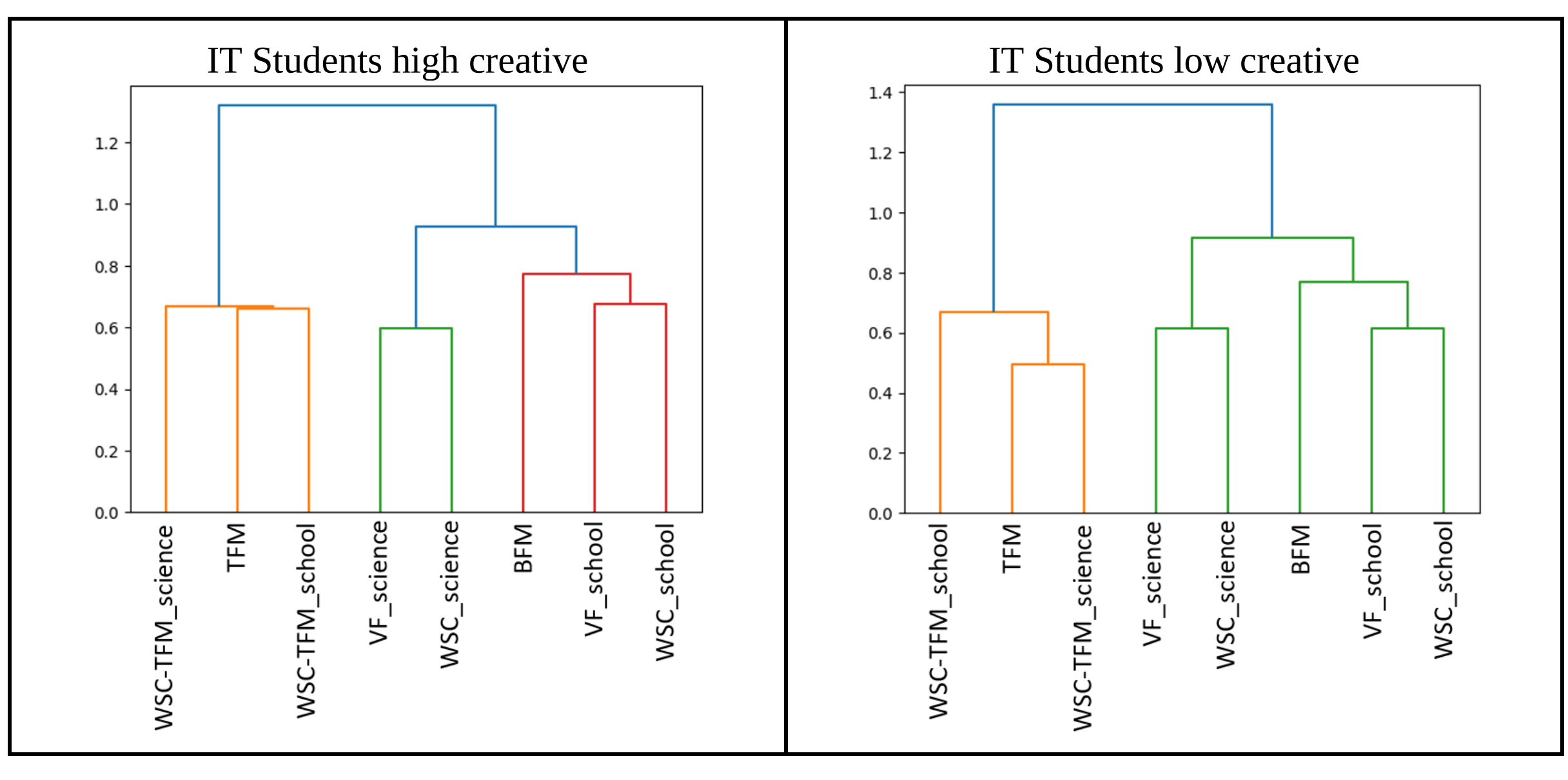

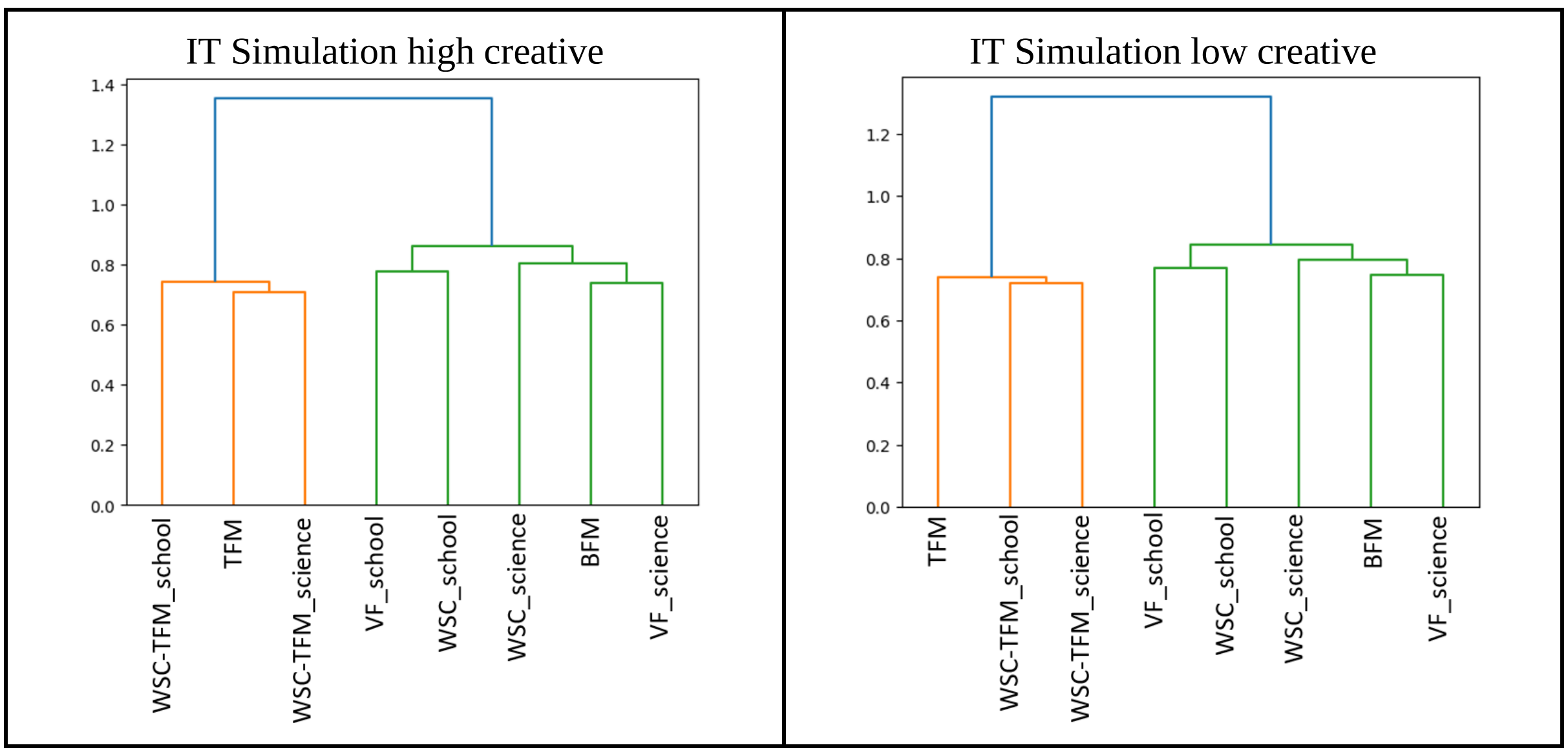

**Figure 10**. Multiplex reducibility analysis conducted separately for high (left) and low (right) creativity groups in the Italian dataset for human (top) and AI-generated (bottom) networks.

| Group | Optimal Layer Configuration | q-value |
|---|---|---|
| IT Students high | 5 layers total, merging [High_VF_science, High_WSC_science] and [TFM, WSC-TFM_school, WSC-TFM_science] | 0.276 |
| IT Students low | 7 layers total, merging [TFM, WSC-TFM_science] | 0.277 |
| IT Simulations high | 7 layers total, merging [TFM, WSC-TFM_science] | 0.334 |
| IT Simulations low | 6 layers total, merging [TFM, WSC-TFM_school, WSC-TFM_science] | 0.345 |

**Table 16**. Optimal multiplex layer configurations and q-values for reducibility analyses conducted separately for high- and low-creativity groups in the Italian dataset.

Best Structural Reducibility Configuration

Across datasets, the layers most frequently merged in the reducibility analyses were the TFM and WSC TFM networks. These layers are constructed from sentence-based text data in contrast to word fluency lists, and contain syntactic-semantic information. When only one WSC TFM topic was aggregated in the best configuration, it was typically the “science” topic, consistent with the focus of the TFM prompt on elaborating on the relationship between creativity and science. Based on this, we merged the TFM layer with both WSC-TFM topic layers (school and science) for each dataset and creativity group, creating a combined TFM+WSC TFM layer. To explore whether this structural reducibility result actually translates into better predictive performance, we ran two versions of the machine learning model: one using the merged TFM+WSC TFM layer and one using the TFM single layer. This lets us compare whether collapsing the layers as suggested by the reducibility analysis improves predictive performance in a ML model.

Aggregated multiplex lexical network structure

To examine the overall structure of the semantic networks, we constructed aggregated multiplex networks by merging all task layers within each dataset and group. The resulting network measures are reported in the Appendix.

Across all datasets, the aggregated human networks were larger and denser than their AI counterparts, with consistently negative degree assortativity and moderate clustering coefficients. Because these networks combine all tasks into a single structure, task-specific differences are no longer preserved. Consequently, the aggregated networks provide only a general overview of the global semantic structure, while distinctions between individual tasks are largely lost. For this reason, no reducibility analysis was conducted for these networks, as the layers were already fully merged.

### 4.3 Differences in networks from human vs AI participants

Comparisons between high- and low-creative human groups did not reveal consistent patterns in the network measures across datasets. Because the networks were derived from tasks with different response formats and construction procedures (e.g. continuous association vs. text-based networks), these measures are not directly comparable across all network types.

Therefore, we focus on comparing network structure from human responses and AI-simulations instead in this section.

Across all datasets, human networks were consistently larger than AI-generated networks, both in the number of nodes and edges. On average across all layers and datasets, human networks contained 420 nodes and 3517 edges in the largest connected component (LCC), whereas AI networks contained 310 nodes and 2055 edges on average. This reflects a greater diversity of responses in the human datasets, compared to the more homogeneous simulated responses. For example, in the Singaporean dataset, high creative participants found an average of 145 non-idiosyncratic concepts for the "school" verbal fluency task, whereas the corresponding AI simulations produced only 28 distinct concepts (see the Appendix). Consistent with their larger size, human networks also showed slightly larger diameters (diameter_human = 7; diameter_AI = 6), and longer average shortest path lengths (ASPL_human = 3.238; ASPL_AI = 2.850). Human networks also exhibited slightly higher clustering coefficients (CC_human = 0.579) than AI networks (CC_AI = 0.544) and slightly higher modularity (Q_human = 0.513; Q_AI = 0.457). Notably, clustering coefficients in the human BFMNs substantially exceeded those of configuration models preserving the same degree sequence (1000 iterations, $p < 0.001$), with the exception of the Italian dataset where the difference was not significant ($p > 0.3$). AI-generated networks showed no such pattern, with empirical clustering coefficients falling close to or below random expectations across all datasets (all $p > 0.5$). This indicates that the clustering structure in human networks, but not in AI-generated networks, is non-random and reflects genuine associative organisation rather than a byproduct of network size or degree distribution.

Furthermore, AI networks showed higher degree centralization (degree_centr_human = 0.053; degree_centr_AI = 0.072) and a larger PageRank range (PageRank_range_human =

0.029; PageRank_range_AI = 0.046), indicating stronger dominance of a smaller set of highly central concepts. Degree assortativity was negative in both cases (human: –0.183, AI: –0.205).

The reducibility analyses also revealed systematic differences between human and AI-generated networks. In the human datasets, layers belonging to high- and low-creative groups were generally not merged across groups (with only one exception in the Italian dataset) but only within groups (aggregating high-high and low-low layers each). In contrast, the AI datasets frequently allowed merging across creativity groups, indicating limited structural differentiation between simulated high- and low-creative networks.

### 4.4 Correlations of Spreading Activation Dynamics with AUT scores

Since spreading activation on group-level fluency networks was an exploratory approach in this study, we first examined how the spreading dynamics correlated with AUT scores before using them as features in the machine learning model. However, before looking at the correlations, it is important to point out the differences in the underlying network structure between the high and low fluency groups.

As shown in Table 17, the high-fluency networks were substantially larger than the low-fluency networks across both tasks. Averaged across topics and nationality samples, the high-fluency VF networks contained around 145 nodes and 428 edges, compared to 89 nodes and 262 edges for the low-fluency group. A similar pattern is observable for the WSC networks (high vs low fluent: 92 vs 50 nodes; 270 vs 143 edges). These differences in size are expected, based on the fact that groups were split based on the number of responses a participant produced, naturally resulting in the high fluency group containing more responses than the low fluency group. The networks also differed in their overall network topology, with the highly fluent group showing longer distances (ASPL and diameter) than the lowly fluent group, higher clustering coefficients and higher modularity (see Table 17). A full table of network measures for all 32 networks (4 nationalities x 4 tasks x 2 fluency groups) is provided in the Appendix.

| Task | Fluency Group | Nodes LCC | Edges LCC | Diameter | ASPL | CC | Q | Avg Degree | Degree centr | Small Worldness |
|---|---|---|---|---|---|---|---|---|---|---|
| VF | High | 145 | 428 | 9.625 | 4.289 | 0.294 | 0.725 | 5.912 | 0.043 | 5.489 |
| | Low | 89 | 262 | 7.875 | 3.867 | 0.272 | 0.682 | 5.864 | 0.067 | 3.216 |
| WSC | High | 92 | 270 | 9.125 | 3.494 | 0.271 | 0.646 | 5.847 | 0.075 | 3.025 |
| | Low | 50 | 143 | 5.625 | 2.702 | 0.234 | 0.574 | 5.739 | 0.128 | 2.059 |

**Table 17.** Network measures for verbal fluency (VF) and Woseco (WSC) networks split by fluency. Measures are reported for the largest connected component (LCC).

Given these differences in network topology, it is unsurprising that the spreading activation dynamics also differed between groups since larger networks offer more pathways for activation to travel, which affects measures like convergence time and peak activation levels. We then examined how the spreading activation features correlated with individual AUT scores within each fluency group separately (Figure 11). The low-fluency group showed only two significant correlations ($p < 0.05$): convergence time on the VF Science network ($\rho = 0.15$) and the stationary activation level on the WSC Science network ($\rho = 0.26$). In contrast, the high-fluency group showed a broader pattern of significant associations. The strongest effect was the stationary activation level on VF School ($\rho = 0.30$), followed by maximum activation on WSC Science ($\rho = 0.21$), the percentage of valid activated words on WSC Science ($\rho = -0.20$), stationary activation on VF Science ($\rho = 0.19$), convergence time on VF Science ($\rho = 0.15$), and time to maximum activation on VF School ($\rho = 0.15$).

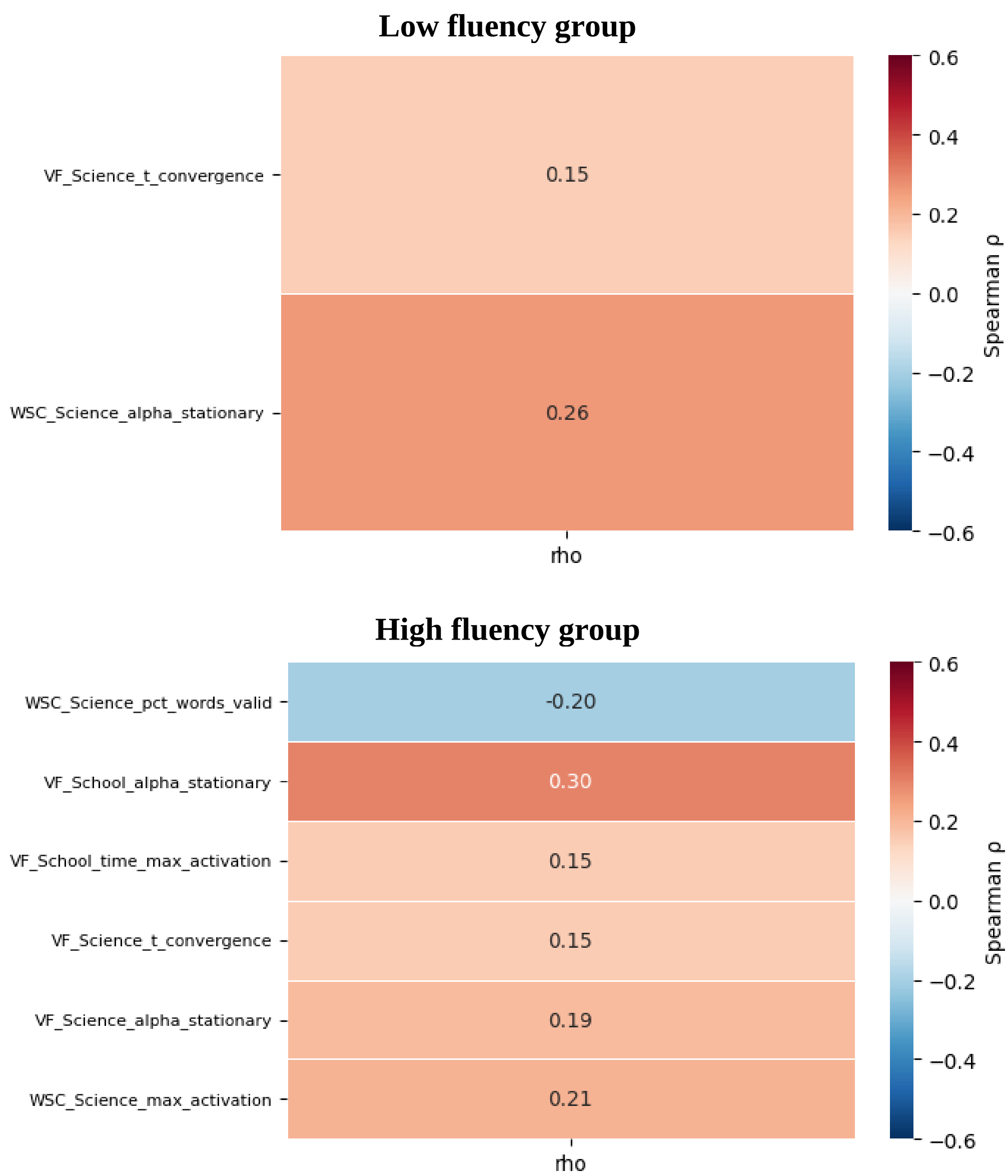

**Figure 11.** Correlogram showing significant correlations ($p < 0.05$) between spreading activation measures and AUT scores separated by fluency group.

Across both fluency groups, the direction of these correlations is consistent: participants with higher AUT scores tended to show higher stationary activation levels, and higher peak activation, which suggest that their underlying semantic networks support richer and more widespread activation. Interestingly, WSC_Science_pct_words_valid showed a negative correlation with the AUT scores ($\rho$ = -0.20). This measure captures the percentage of words a participant produced in the Woseco task that are actually present in the group-level network. To understand why this matters, recall that the group-level fluency and Woseco networks were built using a filtering step that retained only words produced by at least two participants in that group. Thereby, idiosyncratic responses were removed. When we then ran spreading activation for each individual, we could only activate words that existed in the group network (i.g. valid words), which contains only words the participant shared with others. The percentage of valid words therefore reflects how much a participant's vocabulary overlapped with the group. A high percentage means most of what a participant wrote was also written by others; a low percentage means they produced many unique words that nobody else thought of. The negative correlation with AUT scores tells us that more creative participants produced more idiosyncratic words in the Woseco task that did not appear in anyone else's responses. This aligns with the associative theory of creativity, claiming that highly creative individuals tend to make remote and unusual associations, reaching parts of semantic memory that others do not readily access (Kenett, 2019; Mednick, 1962).

### 4.5 Correlations of single and merged TFM Measures with AUT scores

The structural reducibility analysis identified merging the original TFM layer with both WSC-TFM topic layers as the optimal configuration across datasets (Section 4.2). According to the structural reducibility this merged network would retain most of the information of the separate layers with reduced complexity. However, it remains an open question whether this also translates into stronger associations with creativity scores. We therefore examined the Spearman correlations between network measures and AUT scores separately for the single TFM layer and the merged TFM+WSC TFM network (see Figure 12), using the same non-redundancy filter described in Section 3.2.1.

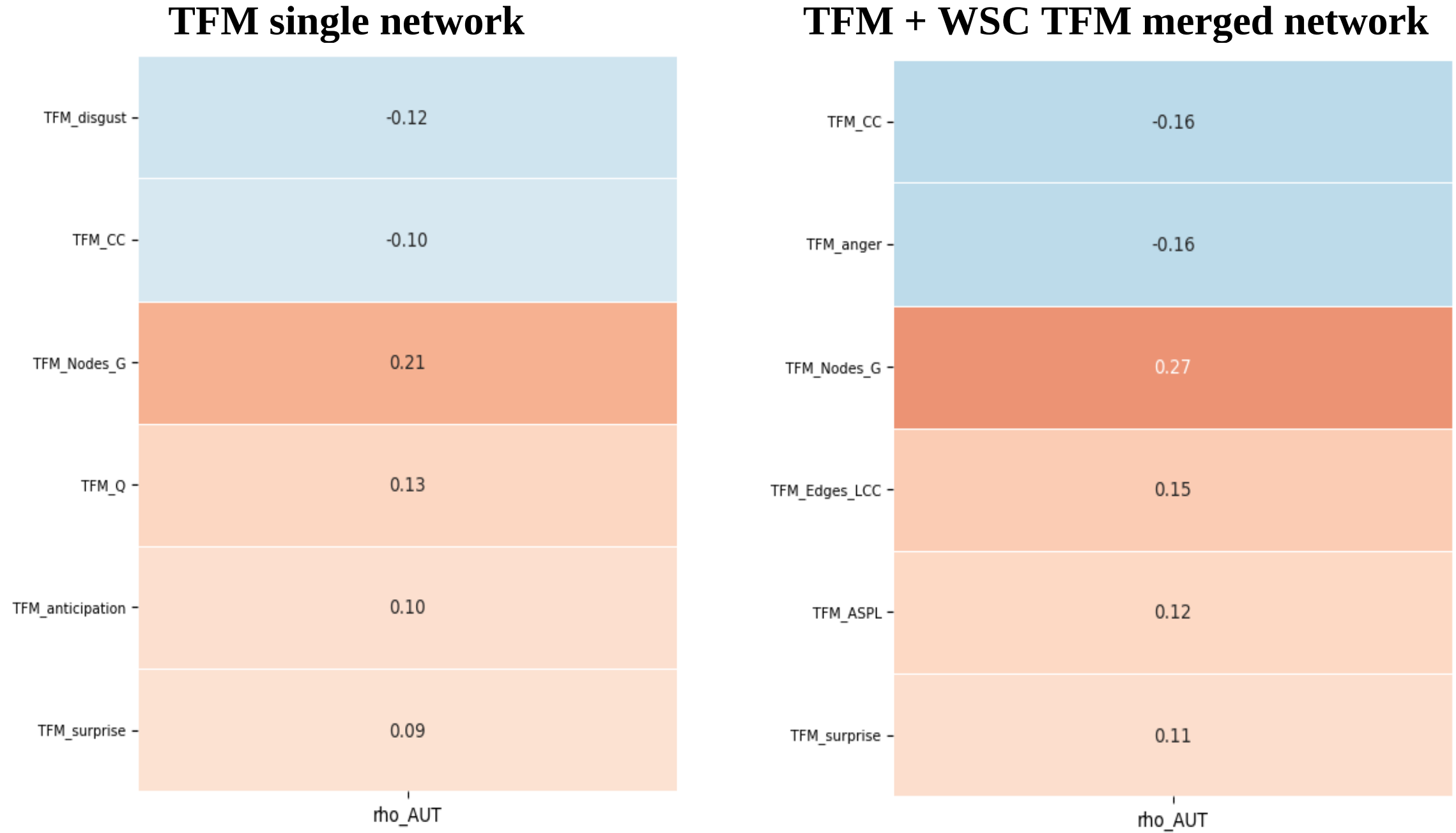

**Figure 12.** Correlograms showing significant Spearman correlations ($p < 0.05$) between TFM network measures and AUT scores. The left panel shows correlations for the single layer TFM network, and the right panel for the merged TFM+WSC TFM network.

Both configurations produced six non-redundant features with significant correlations with AUT scores, but the pattern differed. In both cases, network size (TFM_Nodes_G) showed the strongest positive correlation, and clustering coefficient (TFM_CC) showed a negative one, suggesting that more creative participants tend to have larger but less locally clustered TFM networks. The two configurations differed, however, in which emotional features reached significance. The single TFM layer showed significant correlations ($p < 0.05$) with disgust ($\rho = -0.12$), anticipation ($\rho = 0.10$), and surprise ($\rho = 0.09$) while in the merged networks only anger ($\rho = -0.16$) and surprise ($\rho = 0.11$) reached significance. The merged network also produced a significant correlation with ASPL ($\rho = 0.13$), associating higher average path lengths with higher creativity. Notably, the negative emotional correlations (disgust, anger) and positive ones (anticipation, surprise) are consistent across both configurations in their direction. Participants with higher AUT scores tended to have TFM networks with fewer negative-emotion associations and more surprise- and anticipation-related ones, which may reflect a more open and forward-looking associative style.

To test whether the structural advantage of the merged configuration also produces better AUT score predictions, in the following we ran two separate versions of the machine learning model. One using the single TFM layer features and one using the merged TFM+WSC TFM features as an exploratory comparison.

### 4.6 Exploratory Machine Learning Prediction of AUT Scores

As an exploratory analysis, we tested whether the selected network features could predict individual AUT scores using a Ridge regression model in a nested cross-validation framework (3 outer folds, 3 inner folds). We treat this as exploratory because the model selection phase already indicated modest predictive performance across all tested regressors (see Section 3.2.2).

Table 18 shows the nested cross-validation performance for both model versions. Both models outperformed the null model (RMSE = 0.076), but overall performance was modest. The model using the merged TFM+WSC TFM features ($R^2$ = 0.116, Spearman r = 0.29) outperformed the single TFM layer model ($R^2$ = 0.076, Spearman r = 0.25), even though only six of the twelve selected predictive features were related to the TFM network used. The better performance of the merged network supports the structural reducibility results suggesting that the merged configuration captures more relevant information.

| **Dataset** | **$R^2$** | **RMSE** | **Pearson_r** | **Spearman_r** |
|---|---|---|---|---|
| TFM single | 0.076 | 0.073 | 0.276 | 0.246 |
| TFM+WSC TFM | 0.116 | 0.071 | 0.340 | 0.293 |

**Table 18.** Machine learning performance with the ridge regressor, using nested cross-validation for 3 inner and outer folds. For both cases the null RMSE was 0.076. All correlations had a p-value< 0.001.

To examine how individual features contributed to predictions, we computed SHAP values for the better-performing merged model and visualised them as a beeswarm plot (Figure 13). The feature contributions were largely consistent with the correlations of spreading activation dynamics (Section 4.4) and TFM measures (Section 4.5) with AUT scores. TFM_Nodes_G was the most influential feature: participants with larger TFM networks, reflecting more diverse vocabulary in their texts, were predicted to have higher creativity scores. TFM_CC contributed in the opposite direction, with lower clustering coefficients predicting higher creativity, suggesting that more creative participants have less locally clustered associative structures. Among the emotional features, lower anger and higher surprise in the TFM network both predicted higher creativity, while TFM_disgust showed a positive SHAP contribution, which is an unexpected direction given its negative correlation with AUT scores described in Section 4.5. Among the spreading activation features, higher maximum activation on WSC Science and higher stationary activation levels on VF School both predicted higher creativity. WSC_Science_pct_words_valid showed a negative contribution, replicating the finding in Section 4.4 that more idiosyncratic responders, i.e. those whose words were less represented in the group network, tended to score higher on the AUT. TFM_ASPL contributed positively, with greater average path length in the TFM network predicting higher creativity.

The two spreading activation features VF_School_time_max_activation and VF_Science_t_convergence showed SHAP values close to zero, making their directional contributions difficult to interpret with confidence. Tentatively, shorter times to reach maximum activation and convergence appeared to predict higher creativity, but given the negligible effect sizes we do not interpret these further.

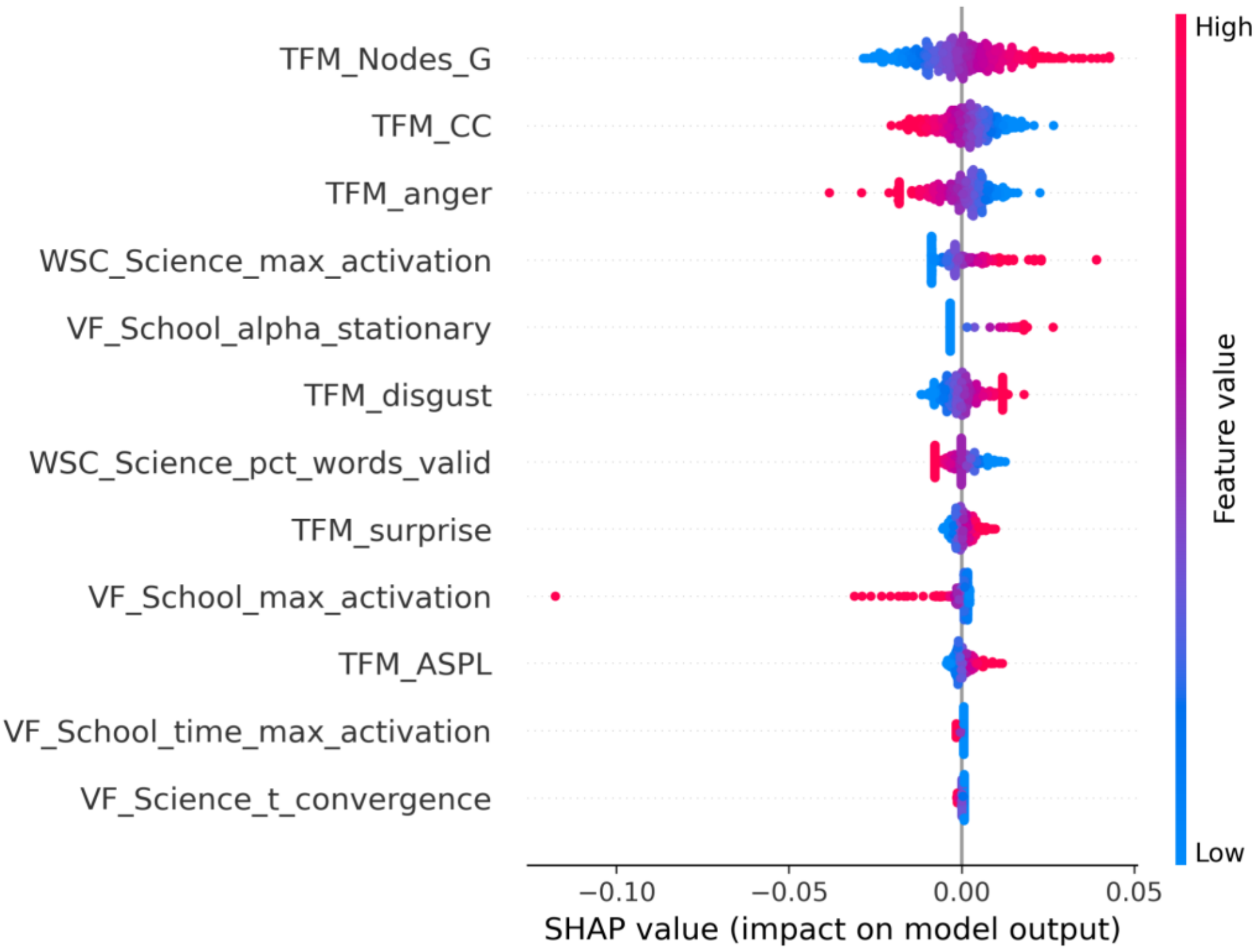

**Figure 13.** Beeswarm plot showing the importance of each feature in the machine learning model. All structural and emotional TFM features are computed on the merged TFM+WSC TFM layer.

One inconsistency can be found in the model regarding the two max_activation features. While higher maximum activation on WSC Science predicted higher creativity, the reverse was true for VF School, where higher maximum activation predicted lower creativity. This may reflect a task-specific difference since the fluency network is created from a simple association list, whereas the Woseco task involved the creation of a chain of sentences, resulting in smaller networks with differing network topology than the fluency networks (see Section 4.2 for a detailed table on the network measures of all four datasets). Such differences in network topology may influence spreading activation dynamics and their correlation with creativity scores.

Given the modest $R^2$ values overall, these SHAP results should be read as exploratory pointers rather than robust conclusions. The direction and ranking of feature contributions nonetheless provide a useful foundation for future work using larger samples or alternative creativity measures.

## 5. Discussion

The present study investigated whether combining multiple cognitive tasks within a multiplex network framework can provide richer information about the semantic organisation underlying creative thinking than any single task alone. Semantic networks were constructed from several language-based tasks and analysed through structural network measures, emotion features and spreading activation dynamics. These features were then used to predict individual creativity scores using machine learning models.

Overall, the results support the value of a multiplex approach in investigating creativity scores and mental representations. Structural reducibility analyses identified the TFM and WSC-TFM layers as the most similar in structure, reflecting their shared basis in semantic-syntactic networks derived from texts, while other layer combinations remained irreducible. This indicates that different tasks capture distinct aspects of semantic memory. In the machine learning model, network structural measures emerged as the strongest individual-level predictors of creativity, with spreading activation dynamics providing independent complementary signal, and emotional features playing a minor role.

**Multiplex semantic networks capture complementary information across tasks.**

A central contribution of this study lies in the use of multiplex semantic networks derived from different cognitive tasks. Previous work has shown that semantic networks constructed from single tasks can reveal structural properties related to creativity (Benedek et al., 2017; Kenett et al., 2014, 2016). However, most studies rely on a single task - commonly verbal fluency - which inevitably captures only a limited portion of an individual's semantic memory organisation.

Multiplex network frameworks are theoretically motivated by the principle that different layers encode non-redundant information about a complex system (Battiston et al., 2014; De Domenico et al., 2015). The multiplex reducibility analyses performed in this study support the assumption that different tasks capture different, relevant network structure. The best reducibility configurations frequently merged the TFM layer with one or both WSC-TFM layers, indicating structural similarity between them. Other layer combinations remained irreducible, demonstrating that each task retains unique information and network structure. This structural finding also had direct predictive implications. A proof-of-concept machine learning model using network measures from the merged TFM+WSC TFM layer outperformed a model using the TFM single layer features alone ($R^2$ = 0.116 vs. 0.076). This supports the premise that merging structurally similar layers enriches rather than reduces the network representation.

Importantly, even when tasks appeared similar in topic - such as the verbal fluency and Woseco tasks - these layers were not consistently merged in the optimal multiplex configuration. Though both verbal fluency and Woseco networks were created from associative responses produced under time constraints, the Woseco task requires participants to generate a chain of sentences rather than isolated word lists. Thereby, the Woseco task captures syntactic-semantic relationships that differ from the associative processes typically measured by verbal fluency (E. Haim et al., 2026). The fact that spreading activation measures derived from both Woseco and verbal fluency networks were retained in the predictive models further supports the idea that semantic memory is multifaceted and different cognitive tasks probe partially distinct aspects of conceptual organisation (Steyvers & Tenenbaum, 2005).

Overall, our findings highlight the advantage of combining multiple tasks within a multiplex framework. Rather than relying on a single behavioral measure, multi-task semantic networks can capture a broader range of cognitive processes involved in creative thinking.

**Cognitive networks highlight structural differences between low- and high-creative people.**

Another key finding concerns the structural differences between low- and high-creative participants. For the human datasets, the multiplex reducibility analysis showed that networks from low- and high-creative groups generally could not be merged without loss of information. In other words, networks of highly creative participants substantially differed from the networks of low creative respondents. One exception for this was found in the Italian dataset, where the sample size was rather small.

This finding confirms previous assumptions that individuals with high levels of creativity exhibit different semantic network structures compared to low creative participants (Benedek et al., 2017; Kenett et al., 2014, 2016). Prior work has argued that creativity is associated with more flexible, well connected semantic networks, facilitating the exploration of remote associations (Kenett et al., 2014, 2016; Mednick, 1962). Nevertheless, it should be noted that in this research we did not test specific hypotheses concerning the structure of these networks such as path lengths and modularity (Kenett et al., 2014; Siew & Guru, 2023). The fact that networks from different creativity groups cannot be merged into one indicates that the overall semantic structure differs systematically between them.

Interestingly, the same pattern was not observed among the artificially generated networks. For this case, networks of simulated high- and low-creative groups could be merged in the reducibility analysis. This suggests that the structural differences characteristic of human networks did not exist in the AI data. This could be explained by the homogeneous nature of the simulated responses that produce networks with similar structures regardless of their creativity level. This underscores the substantial individual variability present in human semantic networks.

**Network structure and spreading activation can predict creativity levels.**

Despite the modest predictive performance of the proof-of-concept ML baseline ($R^2$ = 0.076–0.116), the pattern of feature contributions provide some valuable insights.

The strongest predictor was textual forma mentis (TFM) network size, with larger and more lexically diverse networks predicting higher creativity. This finding is consistent with previous work showing that larger semantic networks relate to higher creativity ratings (E. Haim, Fischer, et al., 2026). At the same time, low clustering and long average path lengths in the TFM network predicted high creativity. This contrasts with typical findings from verbal fluency networks where higher clustering and shorter distances are associated with high creative performance (Kenett & Faust, 2019; Siew & Guru, 2023). This discrepancy may be explained by the nature of text-based and fluency-based networks. In sentence-level co-occurrence networks like the TFM, longer paths and looser local structure may reflect a more elaborated and wide-ranging conceptual elaboration, rather than the efficient associative clustering that characterises fluency search. Haim, Fischer et al. (2026) observed a similar pattern in networks derived from short narratives, where higher ASPL also corresponded with higher creativity ratings.

Among the emotional features, higher creativity was predicted among texts demonstrating lower anger and more surprise. This aligns with mood-creativity research, where positive and negative activating moods were found to enhance creative fluency and originality compared to deactivating moods (Baas et al., 2008; De Dreu et al., 2008; Khalil et al., 2025). However, whether the emotional tone of the resulting product follows the same logic is a different question. Haim, Fischer et al. (2026) found that AI and human raters differed in how they relied on emotional content when assigning creativity scores to short narratives. Therefore, the obtained results should be considered exploratory in nature.

Among the spreading activation features, higher maximum activation and higher stationary activation levels predicted high creativity. When a node accumulates high maximum activation during spreading, this indicates that the node is well-connected and centrally embedded in the underlying network. Moreover, higher stationary activation refers to a node stabilising at a high activation level because the network structure continuously feeds activation to the node rather than briefly activating and then suppressing it. Both findings point to the same underlying network characteristics. Participants that scored higher in AUT were characterised by networks in which key concepts are more densely interconnected and more reachable from the rest of the network. This aligns with research linking creativity to broader and more effective activation of semantic memory (Benedek & Neubauer, 2013; Kenett, 2019; Mednick, 1962).

Finally, an unexpected result was the negative correlation between the percentage of idiosyncratic words and AUT scores. Participants who produced more idiosyncratic words, i.e. responses absent from the group-level network because nobody else produced them, scored higher on the AUT. This result is supported by earlier studies that found that highly creative individuals produce more uncommon responses in free association tasks (Benedek & Neubauer, 2013). This also raises a broader methodological point. Standard methods for creating group-based semantic networks, including verbal fluency and behavioural forma mentis networks, remove idiosyncratic responses as noise. However, our results suggest that this filtering step removes creativity-relevant signal. Highly creative individuals may be precisely those who produce responses that no one else thinks of. Originality, the skill to produce uncommon responses (Acar et al., 2017; Corazza, 2016; Runco & Jaeger, 2012), may be directly found in idiosyncratic responses. We therefore suggest that the proportion of idiosyncratic responses should be retained as an explicit feature in future network-based creativity research rather than discarded by default.

**Limitations and future directions**

Several limitations should be acknowledged. First, sample sizes differed between datasets, with the Italian dataset being substantially smaller (n = 78 vs. 130-160). Thus, the results specific to the Italian dataset should be interpreted with caution. Furthermore, even though the datasets were collected from different national and linguistic backgrounds, the present study does not attempt to assess cultural or linguistic differences in creativity per se. Semantic structure is known to vary across languages (Steyvers & Tenenbaum, 2005), and future cross-cultural studies with larger, balanced samples would be necessary to distinguish whether the observed differences in network characteristics stem from language-specific semantics or other factors.

A further limitation lies with the chosen creativity measure that served as the target for prediction in the machine learning model. In this study, AUT originality scores were generated by two AI programs, which may introduce some measurement noise. AI raters have been found to produce homogeneous scores lacking variance (E. Haim, Fischer, et al., 2026). This fact may have constrained the predictive performance of the machine learning models, resulting in low $R^2$ values. Alternative approaches could use originality ratings from humans, or using a combination of originality with other creativity dimensions like fluency and flexibility.

Lastly, semantic networks were built from task-specific formats and linguistic outputs. While these tasks capture important aspects of semantic processing, they do not exhaustively represent semantic memory functioning in everyday life. Future studies should examine additional task formats and naturalistic language samples to expand the range of the multiplex framework.

## Conclusion

This study introduces multiplex networks as a framework for representing the multifaceted associative knowledge underlying creative cognition. By constructing semantic networks from six cognitive tasks and integrating them into multiplex structures across four nationality samples, we demonstrated that different tasks capture complementary and largely non-redundant aspects of semantic organisation.

In our machine learning framework, network structural measures emerged as the strongest individual-level predictors of creativity, with spreading activation dynamics providing complementary signal. This suggests that both the architecture of semantic memory and the dynamics of activation are relevant to creative performance.

Together, these results argue for a view of creativity emerging from the joint interplay of semantic memory structure and dynamic retrieval processes. Multiplex cognitive network modelling, combined with spreading activation dynamics and interpretable machine learning, offers a promising approach for studying individual creative differences in education, creativity training and assessment.

## Data Availability Statement

The complete datasets used in this study, instructions to human participants and codebooks to prompt digital twins, extracted network measures used for machine learning, and the Python and R notebooks for network construction are publicly available on the OSF database under the following link: https://osf.io/ns952/overview

## Acknowledgements

The authors used OpenAI's ChatGPT (GPT-5, 11/2025 release) and Claude (Sonnet 4.6, 02/2026) to assist with language editing and to improve clarity. All content was carefully reviewed, verified and finalised by the authors.

## Declaration of competing interests

The authors have no competing interests to declare.

# Appendix

## I. Instructions to participants in English, German and Italian

| English | German | Italian |
|---|---|---|
| Welcome!<br><br>Thank you for participating in this study.<br><br>Continue with SPACEBAR. | Willkommen!<br><br>Danke für Ihre Teilnahme an dieser Studie.<br><br>Weiter mit LEERTASTE. | Benvenuto/a!<br><br>Grazie per aver partecipato a questo studio.<br><br>Premere SPAZIO per continuare. |
| **<u>Declaration of consent</u>**<br><br>I hereby declare:<br>° that my participation in the study is my free decision.<br><br>° that my participation is not influenced by financial or other advantages.<br><br>I am informed about the planned activities and know how I can participate in the study.<br><br>Since this is the case:`;<br><br>consent_giving1 = `° I GIVE my consent by pressing the ENTER key.`;<br>consent_giving0 = `° I REFUSE my consent by pressing the ESC key (ESC ends the study).`; | **<u>Einverständniserklärung</u>**<br><br>Ich erkläre hiermit:<br>° dass meine Teilnahme an der Studie meine freie Entscheidung ist.<br><br>° dass meine Teilnahme weder durch finanzielle noch sonstige Vorteile beeinflusst ist.<br><br>Ich bin über die geplanten Aktivitäten informiert und weiß, wie ich an der Studie teilnehmen kann.<br><br>Da dies der Fall ist:`;<br><br>consent_giving1 = `° GEBE ich mein Einverständnis mit der ENTER-Taste.`;<br>consent_giving0 = `° VERWEIGERE ich mein Einverständnis mit der ESC-Taste. (ESC beendet die Studie)`; | **<u>Consenso Informato</u>**<br><br>Con la presente dichiaro:<br>° che la mia partecipazione allo studio è una decisione libera e autonoma.<br><br>° che la mia partecipazione non è influenzata da vantaggi economici o di altro tipo.<br><br>Sono informato/a sulle attività previste e so come posso partecipare allo studio.<br><br>Pertanto :`;<br><br>consent_giving1 = `° PRESTO il mio consenso premendo il tasto INVIO.`;<br><br>consent_giving0 = `° RIFIUTO il mio consenso premendo il tasto ESC (ESC termina lo studio).`; |
| **<u>Important Information</u>**<br><br>°Always read the information carefully. | **<u>Wichtige Informationen</u>**<br><br>° Lesen Sie sich die Angaben immer genau durch. | **<u>Informazioni importanti</u>**<br><br>° Leggere sempre attentamente le informazioni. |
| You CANNOT go back to previous pages.<br><br>° Attention: The ESC button ends the experiment completely. Please do NOT press it accidentally.<br><br>° The bar below shows your progress with the tasks at certain points.<br><br><br>Press SPACEBAR to start the study. | Sie können NICHT zu vorherigen Seiten zurück.<br><br>° Achtung: Die ESC-Taste beendet das Experiment vollständig. Bitte NICHT versehentlich drücken.<br><br>° Der untere Balken zeigt Ihren Fortschritt mit den Aufgaben an.<br><br>Drücken Sie LEERTASTE, um die Studie zu starten. | Non è possibile tornare alle pagine precedenti.<br><br>° Attenzione: il tasto ESC termina completamente l'esperimento. Si prega di non premerlo accidentalmente.<br><br>° La barra sottostante mostra lo stato di avanzamento nello svolgimento delle attività.<br><br>Premere SPAZIO per iniziare lo studio. |
| **PART 1 - STIMULI WORDS**<br><br>In the following, 10 stimulus words are presented one after the other. For each of them, write down three associations that spontaneously come to mind.<br>Do not think about it too long and answer spontaneously.<br><br>Please write each word on a new line, surrounded by a blue box.<br>Use the TAB key [\u21B9] or the mouse to move to the next line.<br><br>Start the task by pressing the SPACEBAR | **Teil 1 - REIZWÖRTER**<br><br>Im folgenden werden nacheinander 10 Reizwörter präsentiert. Schreiben Sie für jedes davon drei Assoziationen auf, die Ihnen spontan in den Sinn kommen. Denken Sie nicht zu lange darüber nach.<br><br>Bitte schreiben Sie jedes Wort in eine neue Zeile. Verwenden Sie die TAB Taste [\u21B9] oder die Maus, um zur nächsten Zeile zu gelangen.<br><br>Starten Sie die Aufgabe mit LEERTASTE. | **PARTE 1 - PAROLE STIMOLO**<br><br>Di seguito verranno presentate 10 parole stimolo, una dopo l'altra. Per ciascuna, scrivi tre associazioni che ti vengono in mente.<br>Non riflettere troppo a lungo e rispondi in modo spontaneo.<br><br>Scrivi ogni parola su una nuova riga, nel riquadro blu che apparirà a schermo.<br>Usa il tasto TAB [\u21B9] o il mouse per passare alla riga successiva.<br><br>Per iniziare il compito premere SPAZIO. |
| Association 1:<br>Association 2:<br>Association 3:<br><br>→ Next | Assoziation 1:<br>Assoziation 2:<br>Assoziation 3:<br><br>→ Weiter | Associazione 1:<br>Associazione 2:<br>Associazione 3:<br><br>→ Avanti |
| • Life<br>• Physics | • Leben<br>• Physik | • Vita<br>• Fisica |

<table>
<tr><td><ul><li>School</li><li>Biology</li><li>Mathematics</li><li>Sustainability</li><li>Creativity</li><li>University</li><li>Art</li><li>Chemistry</li></ul></td><td><ul><li>Schule</li><li>Biologie</li><li>Mathematik</li><li>Nachhaltigkeit</li><li>Kreativität</li><li>Universität</li><li>Kunst</li><li>Chemie</li></ul></td><td><ul><li>Scuola</li><li>Biologia</li><li>Matematica</li><li>Sostenibilità</li><li>Creatività</li><li>Università</li><li>Arte</li><li>Chimica</li></ul></td></tr>
<tr><td>The words you have written for the stimulus words will now be displayed.<br><br>Please rate all the words on a scale from 1 (very negative) to 5 (very positive) according to your personal perception.<br><br>Does the word have a positive or negative rating for you personally?<br><br><br>Start with SPACEBAR.</td><td>Nun werden Ihre Wörter eingeblendet, die Sie zu den Reizwörtern geschrieben haben.<br><br>Bitte bewerten Sie alle Wörter auf einer Skala von 1 (sehr negativ) bis 5 (sehr positiv) nach Ihrer persönlichen Empfindung.<br><br>Hat das Wort für Sie persönlich eine positive oder negative Wertung?<br><br><br>Starten mit LEERTASTE.</td><td>Verranno ora mostrate le parole fornite in precedenza in risposta alle parole stimolo.<br><br>Il compito consiste nel valutare ogni parola su una scala da 1 (molto negativo) a 5 (molto positivo), in base alla propria percezione personale.<br><br>La valutazione deve indicare se la parola ha una valenza personale positiva o negativa.<br><br>Premere SPAZIO per iniziare.</td></tr>
<tr><td><u>PART 2 - FINDING WORDS</u><br><br>Continue with SPACEBAR.</td><td><u>TEIL 2 - WÖRTER FINDEN</u><br><br>Weiter mit LEERTASTE.</td><td><u>PARTE 2 - PRODUZIONE DI PAROLE</u><br><br>Premere SPAZIO per continuare.</td></tr>
<tr><td><u>Finding words</u><br><br>Please write down as many words as you can think of on a given topic.<br><br>All words should be nouns.<br>Example on topic “soccer”: goalkeeper, field, referee, game<br><br>After each entry, press the Enter key [\u23CE]</td><td><u>Wörter finden</u><br><br>Bitte schreiben Sie zu einem vorgegebenen Thema so viele Wörter auf, wie Ihnen einfallen.<br><br>Alle Wörter müssen Nomen (= Hauptwörter) sein.<br>z.B. Thema "Fußball": Torhüter, Feld, Schiedsrichter, Spiel<br><br>Drücken Sie nach jedem Eintrag die Enter-Taste</td><td><u>Produzione di parole</u><br><br>Il compito consiste nello scrivere il maggior numero possibile di parole relative ad un dato argomento.<br><br>Tutte le parole devono essere sostantivi.<br>Esempio sull'argomento “calcio”: portiere, campo, arbitro, partita.<br><br>Dopo ogni inserimento, premi il tasto Invio</td></tr>
</table>

| | | |
|---|---|---|
| before writing the next word.<br>The written word disappears and you can write the next word.<br><br>The words should not be repeated.<br>Feel free to be creative within the topic!<br><br>You have 1 minute per topic.<br><br><br><br>Continue with SPACEBAR. | [\u23CE], bevor Sie das nächste Wort schreiben.<br>Das geschriebene Wort verschwindet und Sie können das nächste Wort schreiben.<br><br>Die Wörter sollen sich nicht wiederholen.<br>Seien Sie innerhalb des Themas gerne kreativ!<br><br>Sie haben pro Thema 1 Minute Zeit.<br><br><br>Weiter mit LEERTASTE. | [\u23CE] prima di scrivere la parola successiva.<br>La parola scomparirà e sarà possibile scrivere quella successiva.<br><br>E' importante non ripetere le parole. Per il resto, la creatività è benvenuta, purché sia attinente al tema.<br>Il tempo a disposizione è di 1 minuto per ogni argomento.<br><br>Premere SPAZIO per continuare. |
| The topic is displayed on the next page.<br>You can then enter the nouns in the empty field.<br><br>There are now 2 topics.<br>You have 1 minute for each topic.<br><br><br>Start with SPACEBAR. | Auf der nächsten Seite wird das Thema eingeblendet.<br>Danach können Sie in das leere Feld die Nomen eintragen.<br><br>Es folgen nun 2 Themen.<br>Für jedes Thema haben Sie 1 Minute Zeit.<br><br>Starten mit LEERTASTE. | L'argomento verrà visualizzato nella pagina successiva.<br>Potrai quindi inserire la parola nella casella vuota.<br><br>Ti verranno ora presentati 2 argomenti.<br>Hai 1 minuto per ognuno.<br><br>Premere SPAZIO per iniziare. |
| Write as many nouns as possible on:<br><br>● School<br>● Natural science | Schreiben Sie möglichst viele Nomen zu:<br><br>● Schule<br>● Naturwissenschaft | Scrivi più sostantivi possibili su:<br><br>● Scuola<br>● Scienze naturali |
| Time is up.<br><br>The study continues in a few seconds. | Die Zeit ist um.<br><br>In wenigen Sekunden geht es weiter. | Il tempo è scaduto.<br><br>Lo studio riprenderà tra pochi secondi. |
| **PART 3 - WORD-SENTENCE-CONSTRUCTIONS**<br><br>Continue with SPACEBAR. | **TEIL 3 - WORT-SATZ-KONSTRUKTIONEN**<br><br>Weiter mit LEERTASTE. | **PARTE 3 - COSTRUZIONI PAROLA-FRASE**<br><br>Premere SPAZIO per continuare. |

<table>
<tr>
<td><u>Word-Sentence-Constructions</u><br><br>In this task, you form sentences on a given topic.<br>Each sentence should contain two technical words. These can be nouns or verbs.<br>The following sentence takes up the last technical word and combines it with a new technical word.<br>This creates a chain of sentences on the topic, each of which is linked by a technical word.<br>After each sentence press the Enter key [\u23CE].<br><br>Example on the topic “soccer”:<br>The TEAM runs onto the FIELD. [Enter key \u23CE]<br>On the FIELD there are two GOALS. [Enter key \u23CE]<br><br>Continue with SPACEBAR.</td>
<td><u>Wort-Satz-Konstruktionen</u><br><br>Bei dieser Aufgabe bilden Sie zu einem vorgegebenen Thema Sätze.<br>Jeder Satz soll zwei Fachwörter enthalten. Dies können Nomen oder Verben sein.<br>Der darauffolgende Satz greift das letzte Fachwort auf und kombiniert es mit einem neuen Fachwort.<br>So entsteht zu dem Thema eine Kette von Sätzen, die jeweils über ein Fachwort verknüpft sind.<br>Drücken Sie nach jedem Satz die Enter-Taste [\u23CE].<br><br>Beispiel zum Thema "Fußball":<br>Die MANNSCHAFT läuft auf das FELD. [Enter-Taste \u23CE]<br>Auf dem FELD stehen zwei TORE. [Enter-Taste \u23CE]<br><br>Weiter mit LEERTASTE.</td>
<td><u>Costruzioni parola-frase</u><br><br>In questo esercizio dovrai formare delle frasi su un argomento dato.<br>Ogni frase deve contenere due parole tecniche.<br>Queste possono essere sostantivi o verbi.<br>La frase che segue riprende l'ultima parola tecnica e la combina con una nuova.<br>In questo modo si crea una catena di frasi sull'argomento, ciascuna delle quali è collegata da una parola tecnica.<br>Dopo ogni frase premi il tasto Invio [\u23CE].<br><br><br>Esempio sul tema “calcio”:<br>La SQUADRA corre sul CAMPO. [Tasto Invio \u23CE]<br>Sul CAMPO ci sono due PORTE. [Tasto Invio \u23CE]<br><br>Premere SPAZIO per continuare.</td>
</tr>
<tr>
<td>More detailed example on the topic "soccer":<br><br>The TEAM runs onto the FIELD. [Enter key \u23CE]<br>On the FIELD there are two GOALS. [Enter key \u23CE]<br>The GOALS are guarded by GOALKEEPERS. [Enter key \u23CE]<br>The GOALKEEPER may catch the BALL. [Enter key \u23CE]<br>Players may KICK the BALL. [Enter key \u23CE]<br>When KICKING...</td>
<td>Ausführlicheres Beispiel zum Thema "Fußball":<br><br>Die MANNSCHAFT läuft auf das FELD. [Enter-Taste \u23CE]<br>Auf dem FELD stehen zwei TORE. [Enter-Taste \u23CE]<br>Die TORE werden von TORHÜTERN bewacht. [Enter-Taste \u23CE]<br>Der TORHÜTER darf den BALL fangen. [Enter-Taste \u23CE]<br>Spieler dürfen den BALL auch KÖPFELN. [Enter-Taste \u23CE]<br>Beim KÖPFELN ...</td>
<td>Esempio più dettagliato sul tema “calcio”:<br><br>La SQUADRA corre sul CAMPO. [Tasto Invio \u23CE]<br>Sul CAMPO ci sono due PORTE. [Tasto Invio \u23CE]<br>Le PORTE sono difese dai PORTIERI. [Tasto Invio \u23CE]<br>Il PORTIERE può prendere la PALLA. [Tasto Invio \u23CE]<br>I giocatori possono CALCIARE la PALLA. [Tasto Invio \u23CE]<br>Quando CALCIANO...</td>
</tr>
</table>

| | | |
|---|---|---|
| Note: The target terms are only capitalized here for clarity.<br>Write all words normally without the capitalization.<br><br>Continue with SPACEBAR. | Anmerkung: Die Fachwörter sind hier nur zur Verdeutlichung groß geschrieben.<br>Schreiben Sie alle Fachwörter normal.<br><br><br>Weiter mit LEERTASTE. | Nota: i termini target sono scritti in maiuscolo solo per chiarezza.<br>Scrivi tutte le parole normalmente senza maiuscole.<br><br>Premere SPAZIO per continuare. |
| Is everything clear?<br><br>Now write chains of sentences like this for 2 topics.<br>Press the Enter key after each sentence.<br><br>You have 3 minutes for each topic.<br><br>Start with SPACEBAR. | Alles klar?<br><br>Schreiben Sie nun zu 2 Themen solche Satzketten.<br>Drücken Sie nach jedem Satz die Enter-Taste.<br><br>Für jedes Thema haben Sie 3 Minuten Zeit.<br><br>Starten mit LEERTASTE. | È tutto chiaro?<br><br>Ora scrivi catene di frasi, come nell'esempio precedente, per 2 argomenti differenti.<br>Premi il tasto Invio dopo ogni frase.<br><br>Hai a disposizione 3 minuti per ogni argomento.<br><br>Premere SPAZIO per iniziare il compito. |
| Write a chain of sentences for:<br>● School<br>● Science | Schreiben Sie ein Wosako zum Thema:<br>● Schule<br>● Naturwissenschaft | Scrivi una serie di frasi su:<br>● Scuola<br>● Scienze naturali |
| **PART 4 - CREATIVITY IN THE NATURAL SCIENCES**<br><br>Please write a text in 6 minutes in which you respond to the following question:<br><br>"To what extent does creativity play a role for you in science?"<br><br>Write enough so that the blue box is completely filled.<br><br>Start with SPACEBAR. | **TEIL 4 - KREATIVITÄT IN DER NATURWISSENSCHAFT**<br><br>Bitte schreiben Sie in 6 Minuten einen Text, in dem Sie auf die folgende Frage eingehen:<br><br>"Inwiefern spielt Kreativität für Sie in der Naturwissenschaft eine Rolle?"<br><br>Schreiben Sie so viel, dass die blaue Box ganz ausgefüllt ist.<br><br>Starten mit LEERTASTE. | **PARTE 4 - CREATIVITÀ NELLE SCIENZE NATURALI**<br><br>Hai 6 minuti per rispondere alla seguente domanda:<br><br>"In che misura, per te, la creatività gioca un ruolo importante nella scienza?"<br><br>Scrivi abbastanza così da riempire completamente il riquadro blu.<br><br>Premere SPAZIO per iniziare il compito. |
| To what extent does creativity play a role for | Inwiefern spielt Kreativität für Sie in der | In che misura, per te, la creatività gioca un ruolo |

| you in science? | Naturwissenschaft eine Rolle? | importante nella scienza? |
|---|---|---|
| Fantastic, you are more than halfway done!<br><br>Take a short break, stretch, have a drink, lean back and rest your eyes for a moment.<br>Clear your head, because you will then have to do some creative tasks.<br><br>The study continues after 1:30 minutes. | Fantastisch, mehr als die Hälfte ist schon geschafft!<br><br>Machen Sie kurz eine Pause, strecken Sie sich, trinken Sie etwas, lehnen Sie sich zurück und entlasten kurz Ihre Augen.<br>Bekommen Sie einen klaren Kopf, denn danach folgen Kreativitätsaufgaben.<br><br>Nach 1:30 Minuten geht es weiter. | Fantastico, sei più che a metà strada!<br><br>Fai una breve pausa, sgranchisciti, bevi qualcosa, rilassati e riposa gli occhi per un momento.<br>Libera la mente, perché poi dovrai svolgere alcuni compiti creativi.<br><br>Lo studio riprende tra 1 minuto e 30 secondi. |
| **PART 5 - CREATIVITY**<br><br>Now follows a series of tasks in which your creativity is required.<br><br>Find as many answers as possible.<br>Also write down creative and original answers.<br><br>Continue with SPACEBAR. | **TEIL 5 - KREATIVITÄT**<br><br>Nun folgt eine Reihe von Aufgaben, in denen Ihre Kreativität gefragt ist.<br><br>Finden Sie so viele Antworten wie möglich.<br>Notieren Sie auch kreative und originelle Antworten.<br><br>Weiter mit LEERTASTE. | **PARTE 5 - CREATIVITÀ**<br><br>Seguono ora una serie di compiti che richiedono la tua creatività.<br><br>Trova il maggior numero possibile di risposte.<br>Scrivi anche risposte creative e originali.<br><br>Premere SPAZIO per continuare. |
| **Task 1**<br><br>Find the most original and alternative uses for two given objects.<br>Also note down creative possibilities.<br><br>Press the Enter key [\u23CE] after each idea you have written.<br>The idea disappears and you can write a new idea.<br><br>You have 3 minutes for each item. | **Aufgabe 1**<br><br>Finden Sie für zwei vorgegebene Gegenstände möglichst originelle und alternative Verwendungsmöglichkeiten.<br>Notieren Sie auch kreative Möglichkeiten.<br><br>Drücken Sie die Enter-Taste [\u23CE] nach jeder geschriebenen Idee.<br>Die Idee verschwindet und Sie können eine neue Idee schreiben.<br><br>Für jeden Gegenstand haben Sie 3 Minuten Zeit. | **Task 1**<br><br>Trova gli usi più originali e alternativi per due oggetti dati.<br>Includi anche le proposte più creative.<br><br>Premi il tasto Invio [\u23CE] dopo ogni idea che hai scritto.<br>L'idea scomparirà e puoi scriverne una nuova.<br><br>Hai 3 minuti per ogni oggetto. |

| Start with SPACEBAR. | Starten mit LEERTASTE. | Premere SPAZIO per iniziare il compito. |
| --- | --- | --- |
| Find alternative uses for:<br>• Fork<br>• Bottle | Finden Sie alternative Verwendungen für:<br>• Gabel<br>• Flasche | Trova usi alternativi per:<br>• Forchetta<br>• Bottiglia |
| **Task 2**<br><br>Now intuitively find a word that matches two stimulus words.<br>Be creative and original without thinking too long.<br><br>Press the Enter key [\u23CE] after you have written the word to go to the next stimuli words.<br><br>Start with SPACEBAR. | **Aufgabe 2**<br><br>Finden Sie nun intuitiv ein Wort, das zu zwei Stimuliwörtern passt.<br>Seien Sie dabei auch kreativ und originell, ohne lange nachzudenken.<br><br>Drücken Sie die Enter-Taste [\u23CE] nachdem sie das Wort geschrieben haben, um zu den nächsten Stimuliwörtern zu gelangen.<br><br>Starten mit LEERTASTE. | **Task 2**<br><br>Ora trova intuitivamente una parola che sia correlata alle due parole stimolo.<br>Sii creativo e originale senza pensarci troppo a lungo.<br><br>Premi il tasto Invio [\u23CE] dopo aver scritto la parola per passare alle parole stimolo successive.<br><br>Premere SPAZIO per iniziare il compito. |
| • round - white<br>• fast - exciting<br>• smooth - angular<br>• bright - weak<br>• strong - lonely<br>• black - fine<br>• slow - thick<br>• sweet - soft<br>• liquid - healthy<br>• malleable - dark<br>• electric - boring<br>• solid - fragrant<br>• fresh - green<br>• dirty - hot<br>• small - wet | • rund - weiß<br>• schnell - spannend<br>• glatt - eckig<br>• hell - schwach<br>• stark - einsam<br>• schwarz - fein<br>• langsam - dick<br>• süß - weich<br>• flüssig - gesund<br>• formbar - dunkel<br>• elektrisch - langweilig<br>• fest - duftend<br>• frisch - grün<br>• schmutzig - heiß<br>• klein - nass | • rotondo - bianco<br>• veloce - eccitante<br>• liscio - spigoloso<br>• luminoso - debole<br>• forte - solitario<br>• nero - sottile<br>• lento - denso<br>• dolce - morbido<br>• liquido - sano<br>• malleabile - scuro<br>• elettrico - noioso<br>• solido - profumato<br>• fresco - verde<br>• sporco - caldo<br>• piccolo - bagnato |

| **PART 6 - PERSONALITY TRAITS**<br><br>To what extent do the following personality traits apply to you?<br><br>Please answer as quickly and intuitively as possible without thinking about it for too long.<br><br>Start with SPACEBAR. | **TEIL 6 - PERSÖNLICHKEITSMERKMALE**<br><br>Inwieweit treffen die folgenden Persönlichkeitsmerkmale auf Sie zu?<br><br>Bitte antworten Sie so rasch und intuitiv wie möglich, ohne lange darüber nachzudenken.<br><br>Starten mit LEERTASTE. | **PARTE 6 - TRATTI DELLA PERSONALITÀ**<br><br>In che misura i seguenti tratti della personalità si applicano a te?<br><br>Rispondi nel modo più rapido e intuitivo possibile, senza riflettere troppo a lungo.<br><br>Premere SPAZIO per iniziare. |
|---|---|---|
| I see myself as someone who...<br><br>Items:<br>● is reserved.<br>● is generally trusting.<br>● does a thorough job.<br>● is relaxed, handles stress well.<br>● has an active imagination.<br>● is outgoing, sociable.<br>● tends to find fault with others.<br>● tends to be lazy.<br>● gets nervous easily.<br>● has few artistic interests.<br><br>Answers:<br>● "strongly disagree",<br>● "disagree",<br>● "neither",<br>● "agree",<br>● "strongly agree" | Ich…<br><br>Items:<br>● bin eher zurückhaltend, reserviert.<br>● schenke anderen leicht Vertrauen, glaube an das Gute im Menschen.<br>● bin bequem, neige zur Faulheit.<br>● erledige Aufgaben gründlich.<br>● bin entspannt, lasse mich durch Stress nicht aus der Ruhe bringen.<br>● habe eine aktive Vorstellungskraft, bin phantasievoll.<br>● gehe aus mir heraus, bin gesellig.<br>● neige dazu, andere zu kritisieren.<br>● werde leicht nervös und unsicher.<br>● habe nur wenig künstlerisches Interesse.<br><br>Answers:<br>● "gar nicht",<br>● "eher nicht",<br>● "weder noch",<br>● "eher",<br>● "voll und ganz" | Mi considero una persona…<br><br>Items:<br>● riservata.<br>● che di solito si fida.<br>● che lavora in modo accurato.<br>● rilassata, gestisce bene lo stress.<br>● con un'immaginazione attiva.<br>● estroversa, socievole.<br>● che tende a trovare da ridire sugli altri.<br>● che tende ad essere pigra.<br>● che diventa facilmente apprensiva.<br>● con pochi interessi artistici.<br><br>Answers:<br>● “per niente”,<br>● “un po´”,<br>● “abbastanza”,<br>● “un bel po’”,<br>● “moltissimo” |
| The following words describe different feelings | Die folgenden Wörter beschreiben unterschiedliche | Le seguenti parole descrivono diversi sentimenti e |

| and perceptions.<br><br>On a five-point scale, indicate how these feelings generally apply to you - not at the moment, but in general.<br><br><br>Start with SPACEBAR. | Gefühle und Empfindungen.<br><br>Tragen Sie auf einer fünfstufigen Skala ein, wie diese Empfindungen im Allgemeinen auf Sie zutreffen - also nicht jetzt im Moment, sondern allgemein.<br><br><br>Starten mit LEERTASTE. | percezioni.<br><br>Su una scala da uno a cinque, indica in che misura questi sentimenti ti riguardano generalmente, non in questo preciso momento.<br><br><br>Premere SPAZIO per iniziare. |
|---|---|---|
| I generally feel…<br><br>Items:<br>● active<br>● distressed<br>● interested<br>● excited<br>● upset<br>● strong<br>● guilty<br>● scared<br>● hostile<br>● inspired<br>● proud<br>● irritable<br>● enthusiastic<br>● ashamed<br>● alert<br>● nervous<br>● determined<br>● attentive<br>● jittery<br>● afraid<br><br>Answers:<br>● "not at all", | Ich fühle mich im Allgemeinen…<br><br>Items:<br>● aktiv<br>● bekümmert<br>● interessiert<br>● freudig erregt<br>● verärgert<br>● stark<br>● schuldig<br>● erschrocken<br>● feindselig<br>● angeregt<br>● stolz<br>● gereizt<br>● begeistert<br>● beschämt<br>● wach<br>● nervös<br>● entschlossen<br>● aufmerksam<br>● durcheinander<br>● ängstlich<br><br>Answers:<br>● "gar nicht", | In generale mi sento...<br><br>Items:<br>● attivo/a<br>● angosciato/a<br>● interessato/a<br>● eccitato/a<br>● turbato/a<br>● forte<br>● colpevole<br>● spaventato/a<br>● ostile<br>● ispirato/a<br>● orgoglioso/a<br>● irritabile<br>● entusiasta<br>● imbarazzato/a<br>● vigile<br>● nervoso/a<br>● determinato/a<br>● attento/a<br>● agitato/a<br>● impaurito/a<br><br>Answers:<br>● “per niente”, |
| ● "a little",<br>● "moderately",<br>● "quite a bit",<br>● "extremely" | ● "ein bisschen",<br>● "ziemlich",<br>● "erheblich",<br>● "äußerst" | ● “un po´”,<br>● “abbastanza”,<br>● “un bel po’”,<br>● “moltissimo” |
| **PART 7 - DEMOGRAPHIC INFORMATION**<br><br>Almost done!<br><br>Now a few more questions about yourself.<br><br>Continue with SPACEBAR. | **TEIL 7 - ANGABEN ZU IHRER PERSON**<br><br>Fast geschafft!<br><br>Nun folgen noch ein paar Fragen zu Ihrer Person.<br><br>Weiter mit LEERTASTE. | **PARTE 7 - INFORMAZIONI DEMOGRAFICHE**<br><br>Ci siamo quasi!<br><br>Ora qualche altra domanda su di te.<br><br>Premere SPAZIO per continuare. |
| Age:<br>[open question]<br><br>Nationality:<br>[open question]<br><br>Gender:<br>● "male",<br>● "female",<br>● "diverse"<br><br>What do you study? (subject, degree, semester)<br>[open question]<br><br>→ Next | Alter:<br>[offene Frage]<br><br>Nationalität:<br>[offene Frage]<br><br>Welchem Geschlecht fühlen Sie sich zugehörig?<br>● "männlich",<br>● "weiblich",<br>● "divers"<br><br>Was studieren Sie? Fach, Bachelor/Master, Semester)<br>[offene Frage]<br><br>→ Weiter | Età:<br>[domanda aperta]<br><br>Nazionalità:<br>[domanda aperta]<br><br>A quale genere senti di appartenere?<br>● “maschio”,<br>● “femmina”,<br>● “altro”<br><br>Cosa studi? (corso, livello, anno)<br>[domanda aperta]<br><br>→ Avanti |
| Highest level of education<br><br>● "Doctorate", | Höchste abgeschlossene Ausbildung<br><br>● "Doktor", | Livello di istruzione più elevato<br><br>● “Dottorato”, |

<table>
<tr><td><ul><li>"Master´s Degree",</li><li>"Bachelor´s Degree"</li></ul><ul><li>"High school diploma",</li><li>"Vocational training",</li><li>"Middle school"</li></ul></td><td><ul><li>"Master",</li><li>"Bachelor"</li></ul><ul><li>"Matura",</li><li>"Lehre",</li><li>"Pflichtschule"</li></ul></td><td><ul><li>“Laurea magistrale”,</li><li>“Laurea triennale”</li></ul><ul><li>“Diploma scuola superiore”,</li><li>“Formazione professionale”,</li><li>“Scuola media”</li></ul></td></tr>
<tr><td>Highest level of education<br>of the parents (at least one parent)<ul><li>"Doctorate",</li><li>"Master´s Degree",</li><li>"Bachelor´s Degree"</li></ul><ul><li>"High school diploma",</li><li>"Vocational training",</li><li>"Middle school"</li></ul></td><td>Höchste abgeschlossene Ausbildung<br>der Eltern (mindestens eines Elternteils)<ul><li>"Doktor",</li><li>"Master",</li><li>"Bachelor"</li></ul><ul><li>"Matura",</li><li>"Lehre",</li><li>"Pflichtschule"</li></ul></td><td>Livello di istruzione più elevato<br>dei genitori (almeno uno dei due)<ul><li>“Dottorato”,</li><li>“Laurea magistrale”,</li><li>“Laurea triennale”</li></ul><ul><li>“Diploma scuola superiore”,</li><li>“Formazione professionale”,</li><li>“Scuola media”</li></ul></td></tr>
<tr><td>Done!<br>Thank you for participating!<br><br>Please wait until a green box appears before closing the page.</td><td>Fertig!<br>Danke für Ihre Teilnahme!<br><br>Bitte warten Sie ab, bis ein grünes Kästchen erscheint, bevor Sie die Seite schließen.<br><br>Bei Fragen schreiben Sie gerne eine Email an edith.haim@unitn.it</td><td>Fatto!<br>Grazie per aver partecipato!<br><br>Attendi fino a quando non compare un riquadro verde prima di chiudere la pagina.</td></tr>
<tr><td>Finished!</td><td>Fertig!</td><td>Finito!</td></tr>
</table>

**II. Full list of 57 potential predictors in the ML model**

| TFM | BFM | VF | WSC |
|---|---|---|---|
| TFM_Nodes_G | BFM_Nodes_G | VF_School_max_activation | WSC_School_max_activation |
| TFM_Nodes_LCC | BFM_Nodes_LCC | | |
| TFM_Edges_G | BFM_Edges_G | VF_School_time_max_activation | WSC_School_time_max_activation |
| TFM_Edges_LCC | BFM_Edges_LCC | | |
| TFM_Diameter | BFM_Diameter | VF_School_alpha_stationary | WSC_School_alpha_stationary |
| TFM_CC | BFM_CC | | |
| TFM_Q | BFM_Q | VF_School_t_convergence | WSC_School_t_convergence |
| TFM_ASPL | BFM_ASPL_LCC | | |
| TFM_anger | BFM_ASPL_mean | VF_School_pct_words_valid | WSC_School_pct_words_valid |
| TFM_trust | BFM_Degree_centr | | |
| TFM_surprise | BFM_Degree_assort | VF_Science_max_activation | WSC_Science_max_activation |
| TFM_disgust | BFM_Pagerank_range | | |
| TFM_joy | BFM_valence_mean | VF_Science_time_max_activation | WSC_Science_time_max_activation |
| TFM_sadness | BFM_valence_sd | | |
| TFM_fear | BFM_valence_skew | VF_Science_alpha_stationary | WSC_Science_alpha_stationary |
| TFM_anticipation | BFM_pct_positive | | |
| | BFM_pct_neutral | VF_Science_t_convergence | WSC_Science_t_convergence |
| | BFM_pct_negative | | |
| | BFM_Components | VF_Science_pct_words_valid | WSC_Science_pct_words_valid |
| | BFM_Components_4plus | | |
| | BFM_LCC_Proportion | | |

**Table A.1.** Full list of features extracted from the multiplex networks as potential predictors in the machine learning model. Here we group them according to the network type they are extracted from: textual forma mentis (TFM), behavioural forma mentis (BFM), verbal fluency (VF) and Woseco (WSC).

### III. Network Measures for AI and aggregated datasets

AT Simulations

| Task | Group | Nodes LCC | Edges LCC | Diameter | ASPL | CC | Q | Degree centr | Degree assort | PageRank Range |
|---|---|---|---|---|---|---|---|---|---|---|
| VF School | Sim High | 64 | 186 | 4 | 2.201 | 0.790 | 0.427 | 0.092 | -0.304 | 0.105 |
| | Sim Low | 65 | 189 | 5 | 2.338 | 0.773 | 0.469 | 0.091 | -0.253 | 0.098 |
| VF Science | Sim High | 58 | 168 | 6 | 3.193 | 0.719 | 0.597 | 0.102 | -0.111 | 0.035 |
| | Sim Low | 58 | 168 | 6 | 3.076 | 0.728 | 0.587 | 0.102 | -0.128 | 0.029 |
| WSC School | Sim High | 69 | 210 | 4 | 2.674 | 0.725 | 0.566 | 0.090 | -0.037 | 0.054 |
| | Sim Low | 68 | 211 | 5 | 2.449 | 0.720 | 0.498 | 0.093 | -0.176 | 0.064 |
| WSC Science | Sim High | 118 | 351 | 11 | 4.873 | 0.696 | 0.704 | 0.051 | -0.030 | 0.017 |
| | Sim Low | 90 | 268 | 7 | 3.510 | 0.700 | 0.668 | 0.067 | -0.003 | 0.030 |
| BFMN | Sim High | 103 | 124 | 6 | 3.688 | 0.033 | 0.654 | 0.024 | -0.768 | 0.067 |
| | Sim Low | 102 | 126 | 6 | 3.543 | 0.024 | 0.631 | 0.024 | -0.738 | 0.076 |
| TFMN | Sim High | 498 | 3114 | 6 | 2.726 | 0.577 | 0.299 | 0.025 | -0.142 | 0.023 |
| | Sim Low | 485 | 2879 | 6 | 2.782 | 0.523 | 0.329 | 0.025 | -0.125 | 0.021 |
| WSC-TFMN School | Sim High | 471 | 1614 | 8 | 3.211 | 0.318 | 0.407 | 0.015 | -0.099 | 0.025 |
| | Sim Low | 484 | 1618 | 8 | 3.205 | 0.338 | 0.415 | 0.014 | -0.102 | 0.025 |
| WSC-TFMN Science | Sim High | 758 | 2639 | 12 | 3.386 | 0.420 | 0.478 | 0.009 | -0.139 | 0.021 |
| | Sim Low | 713 | 2464 | 8 | 3.315 | 0.415 | 0.474 | 0.010 | -0.139 | 0.022 |
| Total Merged | Sim High | 1569 | 8184 | 12 | 3.271 | 0.433 | 0.462 | 0.007 | -0.103 | 0.011 |
| | Sim Low | 1519 | 7729 | 8 | 3.251 | 0.416 | 0.465 | 0.007 | -0.101 | 0.010 |

**Table A.2**. Descriptive network measures for AI-generated semantic networks corresponding to the Austrian dataset. Measures are reported for the largest connected component (LCC).

US Simulations

| Task | Group | Nodes LCC | Edges LCC | Diameter | ASPL | CC | Q | Degree centr | Degree assort | PageRank Range |
|---|---|---|---|---|---|---|---|---|---|---|
| VF School | Sim High | 27 | 75 | 4 | 2.137 | 0.756 | 0.321 | 0.214 | -0.258 | 0.070 |
| | Sim Low | 29 | 81 | 4 | 1.963 | 0.771 | 0.327 | 0.200 | -0.312 | 0.110 |
| VF Science | Sim High | 26 | 75 | 4 | 2.175 | 0.719 | 0.424 | 0.231 | -0.134 | 0.058 |
| | Sim Low | 28 | 81 | 4 | 2.101 | 0.726 | 0.376 | 0.214 | -0.221 | 0.077 |
| WSC School | Sim High | 105 | 309 | 7 | 2.942 | 0.742 | 0.594 | 0.057 | -0.168 | 0.067 |
| | Sim Low | 112 | 330 | 8 | 3.077 | 0.735 | 0.631 | 0.053 | -0.118 | 0.057 |
| WSC Science | Sim High | 86 | 252 | 6 | 2.807 | 0.749 | 0.587 | 0.069 | -0.157 | 0.061 |
| | Sim Low | 95 | 279 | 7 | 3.314 | 0.732 | 0.675 | 0.062 | -0.104 | 0.036 |
| BFMN | Sim High | 116 | 142 | 7 | 3.807 | 0.033 | 0.672 | 0.021 | -0.826 | 0.062 |
| | Sim Low | 109 | 131 | 7 | 4.105 | 0.045 | 0.668 | 0.022 | -0.820 | 0.061 |
| TFMN | Sim High | 610 | 5701 | 6 | 2.499 | 0.548 | 0.265 | 0.031 | -0.145 | 0.023 |
| | Sim Low | 574 | 5524 | 5 | 2.469 | 0.553 | 0.247 | 0.034 | -0.154 | 0.021 |
| WSC-TFMN School | Sim High | 491 | 3503 | 5 | 2.641 | 0.444 | 0.324 | 0.029 | -0.063 | 0.020 |
| | Sim Low | 454 | 3653 | 5 | 2.517 | 0.459 | 0.294 | 0.036 | -0.089 | 0.020 |
| WSC-TFMN Science | Sim High | 443 | 3139 | 6 | 2.691 | 0.511 | 0.323 | 0.032 | 0.003 | 0.016 |
| | Sim Low | 476 | 3235 | 6 | 2.761 | 0.482 | 0.346 | 0.029 | 0.021 | 0.014 |
| Total Merged | Sim High | 1237 | 12415 | 6 | 2.765 | 0.462 | 0.368 | 0.016 | -0.077 | 0.011 |
| | Sim Low | 1207 | 12561 | 6 | 2.725 | 0.465 | 0.370 | 0.017 | -0.082 | 0.009 |

**Table A.3**. Descriptive network measures for AI-generated semantic networks corresponding to the US dataset. Measures are reported for the largest connected component (LCC).

SG Simulations

| Task | Group | Nodes LCC | Edges LCC | Diameter | ASPL | CC | Q | Degree centr | Degree assort | PageRank Range |
|---|---|---|---|---|---|---|---|---|---|---|
| VF School | Sim High | 28 | 81 | 4 | 1.902 | 0.749 | 0.354 | 0.214 | -0.261 | 0.115 |
| | Sim Low | 32 | 92 | 4 | 2.036 | 0.775 | 0.348 | 0.185 | -0.302 | 0.107 |
| VF Science | Sim High | 25 | 72 | 4 | 2.097 | 0.716 | 0.369 | 0.240 | -0.308 | 0.068 |
| | Sim Low | 28 | 81 | 5 | 2.214 | 0.717 | 0.457 | 0.214 | -0.251 | 0.054 |
| WSC School | Sim High | 84 | 246 | 6 | 2.527 | 0.768 | 0.512 | 0.071 | -0.247 | 0.091 |
| | Sim Low | 91 | 267 | 5 | 2.844 | 0.754 | 0.592 | 0.065 | -0.185 | 0.075 |
| WSC Science | Sim High | 78 | 228 | 7 | 3.145 | 0.755 | 0.546 | 0.076 | -0.228 | 0.065 |
| | Sim Low | 78 | 228 | 6 | 2.829 | 0.746 | 0.579 | 0.076 | -0.154 | 0.061 |
| BFMN | Sim High | 96 | 111 | 10 | 4.232 | 0.027 | 0.696 | 0.024 | -0.748 | 0.071 |
| | Sim Low | 100 | 113 | 10 | 4.929 | 0.055 | 0.734 | 0.023 | -0.843 | 0.055 |
| TFMN | Sim High | 469 | 4399 | 5 | 2.434 | 0.548 | 0.260 | 0.040 | -0.153 | 0.021 |
| | Sim Low | 486 | 4489 | 5 | 2.449 | 0.566 | 0.271 | 0.038 | -0.152 | 0.022 |
| WSC-TFMN School | Sim High | 400 | 2955 | 6 | 2.547 | 0.454 | 0.301 | 0.037 | -0.098 | 0.022 |
| | Sim Low | 415 | 3392 | 6 | 2.508 | 0.440 | 0.300 | 0.039 | -0.092 | 0.018 |
| WSC-TFMN Science | Sim High | 400 | 2736 | 6 | 2.732 | 0.479 | 0.353 | 0.034 | 0.058 | 0.014 |
| | Sim Low | 419 | 2819 | 6 | 2.740 | 0.482 | 0.348 | 0.032 | 0.031 | 0.012 |
| Total Merged | Sim High | 1050 | 10259 | 6 | 2.753 | 0.467 | 0.377 | 0.019 | -0.075 | 0.009 |
| | Sim Low | 1050 | 10771 | 6 | 2.713 | 0.471 | 0.373 | 0.020 | -0.091 | 0.009 |

**Table A.4**. Descriptive network measures for AI-generated semantic networks corresponding to the Singaporean dataset. Measures are reported for the largest connected component (LCC).

IT Simulations

| Task | Group | Nodes LCC | Edges LCC | Diameter | ASPL | CC | Q | Degree centr | Degree assort | PageRank Range |
|---|---|---|---|---|---|---|---|---|---|---|
| VF School | Sim High | 31 | 89 | 4 | 2.215 | 0.735 | 0.447 | 0.191 | -0.110 | 0.074 |
| | Sim Low | 38 | 113 | 5 | 2.432 | 0.710 | 0.460 | 0.161 | -0.086 | 0.065 |
| VF Science | Sim High | 31 | 87 | 4 | 1.978 | 0.758 | 0.366 | 0.187 | -0.265 | 0.112 |
| | Sim Low | 36 | 102 | 4 | 2.363 | 0.744 | 0.466 | 0.162 | -0.209 | 0.078 |
| WSC School | Sim High | 53 | 153 | 5 | 2.482 | 0.743 | 0.526 | 0.111 | -0.166 | 0.069 |
| | Sim Low | 58 | 168 | 5 | 2.547 | 0.751 | 0.504 | 0.102 | -0.179 | 0.060 |
| WSC Science | Sim High | 60 | 177 | 5 | 2.676 | 0.725 | 0.555 | 0.100 | -0.078 | 0.049 |
| | Sim Low | 57 | 168 | 6 | 2.690 | 0.711 | 0.533 | 0.105 | -0.095 | 0.061 |
| BFMN | Sim High | 54 | 59 | 8 | 4.368 | 0.000 | 0.697 | 0.041 | -0.901 | 0.061 |
| | Sim Low | 46 | 53 | 6 | 3.742 | 0.009 | 0.619 | 0.051 | -0.808 | 0.076 |
| TFMN | Sim High | 300 | 2399 | 5 | 2.428 | 0.575 | 0.311 | 0.053 | -0.119 | 0.023 |
| | Sim Low | 310 | 2470 | 5 | 2.443 | 0.584 | 0.301 | 0.052 | -0.093 | 0.021 |
| WSC-TFMN School | Sim High | 227 | 1198 | 5 | 2.547 | 0.493 | 0.322 | 0.047 | -0.128 | 0.027 |
| | Sim Low | 225 | 1250 | 5 | 2.563 | 0.489 | 0.308 | 0.050 | -0.062 | 0.023 |
| WSC-TFMN Science | Sim High | 258 | 1104 | 8 | 3.009 | 0.456 | 0.423 | 0.033 | -0.027 | 0.024 |
| | Sim Low | 260 | 1045 | 6 | 2.985 | 0.456 | 0.443 | 0.031 | -0.040 | 0.020 |
| Total Merged | Sim High | 737 | 5081 | 7 | 2.964 | 0.492 | 0.460 | 0.019 | -0.054 | 0.014 |
| | Sim Low | 728 | 5207 | 6 | 2.969 | 0.489 | 0.465 | 0.020 | 0.003 | 0.010 |

**Table A.5**. Descriptive network measures for AI-generated semantic networks corresponding to the Italian dataset. Measures are reported for the largest connected component (LCC).

Aggregated networks

| Group | Nodes LCC | Edges LCC | Diameter | ASPL | CC | Q | Degree centr | Degree assort | PageRank Range |
|---|---|---|---|---|---|---|---|---|---|
| AT_Stud_High | 1879 | 9571 | 10 | 3.160 | 0.516 | 0.368 | 0.005 | -0.122 | 0.016 |
| AT_Stud_Low | 1789 | 9101 | 8 | 3.124 | 0.511 | 0.365 | 0.006 | -0.122 | 0.014 |
| US_Stud_High | 1497 | 14894 | 6 | 2.740 | 0.495 | 0.282 | 0.013 | -0.124 | 0.012 |
| US_Stud_Low | 1116 | 9933 | 6 | 2.778 | 0.501 | 0.310 | 0.016 | -0.117 | 0.014 |
| SG_Stud_High | 1762 | 17279 | 6 | 2.813 | 0.502 | 0.278 | 0.011 | -0.110 | 0.010 |
| SG_Stud_Low | 1568 | 15817 | 7 | 2.762 | 0.499 | 0.285 | 0.013 | -0.117 | 0.010 |
| IT_Stud_High | 1085 | 14946 | 6 | 2.624 | 0.589 | 0.359 | 0.025 | -0.107 | 0.010 |
| IT_Stud_Low | 1021 | 30259 | 5 | 2.410 | 0.656 | 0.320 | 0.058 | -0.171 | 0.007 |
| AT_Sim_High | 1569 | 8184 | 12 | 3.271 | 0.433 | 0.462 | 0.007 | -0.103 | 0.011 |
| AT_Sim_Low | 1519 | 7729 | 8 | 3.251 | 0.416 | 0.465 | 0.007 | -0.101 | 0.010 |
| US_Sim_High | 1237 | 12415 | 6 | 2.765 | 0.462 | 0.368 | 0.016 | -0.077 | 0.011 |
| US_Sim_Low | 1207 | 12561 | 6 | 2.725 | 0.465 | 0.370 | 0.017 | -0.082 | 0.009 |
| SG_Sim_High | 1050 | 10259 | 6 | 2.753 | 0.467 | 0.377 | 0.019 | -0.075 | 0.009 |
| SG_Sim_Low | 1050 | 10771 | 6 | 2.713 | 0.471 | 0.373 | 0.020 | -0.091 | 0.009 |
| IT_Sim_High | 737 | 5081 | 7 | 2.964 | 0.492 | 0.460 | 0.019 | -0.054 | 0.014 |
| IT_Sim_Low | 728 | 5207 | 6 | 2.969 | 0.489 | 0.465 | 0.020 | 0.003 | 0.010 |

**Table A.6**. Descriptive network measures for the aggregated multiplex semantic networks obtained by merging all task layers within each dataset and group for human and AI-generated responses. Measures are reported for the largest connected component (LCC).

## IV. Network Measures for VF and WSC networks split by fluency group

| Task | Fluency Group | Nodes LCC | Edges LCC | Diameter | ASPL | CC | Q | Avg Degree | Degree centr | Small Worldness |
|---|---|---|---|---|---|---|---|---|---|---|
| **AT Dataset** | | | | | | | | | | |
| VF School | AT High | 161 | 477 | 9 | 4.014 | 0.280 | 0.732 | 5.925 | 0.037 | 9.369 |
| | AT Low | 120 | 354 | 11 | 4.346 | 0.274 | 0.698 | 5.900 | 0.050 | 2.722 |
| VF Science | AT High | 141 | 417 | 10 | 4.613 | 0.279 | 0.733 | 5.915 | 0.042 | 3.549 |
| | AT Low | 88 | 258 | 6 | 3.197 | 0.259 | 0.659 | 5.864 | 0.067 | 2.814 |
| WSC School | AT High | 89 | 261 | 8 | 3.428 | 0.285 | 0.678 | 5.865 | 0.067 | 3.419 |
| | AT Low | 68 | 198 | 7 | 3.128 | 0.299 | 0.614 | 5.824 | 0.087 | 3.147 |
| WSC Science | AT High | 87 | 255 | 12 | 4.478 | 0.254 | 0.693 | 5.862 | 0.068 | 2.405 |
| | AT Low | 60 | 174 | 7 | 2.756 | 0.182 | 0.580 | 5.800 | 0.098 | 2.249 |
| **US Dataset** | | | | | | | | | | |
| VF School | US High | 171 | 507 | 9 | 3.988 | 0.294 | 0.732 | 5.930 | 0.035 | 6.663 |
| | US Low | 110 | 324 | 9 | 3.454 | 0.279 | 0.644 | 5.891 | 0.054 | 3.576 |
| VF Science | US High | 153 | 453 | 11 | 4.919 | 0.291 | 0.746 | 5.922 | 0.039 | 4.481 |
| | US Low | 94 | 276 | 7 | 3.194 | 0.261 | 0.658 | 5.872 | 0.063 | 4.666 |
| WSC School | US High | 121 | 357 | 9 | 3.866 | 0.285 | 0.712 | 5.901 | 0.049 | 3.617 |
| | US Low | 38 | 108 | 5 | 2.660 | 0.175 | 0.523 | 5.684 | 0.154 | 1.004 |
| WSC Science | US High | 92 | 270 | 12 | 4.300 | 0.245 | 0.699 | 5.870 | 0.065 | 2.666 |
| | US Low | 29 | 81 | 5 | 2.236 | 0.164 | 0.492 | 5.586 | 0.200 | 0.632 |
| **SG Dataset** | | | | | | | | | | |
| VF School | SG High | 172 | 510 | 11 | 4.687 | 0.285 | 0.745 | 5.930 | 0.035 | 4.904 |
| | SG Low | 119 | 351 | 10 | 4.361 | 0.262 | 0.704 | 5.899 | 0.050 | 4.392 |
| VF Science | SG High | 165 | 489 | 13 | 4.919 | 0.301 | 0.761 | 5.927 | 0.036 | 5.159 |
| | SG Low | 87 | 255 | 8 | 3.572 | 0.255 | 0.691 | 5.862 | 0.068 | 2.530 |
| WSC School | SG High | 114 | 336 | 7 | 3.850 | 0.265 | 0.720 | 5.895 | 0.052 | 4.433 |
| | SG Low | 54 | 156 | 5 | 2.772 | 0.298 | 0.605 | 5.778 | 0.109 | 4.149 |
| WSC Science | SG High | 98 | 288 | 8 | 3.575 | 0.270 | 0.706 | 5.878 | 0.061 | 3.723 |
| | SG Low | 61 | 177 | 6 | 2.762 | 0.202 | 0.622 | 5.803 | 0.097 | 1.565 |

| IT Dataset | | | | | | | | | | |
|---|---|---|---|---|---|---|---|---|---|---|
| VF School | IT High | 103 | 303 | 7 | 3.347 | 0.316 | 0.655 | 5.883 | 0.058 | 6.343 |
| | IT Low | 58 | 168 | 6 | 3.139 | 0.302 | 0.577 | 5.793 | 0.102 | 1.991 |
| VF Science | IT High | 90 | 264 | 7 | 3.825 | 0.307 | 0.695 | 5.867 | 0.066 | 3.441 |
| | IT Low | 39 | 111 | 6 | 2.688 | 0.279 | 0.535 | 5.692 | 0.150 | 1.507 |
| WSC School | IT High | 69 | 201 | 8 | 3.611 | 0.289 | 0.621 | 5.826 | 0.086 | 3.669 |
| | IT Low | 48 | 138 | 5 | 2.746 | 0.270 | 0.602 | 5.750 | 0.122 | 2.039 |
| WSC Science | IT High | 65 | 189 | 9 | 3.826 | 0.285 | 0.630 | 5.815 | 0.091 | 1.792 |
| | IT Low | 38 | 108 | 5 | 2.556 | 0.284 | 0.551 | 5.684 | 0.154 | 1.690 |

**Table A.7.** Full descriptive network measures for verbal fluency (VF) and Woseco (WSC) networks split by total fluency. Measures are reported for the largest connected component (LCC).

**V. Full Structural Reducibility - Comparisons high vs low**

Labelling:

| Layer | Abbreviation | Creativity | Task Title |
|---|---|---|---|
| 0 | High_BFM | High creative | Behavioural forma mentis |
| 1 | High_TFM | High creative | Textual forma mentis |
| 2 | High_WSC-TFM_school | High creative | Woseco-TFM method, topic “school” |
| 3 | High_WSC-TFM_science | High creative | Woseco-TFM method, topic “science” |
| 4 | High_VF_school | High creative | Verbal Fluency, topic “school” |
| 5 | High_VF_science | High creative | Verbal Fluency, topic “science” |
| 6 | High_WSC_school | High creative | Woseco, topic “school” |
| 7 | High_WSC_science | High creative | Woseco, topic “science” |
| 8 | Low_BFM | Low creative | Behavioural forma mentis |
| 9 | Low_TFM | Low creative | Textual forma mentis |
| 10 | Low_WSC-TFM_school | Low creative | Woseco-TFM method, topic “school” |
| 11 | Low_WSC-TFM_science | Low creative | Woseco-TFM method, topic “science” |
| 12 | Low_VF_school | Low creative | Verbal Fluency, topic “school” |
| 13 | Low_VF_science | Low creative | Verbal Fluency, topic “science” |
| 14 | Low_WSC_school | Low creative | Woseco, topic “school” |
| 15 | Low_WSC_science | Low creative | Woseco, topic “science” |

**Table A.8**. Explanation of layer labels and abbreviations in the structural reducibility analyses.

Highlighted in the tables below is the **highest q-value** which represents the best configuration for reducing complexity while maintaining relevant information.

**Austria - Human**

| q-value | Layer configuration |
|---|---|
| 0.286776 | [[0], [1], [2], [3], [4], [5], [6], [7], [8], [9], [10], [11], [12], [13], [14], [15]] |
| 0.293003 | [[0], [1], [2], [3], [4], [5], [6], [7], [8], [10], [12], [13], [14], [15], [9, 11]] |
| 0.300501 | [[0], [2], [4], [5], [6], [7], [8], [10], [12], [13], [14], [15], [9, 11], [1, 3]] |
| 0.28961 | [[0], [2], [4], [6], [8], [10], [12], [13], [14], [15], [9, 11], [1, 3], [5, 7]] |
| 0.298945 | [[0], [2], [4], [6], [8], [12], [13], [14], [15], [1, 3], [5, 7], [10, 9, 11]] |
| **0.310405** | **[[0], [4], [6], [8], [12], [13], [14], [15], [5, 7], [10, 9, 11], [2, 1, 3]]** |
| 0.293653 | [[0], [4], [6], [12], [13], [14], [5, 7], [10, 9, 11], [2, 1, 3], [8, 15]] |
| 0.28031 | [[4], [6], [12], [13], [14], [10, 9, 11], [2, 1, 3], [8, 15], [0, 5, 7]] |
| 0.261478 | [[4], [6], [13], [10, 9, 11], [2, 1, 3], [8, 15], [0, 5, 7], [12, 14]] |
| 0.236803 | [[13], [10, 9, 11], [2, 1, 3], [8, 15], [0, 5, 7], [12, 14], [4, 6]] |
| 0.209953 | [[10, 9, 11], [2, 1, 3], [0, 5, 7], [12, 14], [4, 6], [13, 8, 15]] |
| 0.232534 | [[0, 5, 7], [12, 14], [4, 6], [13, 8, 15], [10, 9, 11, 2, 1, 3]] |
| 0.200819 | [[0, 5, 7], [4, 6], [10, 9, 11, 2, 1, 3], [12, 14, 13, 8, 15]] |
| 0.146996 | [[10, 9, 11, 2, 1, 3], [12, 14, 13, 8, 15], [0, 5, 7, 4, 6]] |
| 0.068944 | [[10, 9, 11, 2, 1, 3], [12, 14, 13, 8, 15, 0, 5, 7, 4, 6]] |
| 0 | [[10, 9, 11, 2, 1, 3, 12, 14, 13, 8, 15, 0, 5, 7, 4, 6]] |

**Table A.9**. Complete results for the structural reducibility analysis for the Austrian dataset comparing high and low creative groups.

**USA - Human**

| q-value | Layer configuration |
|---|---|
| 0.284152 | [[0], [1], [2], [3], [4], [5], [6], [7], [8], [9], [10], [11], [12], [13], [14], [15]] |
| 0.278062 | [[0], [1], [2], [3], [5], [7], [8], [9], [10], [11], [12], [13], [14], [15], [4, 6]] |
| 0.283415 | [[0], [1], [5], [7], [8], [9], [10], [11], [12], [13], [14], [15], [4, 6], [2, 3]] |
| 0.294351 | [[0], [5], [7], [8], [9], [10], [11], [12], [13], [14], [15], [4, 6], [1, 2, 3]] |
| **0.296197** | **[[0], [5], [7], [8], [9], [12], [13], [14], [15], [4, 6], [1, 2, 3], [10, 11]]** |
| 0.284167 | [[0], [8], [9], [12], [13], [14], [15], [4, 6], [1, 2, 3], [10, 11], [5, 7]] |

| 0.292797 | [[0], [8], [12], [13], [14], [15], [4, 6], [1, 2, 3], [5, 7], [9, 10, 11]] |
|---|---|
| 0.275085 | [[0], [8], [12], [14], [4, 6], [1, 2, 3], [5, 7], [9, 10, 11], [13, 15]] |
| 0.258244 | [[0], [14], [4, 6], [1, 2, 3], [5, 7], [9, 10, 11], [13, 15], [8, 12]] |
| 0.238716 | [[14], [4, 6], [1, 2, 3], [9, 10, 11], [13, 15], [8, 12], [0, 5, 7]] |
| 0.204364 | [[4, 6], [1, 2, 3], [9, 10, 11], [13, 15], [0, 5, 7], [14, 8, 12]] |
| 0.17917 | [[1, 2, 3], [9, 10, 11], [13, 15], [14, 8, 12], [4, 6, 0, 5, 7]] |
| 0.132229 | [[1, 2, 3], [9, 10, 11], [4, 6, 0, 5, 7], [13, 15, 14, 8, 12]] |
| 0.133462 | [[4, 6, 0, 5, 7], [13, 15, 14, 8, 12], [1, 2, 3, 9, 10, 11]] |
| 0.060979 | [[1, 2, 3, 9, 10, 11], [4, 6, 0, 5, 7, 13, 15, 14, 8, 12]] |
| 0 | [[1, 2, 3, 9, 10, 11, 4, 6, 0, 5, 7, 13, 15, 14, 8, 12]] |

**Table A.10**. Complete results for the structural reducibility analysis for the US dataset comparing high and low creative groups.

**Singapore - Human**

| q-value | Layer configuration |
|---|---|
| 0.280993 | [[0], [1], [2], [3], [4], [5], [6], [7], [8], [9], [10], [11], [12], [13], [14], [15]] |
| 0.284941 | [[0], [1], [2], [3], [4], [5], [6], [7], [8], [9], [12], [13], [14], [15], [10, 11]] |
| 0.292724 | [[0], [2], [4], [5], [6], [7], [8], [9], [12], [13], [14], [15], [10, 11], [1, 3]] |
| 0.283441 | [[0], [2], [5], [7], [8], [9], [12], [13], [14], [15], [10, 11], [1, 3], [4, 6]] |
| 0.292645 | [[0], [5], [7], [8], [9], [12], [13], [14], [15], [10, 11], [4, 6], [2, 1, 3]] |
| **0.30439** | **[[0], [5], [7], [8], [12], [13], [14], [15], [4, 6], [2, 1, 3], [9, 10, 11]]** |
| 0.292239 | [[0], [8], [12], [13], [14], [15], [4, 6], [2, 1, 3], [9, 10, 11], [5, 7]] |
| 0.276318 | [[0], [8], [12], [14], [4, 6], [2, 1, 3], [9, 10, 11], [5, 7], [13, 15]] |
| 0.255989 | [[0], [8], [4, 6], [2, 1, 3], [9, 10, 11], [5, 7], [13, 15], [12, 14]] |
| 0.234708 | [[0], [4, 6], [2, 1, 3], [9, 10, 11], [5, 7], [12, 14], [8, 13, 15]] |
| 0.20478 | [[4, 6], [2, 1, 3], [9, 10, 11], [12, 14], [8, 13, 15], [0, 5, 7]] |
| 0.175531 | [[4, 6], [2, 1, 3], [9, 10, 11], [0, 5, 7], [12, 14, 8, 13, 15]] |
| 0.131162 | [[2, 1, 3], [9, 10, 11], [12, 14, 8, 13, 15], [4, 6, 0, 5, 7]] |
| 0.137797 | [[12, 14, 8, 13, 15], [4, 6, 0, 5, 7], [2, 1, 3, 9, 10, 11]] |
| 0.063201 | [[2, 1, 3, 9, 10, 11], [12, 14, 8, 13, 15, 4, 6, 0, 5, 7]] |

| 0 | [[2, 1, 3, 9, 10, 11, 12, 14, 8, 13, 15, 4, 6, 0, 5, 7]] |
|---|---|

**Table A.11**. Complete results for the structural reducibility analysis for the Singapore dataset comparing high and low creative groups.

**Italy - Human**

| q-value | Layer configuration |
|---|---|
| 0.29737 | [[0], [1], [2], [3], [4], [5], [6], [7], [8], [9], [10], [11], [12], [13], [14], [15]] |
| 0.305263 | [[0], [1], [2], [3], [4], [5], [6], [7], [8], [10], [12], [13], [14], [15], [9, 11]] |
| **0.314277** | **[[0], [2], [3], [4], [5], [6], [7], [8], [12], [13], [14], [15], [9, 11], [1, 10]]** |
| 0.302071 | [[0], [2], [3], [4], [6], [8], [12], [13], [14], [15], [9, 11], [1, 10], [5, 7]] |
| 0.288262 | [[0], [2], [3], [4], [6], [8], [12], [14], [9, 11], [1, 10], [5, 7], [13, 15]] |
| 0.274118 | [[0], [2], [3], [4], [6], [8], [9, 11], [1, 10], [5, 7], [13, 15], [12, 14]] |
| 0.277432 | [[0], [4], [6], [8], [9, 11], [1, 10], [5, 7], [13, 15], [12, 14], [2, 3]] |
| 0.258782 | [[0], [8], [9, 11], [1, 10], [5, 7], [13, 15], [12, 14], [2, 3], [4, 6]] |
| 0.272247 | [[0], [8], [5, 7], [13, 15], [12, 14], [2, 3], [4, 6], [9, 11, 1, 10]] |
| 0.292796 | [[0], [8], [5, 7], [13, 15], [12, 14], [4, 6], [2, 3, 9, 11, 1, 10]] |
| 0.272164 | [[0], [5, 7], [13, 15], [4, 6], [2, 3, 9, 11, 1, 10], [8, 12, 14]] |
| 0.244889 | [[5, 7], [13, 15], [2, 3, 9, 11, 1, 10], [8, 12, 14], [0, 4, 6]] |
| 0.206546 | [[5, 7], [2, 3, 9, 11, 1, 10], [0, 4, 6], [13, 15, 8, 12, 14]] |
| 0.142183 | [[2, 3, 9, 11, 1, 10], [13, 15, 8, 12, 14], [5, 7, 0, 4, 6]] |
| 0.064946 | [[2, 3, 9, 11, 1, 10], [13, 15, 8, 12, 14, 5, 7, 0, 4, 6]] |
| 0 | [[2, 3, 9, 11, 1, 10, 13, 15, 8, 12, 14, 5, 7, 0, 4, 6]] |

**Table A.12**. Complete results for the structural reducibility analysis for the Italian dataset comparing high and low creative groups.

-------------------------------------- AI Participants --------------------------------------

**Austria - Simulations**

| q-value | Layer configuration |
|---|---|
| 0.332769 | [[0], [1], [2], [3], [4], [5], [6], [7], [8], [9], [10], [11], [12], [13], [14], [15]] |
| **0.341638** | **[[0], [1], [2], [3], [4], [5], [6], [8], [9], [10], [11], [12], [13], [14], [7, 15]]** |
| 0.329471 | [[0], [1], [3], [5], [6], [8], [9], [10], [11], [12], [13], [14], [7, 15], [2, 4]] |
| 0.338321 | [[0], [3], [5], [6], [8], [9], [10], [11], [12], [13], [14], [2, 4], [1, 7, 15]] |
| 0.323294 | [[0], [3], [5], [6], [8], [9], [11], [13], [14], [2, 4], [1, 7, 15], [10, 12]] |
| 0.32778 | [[0], [3], [5], [6], [8], [11], [13], [2, 4], [1, 7, 15], [10, 12], [9, 14]] |
| 0.336901 | [[0], [3], [5], [8], [11], [13], [2, 4], [1, 7, 15], [10, 12], [6, 9, 14]] |
| 0.318467 | [[0], [8], [11], [13], [2, 4], [1, 7, 15], [10, 12], [6, 9, 14], [3, 5]] |
| 0.292506 | [[11], [13], [2, 4], [1, 7, 15], [10, 12], [6, 9, 14], [3, 5], [0, 8]] |
| 0.264221 | [[2, 4], [1, 7, 15], [10, 12], [6, 9, 14], [3, 5], [0, 8], [11, 13]] |
| 0.287388 | [[2, 4], [10, 12], [3, 5], [0, 8], [11, 13], [1, 7, 15, 6, 9, 14]] |
| 0.259017 | [[2, 4], [10, 12], [11, 13], [1, 7, 15, 6, 9, 14], [3, 5, 0, 8]] |
| 0.231483 | [[2, 4], [10, 12], [1, 7, 15, 6, 9, 14], [11, 13, 3, 5, 0, 8]] |
| 0.155412 | [[1, 7, 15, 6, 9, 14], [11, 13, 3, 5, 0, 8], [2, 4, 10, 12]] |
| 0.069105 | [[1, 7, 15, 6, 9, 14], [11, 13, 3, 5, 0, 8, 2, 4, 10, 12]] |
| 0 | [[1, 7, 15, 6, 9, 14, 11, 13, 3, 5, 0, 8, 2, 4, 10, 12]] |

**Table A.13**. Complete results for the structural reducibility analysis for the Austrian digital twin dataset comparing high and low creative groups.

**USA - Simulations**

| q-value | Layer configuration |
|---|---|
| 0.332821 | [[0], [1], [2], [3], [4], [5], [6], [7], [8], [9], [10], [11], [12], [13], [14], [15]] |
| 0.340688 | [[0], [1], [4], [5], [6], [7], [8], [9], [10], [11], [12], [13], [14], [15], [2, 3]] |
| 0.349578 | [[0], [1], [4], [5], [6], [7], [8], [9], [12], [13], [14], [15], [2, 3], [10, 11]] |
| 0.360484 | [[0], [4], [5], [6], [7], [8], [12], [13], [14], [15], [2, 3], [10, 11], [1, 9]] |
| 0.379561 | [[0], [4], [5], [6], [7], [8], [12], [13], [14], [15], [2, 3], [10, 11, 1, 9]] |

| **0.404232** | **[[0], [4], [5], [6], [7], [8], [12], [13], [14], [15], [2, 3, 10, 11, 1, 9]]** |
|---|---|
| 0.397213 | [[4], [5], [7], [8], [12], [13], [14], [15], [2, 3, 10, 11, 1, 9], [0, 6]] |
| 0.393019 | [[4], [5], [8], [12], [13], [14], [15], [2, 3, 10, 11, 1, 9], [7, 0, 6]] |
| 0.387187 | [[4], [5], [12], [13], [14], [15], [2, 3, 10, 11, 1, 9], [8, 7, 0, 6]] |
| 0.378563 | [[4], [5], [12], [13], [2, 3, 10, 11, 1, 9], [8, 7, 0, 6], [14, 15]] |
| 0.333837 | [[4], [13], [2, 3, 10, 11, 1, 9], [8, 7, 0, 6], [14, 15], [5, 12]] |
| 0.280542 | [[4], [2, 3, 10, 11, 1, 9], [8, 7, 0, 6], [14, 15], [13, 5, 12]] |
| 0.20315 | [[2, 3, 10, 11, 1, 9], [8, 7, 0, 6], [14, 15], [4, 13, 5, 12]] |
| 0.162342 | [[2, 3, 10, 11, 1, 9], [4, 13, 5, 12], [8, 7, 0, 6, 14, 15]] |
| 0.071441 | [[2, 3, 10, 11, 1, 9], [4, 13, 5, 12, 8, 7, 0, 6, 14, 15]] |
| 0 | [[2, 3, 10, 11, 1, 9, 4, 13, 5, 12, 8, 7, 0, 6, 14, 15]] |

**Table A.14**. Complete results for the structural reducibility analysis for the US digital twin dataset comparing high and low creative groups.

**Singapore - Simulations**

| q-value | Layer configuration |
|---|---|
| 0.335415 | [[0], [1], [2], [3], [4], [5], [6], [7], [8], [9], [10], [11], [12], [13], [14], [15]] |
| 0.343854 | [[0], [1], [2], [3], [4], [5], [6], [7], [8], [9], [12], [13], [14], [15], [10, 11]] |
| 0.352552 | [[0], [1], [4], [5], [6], [7], [8], [9], [12], [13], [14], [15], [10, 11], [2, 3]] |
| 0.3629 | [[0], [4], [5], [6], [7], [8], [12], [13], [14], [15], [10, 11], [2, 3], [1, 9]] |
| 0.38207 | [[0], [4], [5], [6], [7], [8], [12], [13], [14], [15], [2, 3], [10, 11, 1, 9]] |
| **0.407219** | **[[0], [4], [5], [6], [7], [8], [12], [13], [14], [15], [2, 3, 10, 11, 1, 9]]** |
| 0.394488 | [[0], [4], [5], [6], [7], [8], [13], [15], [2, 3, 10, 11, 1, 9], [12, 14]] |
| 0.383974 | [[0], [4], [5], [7], [13], [15], [2, 3, 10, 11, 1, 9], [12, 14], [6, 8]] |
| 0.370268 | [[4], [5], [13], [15], [2, 3, 10, 11, 1, 9], [12, 14], [6, 8], [0, 7]] |
| 0.344293 | [[4], [5], [2, 3, 10, 11, 1, 9], [12, 14], [6, 8], [0, 7], [13, 15]] |
| 0.335382 | [[4], [5], [2, 3, 10, 11, 1, 9], [12, 14], [13, 15], [6, 8, 0, 7]] |
| 0.273589 | [[2, 3, 10, 11, 1, 9], [12, 14], [13, 15], [6, 8, 0, 7], [4, 5]] |
| 0.243481 | [[2, 3, 10, 11, 1, 9], [12, 14], [4, 5], [13, 15, 6, 8, 0, 7]] |
| 0.178203 | [[2, 3, 10, 11, 1, 9], [12, 14], [4, 5, 13, 15, 6, 8, 0, 7]] |
| 0.075871 | [[2, 3, 10, 11, 1, 9], [12, 14, 4, 5, 13, 15, 6, 8, 0, 7]] |
| 0 | [[2, 3, 10, 11, 1, 9, 12, 14, 4, 5, 13, 15, 6, 8, 0, 7]] |

**Table A.15**. Complete results for the structural reducibility analysis for the Singapore digital twin dataset comparing high and low creative groups.

**Italy - Simulations**

| q-value | Layer configuration |
|---|---|
| 0.340006 | [[0], [1], [2], [3], [4], [5], [6], [7], [8], [9], [10], [11], [12], [13], [14], [15]] |
| 0.346975 | [[0], [1], [2], [4], [5], [6], [7], [8], [10], [11], [12], [13], [14], [15], [3, 9]] |
| 0.354633 | [[0], [2], [4], [5], [6], [7], [8], [10], [12], [13], [14], [15], [3, 9], [1, 11]] |
| 0.369997 | [[0], [2], [4], [5], [6], [7], [8], [10], [12], [13], [14], [15], [3, 9, 1, 11]] |
| **0.376806** | **[[0], [4], [5], [6], [7], [8], [12], [13], [14], [15], [3, 9, 1, 11], [2, 10]]** |
| 0.361158 | [[4], [6], [7], [8], [12], [13], [14], [15], [3, 9, 1, 11], [2, 10], [0, 5]] |
| 0.34345 | [[4], [6], [7], [12], [14], [15], [3, 9, 1, 11], [2, 10], [0, 5], [8, 13]] |
| 0.365851 | [[4], [6], [7], [12], [14], [15], [0, 5], [8, 13], [3, 9, 1, 11, 2, 10]] |
| 0.350759 | [[4], [6], [12], [14], [0, 5], [8, 13], [3, 9, 1, 11, 2, 10], [7, 15]] |
| 0.325952 | [[4], [6], [0, 5], [8, 13], [3, 9, 1, 11, 2, 10], [7, 15], [12, 14]] |
| 0.288796 | [[0, 5], [8, 13], [3, 9, 1, 11, 2, 10], [7, 15], [12, 14], [4, 6]] |
| 0.261925 | [[0, 5], [3, 9, 1, 11, 2, 10], [12, 14], [4, 6], [8, 13, 7, 15]] |
| 0.211646 | [[3, 9, 1, 11, 2, 10], [12, 14], [8, 13, 7, 15], [0, 5, 4, 6]] |
| 0.1698 | [[3, 9, 1, 11, 2, 10], [12, 14], [8, 13, 7, 15, 0, 5, 4, 6]] |
| 0.071446 | [[3, 9, 1, 11, 2, 10], [12, 14, 8, 13, 7, 15, 0, 5, 4, 6]] |
| 0 | [[3, 9, 1, 11, 2, 10, 12, 14, 8, 13, 7, 15, 0, 5, 4, 6]] |

**Table A.16**. Complete results for the structural reducibility analysis for the Italian digital twin dataset comparing high and low creative groups.

**VI. Full Structural Reducibility - Separate datasets**

Labelling:

| Layer Number | Abbreviation | Task Title |
|---|---|---|
| 0 | BFM | Behavioural forma mentis |
| 1 | TFM | Textual forma mentis |
| 2 | WSC-TFM_school | Woseco-TFM method, topic “school” |
| 3 | WSC-TFM_science | Woseco-TFM method, topic “science” |
| 4 | VF_school | Verbal Fluency, topic “school” |
| 5 | VF_science | Verbal Fluency, topic “science” |
| 6 | WSC_school | Woseco, topic “school” |
| 7 | WSC_science | Woseco, topic “science” |

**Table A.17**. Explanation of layer labels and abbreviations in the structural reducibility analyses.

Highlighted in the tables below is the **highest q-value** which represents the best configuration for reducing complexity while maintaining relevant information.

**Austria - Students high**

| q-value | Layer configuration |
|---|---|
| 0.254587 | [[0], [1], [2], [3], [4], [5], [6], [7]] |
| **0.269002** | **[[0], [2], [4], [5], [6], [7], [1, 3]]** |
| 0.244089 | [[0], [2], [4], [6], [1, 3], [5, 7]] |
| 0.265329 | [[0], [4], [6], [5, 7], [2, 1, 3]] |
| 0.234429 | [[4], [6], [2, 1, 3], [0, 5, 7]] |
| 0.175818 | [[2, 1, 3], [0, 5, 7], [4, 6]] |
| 0.096075 | [[2, 1, 3], [0, 5, 7, 4, 6]] |
| 0 | [[2, 1, 3, 0, 5, 7, 4, 6]] |

**Table A.18**. Complete structural reducibility analysis for the Austrian high creative students.

**Austria - Students low**

| q-value | Layer configuration |
|---|---|
| 0.253233 | [[0], [1], [2], [3], [4], [5], [6], [7]] |
| 0.267532 | [[0], [2], [4], [5], [6], [7], [1, 3]] |
| **0.289359** | **[[0], [4], [5], [6], [7], [2, 1, 3]]** |
| 0.256852 | [[4], [5], [6], [2, 1, 3], [0, 7]] |
| 0.220418 | [[5], [2, 1, 3], [0, 7], [4, 6]] |
| 0.171265 | [[2, 1, 3], [4, 6], [5, 0, 7]] |
| 0.093174 | [[2, 1, 3], [4, 6, 5, 0, 7]] |
| 0 | [[2, 1, 3, 4, 6, 5, 0, 7]] |

**Table A.19**. Complete structural reducibility analysis for the Austrian low creative students.

**USA - Students high**

| q-value | Layer configuration |
|---|---|
| 0.235374 | [[0], [1], [2], [3], [4], [5], [6], [7]] |
| 0.218761 | [[0], [1], [2], [3], [5], [7], [4, 6]] |
| 0.22644 | [[0], [1], [5], [7], [4, 6], [2, 3]] |
| **0.250036** | **[[0], [5], [7], [4, 6], [1, 2, 3]]** |
| 0.210871 | [[0], [4, 6], [1, 2, 3], [5, 7]] |
| 0.157218 | [[4, 6], [1, 2, 3], [0, 5, 7]] |
| 0.083445 | [[1, 2, 3], [4, 6, 0, 5, 7]] |
| 0 | [[1, 2, 3, 4, 6, 0, 5, 7]] |

**Table A.20**. Complete structural reducibility analysis for the US high creative students.

**USA - Students low**

| q-value | Layer configuration |
|---|---|
| 0.249495 | [[0], [1], [2], [3], [4], [5], [6], [7]] |
| 0.254524 | [[0], [1], [4], [5], [6], [7], [2, 3]] |

| **0.274603** | **[[0], [4], [5], [6], [7], [1, 2, 3]]** |
|---|---|
| 0.2478 | [[0], [4], [6], [1, 2, 3], [5, 7]] |
| 0.219034 | [[6], [1, 2, 3], [5, 7], [0, 4]] |
| 0.158493 | [[1, 2, 3], [5, 7], [6, 0, 4]] |
| 0.079456 | [[1, 2, 3], [5, 7, 6, 0, 4]] |
| 0 | [[1, 2, 3, 5, 7, 6, 0, 4]] |

**Table A.21**. Complete structural reducibility analysis for the US low creative students.

**Singapore - Students high**

| q-value | Layer configuration |
|---|---|
| 0.244179 | [[0], [1], [2], [3], [4], [5], [6], [7]] |
| **0.259059** | **[[0], [2], [4], [5], [6], [7], [1, 3]]** |
| 0.237603 | [[0], [2], [5], [7], [1, 3], [4, 6]] |
| 0.257744 | [[0], [5], [7], [4, 6], [2, 1, 3]] |
| 0.221969 | [[0], [4, 6], [2, 1, 3], [5, 7]] |
| 0.16639 | [[4, 6], [2, 1, 3], [0, 5, 7]] |
| 0.087203 | [[2, 1, 3], [4, 6, 0, 5, 7]] |
| 0 | [[2, 1, 3, 4, 6, 0, 5, 7]] |

**Table A.22**. Complete structural reducibility analysis for the Singapore high creative students.

**Singapore - Students low**

| q-value | Layer configuration |
|---|---|
| 0.243415 | [[0], [1], [2], [3], [4], [5], [6], [7]] |
| 0.253171 | [[0], [1], [4], [5], [6], [7], [2, 3]] |
| **0.276481** | **[[0], [4], [5], [6], [7], [1, 2, 3]]** |
| 0.251439 | [[0], [4], [6], [1, 2, 3], [5, 7]] |
| 0.212968 | [[0], [1, 2, 3], [5, 7], [4, 6]] |
| 0.160857 | [[1, 2, 3], [4, 6], [0, 5, 7]] |
| 0.085219 | [[1, 2, 3], [4, 6, 0, 5, 7]] |
| 0 | [[1, 2, 3, 4, 6, 0, 5, 7]] |

**Table A.23**. Complete structural reducibility analysis for the Singapore low creative students.

**Italy - Students high**

| q-value | Layer configuration |
|---|---|
| 0.271636 | [[0], [1], [2], [3], [4], [5], [6], [7]] |
| 0.245592 | [[0], [1], [2], [3], [4], [6], [5, 7]] |
| 0.25709 | [[0], [3], [4], [6], [5, 7], [1, 2]] |
| **0.275592** | **[[0], [4], [6], [5, 7], [3, 1, 2]]** |
| 0.23817 | [[0], [5, 7], [3, 1, 2], [4, 6]] |
| 0.188492 | [[5, 7], [3, 1, 2], [0, 4, 6]] |
| 0.093746 | [[3, 1, 2], [5, 7, 0, 4, 6]] |
| 0 | [[3, 1, 2, 5, 7, 0, 4, 6]] |

**Table A.24**. Complete structural reducibility analysis for the Italian high creative students.

**Italy - Students low**

| q-value | Layer configuration |
|---|---|
| 0.259415 | [[0], [1], [2], [3], [4], [5], [6], [7]] |
| **0.277308** | **[[0], [2], [4], [5], [6], [7], [1, 3]]** |
| 0.250403 | [[0], [2], [4], [6], [1, 3], [5, 7]] |
| 0.217788 | [[0], [2], [1, 3], [5, 7], [4, 6]] |
| 0.231886 | [[0], [5, 7], [4, 6], [2, 1, 3]] |
| 0.180934 | [[5, 7], [2, 1, 3], [0, 4, 6]] |
| 0.088695 | [[2, 1, 3], [5, 7, 0, 4, 6]] |
| 0 | [[2, 1, 3, 5, 7, 0, 4, 6]] |

**Table A.25**. Complete structural reducibility analysis for the Italian low creative students.

------------------------------------- AI Participants -------------------------------------

**AT - Simulation high**

| q-value | Layer configuration |
|---|---|
| 0.299169 | [[0], [1], [2], [3], [4], [5], [6], [7]] |
| **0.313509** | **[[0], [2], [3], [4], [5], [6], [1, 7]]** |
| 0.282919 | [[0], [3], [5], [6], [1, 7], [2, 4]] |
| 0.299786 | [[0], [3], [5], [2, 4], [6, 1, 7]] |
| 0.257822 | [[0], [2, 4], [6, 1, 7], [3, 5]] |
| 0.193301 | [[2, 4], [6, 1, 7], [0, 3, 5]] |
| 0.093538 | [[6, 1, 7], [2, 4, 0, 3, 5]] |
| 0 | [[6, 1, 7, 2, 4, 0, 3, 5]] |

**Table A.26**. Complete structural reducibility analysis for Austrian high creative digital twins.

**AT - Simulation low**

| q-value | Layer configuration |
|---|---|
| 0.303017 | [[0], [1], [2], [3], [4], [5], [6], [7]] |
| **0.317626** | **[[0], [2], [3], [4], [5], [6], [1, 7]]** |
| 0.289309 | [[0], [3], [5], [6], [1, 7], [2, 4]] |
| 0.307934 | [[0], [3], [5], [2, 4], [6, 1, 7]] |
| 0.265445 | [[0], [2, 4], [6, 1, 7], [3, 5]] |
| 0.19955 | [[2, 4], [6, 1, 7], [0, 3, 5]] |
| 0.099006 | [[6, 1, 7], [2, 4, 0, 3, 5]] |
| 0 | [[6, 1, 7, 2, 4, 0, 3, 5]] |

**Table A.27**. Complete structural reducibility analysis for Austrian low creative digital twins.

**USA - Simulation high**

| q-value | Layer configuration |
|---|---|
| 0.30816 | [[0], [1], [2], [3], [4], [5], [6], [7]] |
| 0.326045 | [[0], [1], [4], [5], [6], [7], [2, 3]] |
| **0.355824** | **[[0], [4], [5], [6], [7], [1, 2, 3]]** |
| 0.336367 | [[4], [5], [7], [1, 2, 3], [0, 6]] |
| 0.317446 | [[4], [5], [1, 2, 3], [7, 0, 6]] |
| 0.211423 | [[1, 2, 3], [7, 0, 6], [4, 5]] |
| 0.10243 | [[1, 2, 3], [7, 0, 6, 4, 5]] |
| 0 | [[1, 2, 3, 7, 0, 6, 4, 5]] |

**Table A.28**. Complete structural reducibility analysis for US high creative digital twins.

**USA - Simulation low**

| q-value | Layer configuration |
|---|---|
| 0.302052 | [[0], [1], [2], [3], [4], [5], [6], [7]] |
| 0.319075 | [[0], [1], [4], [5], [6], [7], [2, 3]] |
| **0.347456** | **[[0], [4], [5], [6], [7], [1, 2, 3]]** |
| 0.332134 | [[0], [4], [5], [1, 2, 3], [6, 7]] |
| 0.309397 | [[4], [5], [1, 2, 3], [0, 6, 7]] |
| 0.203184 | [[1, 2, 3], [0, 6, 7], [4, 5]] |
| 0.095718 | [[1, 2, 3], [0, 6, 7, 4, 5]] |
| 0 | [[1, 2, 3, 0, 6, 7, 4, 5]] |

**Table A.29**. Complete structural reducibility analysis for US low creative digital twins.

**SG - Simulation high**

| q-value | Layer configuration |
|---|---|
| 0.314131 | [[0], [1], [2], [3], [4], [5], [6], [7]] |
| 0.331779 | [[0], [1], [4], [5], [6], [7], [2, 3]] |

| **0.360439** | **[[0], [4], [5], [6], [7], [1, 2, 3]]** |
|---|---|
| 0.337902 | [[4], [5], [6], [1, 2, 3], [0, 7]] |
| 0.316247 | [[4], [5], [1, 2, 3], [6, 0, 7]] |
| 0.211678 | [[1, 2, 3], [6, 0, 7], [4, 5]] |
| 0.106863 | [[1, 2, 3], [6, 0, 7, 4, 5]] |
| 0 | [[1, 2, 3, 6, 0, 7, 4, 5]] |

**Table A.30**. Complete structural reducibility analysis for Singapore high creative digital twins.

**SG - Simulation low**

| q-value | Layer configuration |
|---|---|
| 0.304256 | [[0], [1], [2], [3], [4], [5], [6], [7]] |
| 0.322234 | [[0], [1], [4], [5], [6], [7], [2, 3]] |
| **0.350193** | **[[0], [4], [5], [6], [7], [1, 2, 3]]** |
| 0.316905 | [[0], [5], [7], [1, 2, 3], [4, 6]] |
| 0.26034 | [[7], [1, 2, 3], [4, 6], [0, 5]] |
| 0.200379 | [[1, 2, 3], [4, 6], [7, 0, 5]] |
| 0.101952 | [[1, 2, 3], [4, 6, 7, 0, 5]] |
| 0 | [[1, 2, 3, 4, 6, 7, 0, 5]] |

**Table A.31**.Complete structural reducibility analysis for Singapore low creative digital twins.

**IT - Simulation high**

| q-value | Layer configuration |
|---|---|
| 0.319584 | [[0], [1], [2], [3], [4], [5], [6], [7]] |
| **0.334079** | **[[0], [2], [4], [5], [6], [7], [1, 3]]** |
| 0.301153 | [[2], [4], [6], [7], [1, 3], [0, 5]] |
| 0.318653 | [[4], [6], [7], [0, 5], [2, 1, 3]] |
| 0.265418 | [[7], [0, 5], [2, 1, 3], [4, 6]] |
| 0.202673 | [[2, 1, 3], [4, 6], [7, 0, 5]] |
| 0.102144 | [[2, 1, 3], [4, 6, 7, 0, 5]] |
| 0 | [[2, 1, 3, 4, 6, 7, 0, 5]] |

**Table A.32**. Complete structural reducibility analysis for Italian high creative digital twins.

**IT - Simulation low**

| q-value | Layer configuration |
|---|---|
| 0.313163 | [[0], [1], [2], [3], [4], [5], [6], [7]] |
| 0.322996 | [[0], [1], [4], [5], [6], [7], [2, 3]] |
| **0.3453** | **[[0], [4], [5], [6], [7], [1, 2, 3]]** |
| 0.309787 | [[4], [6], [7], [1, 2, 3], [0, 5]] |
| 0.26009 | [[7], [1, 2, 3], [0, 5], [4, 6]] |
| 0.195337 | [[1, 2, 3], [4, 6], [7, 0, 5]] |
| 0.09878 | [[1, 2, 3], [4, 6, 7, 0, 5]] |
| 0 | [[1, 2, 3, 4, 6, 7, 0, 5]] |

**Table A.33**. Complete structural reducibility analysis for Italian low creative digital twins.